\documentclass{article}

\usepackage{arxiv}

\usepackage[utf8]{inputenc} 
\usepackage[T1]{fontenc}    
\usepackage{hyperref}       
\usepackage{url}            
\usepackage{booktabs}       
\usepackage{amsfonts}       
\usepackage{nicefrac}       
\usepackage{microtype}      
\usepackage{lipsum}		
\usepackage{graphicx}

\usepackage{dcolumn}
\usepackage{bm}
\usepackage{amsmath}
\usepackage{amssymb}
\usepackage{amscd}
\usepackage{amsthm}
\usepackage{amsfonts}
\usepackage{graphicx}
\usepackage{multirow}

\date{\today}

\author{
Salom\'{e} A. Sep\'{u}veda Fontaine\\
Centro de Investigaci\'{o}n Operativa, \\
Universidad Miguel Hern\'{a}ndez, \\
Elche, 03202 Alicante, Spain \\
\texttt{salome.sepulveda.fontaine@gmail.com}
\And
Jos\'{e} M. Amig\'{o} \\
Centro de Investigaci\'{o}n Operativa, \\
Universidad Miguel Hern\'{a}ndez, \\
Elche, 03202 Alicante, Spain \\
\texttt{jm.amigo@umh.es}
}
\renewcommand{\headeright}{}
\renewcommand{\undertitle}{}

\begin{document}

\title{Applications of Entropy in Data Analysis and Machine Learning: A Review}
\maketitle

\begin{abstract}
Since its origin in the thermodynamics of the 19th
century, the concept of entropy has also permeated other fields of physics
and mathematics, such as Classical and Quantum Statistical Mechanics,
Information Theory, Probability Theory, Ergodic Theory and the Theory of
Dynamical Systems. Specifically, we are referring to the classical entropies: the Boltzmann-Gibbs, von Neumann, Shannon, Kolmogorov-Sinai and topological entropies. In addition to their common name, which is historically justified (as we briefly describe in this review), other commonality of the classical entropies is the important role that they have played and are still playing in the theory and applications of their respective fields and beyond. Therefore, it is not surprising that, in the course of time, many other instances of the overarching concept of entropy have been proposed, most of them tailored to specific purposes. Following the current usage, we will refer to all of them, whether classical or new, simply as entropies. Precisely, the subject of this review is their applications in data analysis and machine learning. The reason for these particular applications is that entropies are very well suited to characterize probability mass distributions, typically generated by finite-state processes or symbolized signals. Therefore, we will focus on entropies defined as positive functionals on probability mass distributions and provide an axiomatic characterization that goes back to Shannon and Khinchin. Given the plethora of entropies in the literature, we have selected a representative group, including the classical ones. The applications summarized in this review finely illustrate the power and versatility of entropy in data analysis and machine learning.
\end{abstract}


\section{\large Introduction}
\label{secIntro}

\subsection{Aims and scope}
\label{sec1.2}

The motivation for this review is twofold. First, to give researchers in data analysis and machine learning an overview of the applications of the concept of entropy (in a broad sense) in those two fields, both in past and recent times. Needless to say, data analysis and machine learning are hot topics in current research, this being the ultimate reason for choosing them. The second motivation is to familiarize the interested researchers with the entropy toolkit. To pave the way, we give a historical and mathematical background of different entropies in use.    

Entropy is a concept that appears in different areas of physics and
mathematics with different meanings. Thus, entropy is a measure of: (i)
disorder in Statistical Mechanics, (ii) uncertainty in Information and
Probability Theories, (iii) (pseudo-)randomness in the Theory of
Measure-preserving Dynamical Systems, and (iv) complexity in Topological
Dynamics. This versatility explains why entropy has found extensive
applications across various scientific disciplines since its inception in
the 19th century.

Precisely, this paper aims to provide an up-to-date overview of the
applications of entropy in data analysis and machine learning, where entropy stands here not only for the traditional instances but also for more recent proposals inspired by them. In data analysis, entropy is a powerful tool
for detection of dynamical changes, segmentation, clustering, discrimination, etc. In machine learning, it is used for classification, feature extraction, optimization of algorithms, anomaly detection, and more. 

The applications of entropy to data analysis date back to Shannon's founding paper of Information Theory \cite{Shannon1948}. Thus, in Section 7 of \cite{Shannon1948} Shannon introduces the entropy-based concept of redundancy of a communication source (i.e., a stationary random process) and compares the redundancies of Basic English and James Joyce's prose. With the advent of nonlinear time series analysis in the 1980s \cite{Kantz1999}, entropy gained momentum and became a standard tool of the data analysts. The applications of entropy to machine learning may also be traced back to the 1980s at least, when the first energy-based models were formulated \cite{Goodfellow2016}. They are also called Boltzmann machines because the probability of finding the model in a certain configuration is given by the Boltzmann distribution (see Section \ref{secSKaxioms} for details). Since then, entropy has found many applications in machine learning, especially via cost functions and the maximum entropy principle, as we will see in Section \ref{secEntropyList}.

The ability of entropy to provide
insights into data structure and algorithm performance has led to a
widespread search for further applications and new proposals tailored to specific needs, both in data analysis and machine learning. This being the case, the present review will be useful for researchers in the above two fields, interested in the theoretical basics and/or the current applications of entropy. Along with established applications, the authors also have taken into account innovative proposals to reflect the intense research activity on entropy that is currently underway.

At this point, the reader may be wondering what entropy is. A search for the
word \textquotedblleft entropy\textquotedblright\ on the Internet returns a
large number of results, some of them also called entropy metrics, entropy-like measures or entropy-based indices in the literature. So, what is entropy actually?


\subsection{Classical entropies}
\label{secSKaxioms}

Historically, the word \textquotedblleft entropy\textquotedblright\ was
introduced by the German physicist Clausius in Thermodynamics in 1865 to designate the amount of internal energy in a system that cannot be
transformed into work. In particular, entropy determines the equilibrium of a thermodynamical system, namely, the state of maximum entropy consistent with the macroscopic constraints. In the second half of the 19th century, entropy was given a microscopic
interpretation in the foundational works of Boltzmann and Gibbs on
Statistical Mechanics. In 1927, von Neumann generalized the
Boltzmann-Gibbs entropy to the then-emerging theory of Quantum Mechanics \cite{MathematischeGrundlagen}. In 1948, the word entropy appeared in a completely different context: Information Theory. If entropy is a measure of disorder in Statistical Mechanics; in the seminal paper of Shannon \cite{Shannon1948}, the creator of Information Theory, entropy stands for the average uncertainty about the outcome of a random variable (or the information conveyed by knowing it). Albeit in different realms, the coincidence in names is explained because Shannon's formula (see equation (\ref{Shannon}) below) is formally the same as Gibbs' for the entropy of a system in thermal equilibrium with a heat bath at constant temperature \cite{gibbs1902}. 

This abridged history of entropy continues with Kolmogorov, who
crafted Shannon's entropy into a useful invariant in Ergodic Theory \cite{kolmogorov1959}, and his student Sinai, who adapted Kolmogorov's ideas to the theory of
measure-preserving dynamical systems \cite{sinai1959}. In turn, Adler,
Konheim and McAndrew \cite{Adler1965} generalized the Kolmogorov-Sinai (KS) entropy from
measure-preserving dynamics to topological dynamics under the name of
topological entropy. According to the Variational Principle, topological
entropy is a tight upper bound of the KS entropies of dynamical systems endowed with certain probability measures \cite{walters2000}.

To get down to the mathematical formulas, let $\mathcal{P}$ be the set of
probability mass distributions $\{p_{1},...,p_{W}\}$ for all $W\geq 2$. Then,
the Shannon entropy of the probability distribution $\mathrm{p}%
_{W}=\{p_{1},...,p_{W}\}$ is defined as%
\begin{equation}
S(\mathrm{p}_{W})=S(p_{1},...,p_{W})=-\sum\limits_{i=1}^{W}p_{i}\log p_{i}
\label{Shannon}
\end{equation}%
where the choice of the logarithm base fixes the unit of the entropy. The
usual choices being $2$ (bit), $e$ (nat) or $10$ (dit). If $p_{i}=0$, then $%
0\log 0:=\lim_{x\rightarrow 0+}x\log x=0$. Mathematically, equation (\ref%
{Shannon}) is the expected value of the \textit{information function} $%
I(X)=-\log p(X)$, where $X$ is a random variable with probability
distribution $\mathrm{p}_{W}$. Since entropy is the cornerstone of
Information Theory, Shannon also justified definition (\ref{Shannon}) by
proving in his seminal paper \cite{Shannon1948} that it is unique (except for a positive
factor) under a few, general assumptions. In their modern (equivalent)
formulation, these assumptions are called the \textit{Shannon-Khinchin axioms} \cite{khinchin1957}, that we state below.

A positive functional $H$ on $\mathcal{P}$, i.e., a map $H:\mathcal{P}%
\rightarrow \textbf{R}^{+}$ ($\textbf{R}^{+}$ being the non-negative real
numbers), is an entropy if it satisfies the following properties:

\begin{description}
\item[SK1] \textit{Continuity}. $H(p_{1},...,p_{W})$ depends continuously on
all variables for each $W$.

\item[SK2] \textit{Maximality}. For all $W$, 
\begin{equation*}
H(p_{1},...,p_{W})\leq H(\tfrac{1}{W},...,\tfrac{1}{W}).
\end{equation*}

\item[SK3] \textit{Expansibility}. For all $W$ and $1\leq i\leq W$,%
\begin{equation*}
H(0,p_{1},...,p_{W})=H(p_{1},...,p_{i},0,p_{i+1},...,p_{W})=H(p_{1},...,p_{i},p_{i+1},...,p_{W}).
\end{equation*}

\item[SK4] \textit{Strong additivity} (or \textit{separability}). For all $%
W,U$,%
\begin{align}
& H(p_{11},\ldots ,p_{1U},p_{21},\ldots p_{2U},\ldots ,p_{W1},\ldots ,p_{WU})%
\hspace*{3cm}  \label{add} \\
& =H(p_{1\ast },p_{2\ast },\ldots ,p_{W\ast })+\sum_{i=1}^{W}p_{i\ast
}H\left( \frac{p_{i1}}{p_{i\ast }},\frac{p_{i2}}{p_{i\ast }},\ldots ,\frac{%
p_{iU}}{p_{i\ast }}\right) ,  \notag
\end{align}%
where $p_{i\ast }=\sum_{j=1}^{U}p_{ij}$.
\end{description}

Axiom SK4 can be formulated in a more compact way as 
\begin{equation*}
H(X,Y)=H(X)+H(Y\left\vert X\right) ,
\end{equation*}%
where $X$ and $Y$ are random variables with probability distributions $%
\{p_{i\ast }:1\leq i\leq W\}$ and $\{p_{\ast j}=\sum_{i=1}^{W}p_{ij}:1\leq
j\leq U\}$ respectively, $H(X,Y)=H(p_{11},\ldots ,p_{1U},...,p_{W1},\ldots
,p_{WU})$ and $H(Y\left\vert X\right) $ is the entropy of $Y$ conditional on 
$X$, i.e., the expected value of the conditional distributions $p(y\left\vert x\right)$, averaged over the conditioning variable $X$ \cite{Cover2006}. In particular, if $X$ and $Y$ are independent (i.e., $%
p_{ij}=p_{i\ast }p_{\ast j}$), then $H(Y\left\vert X\right) =H(Y)$ and%
\begin{equation}
H(X,Y)=H(X)+H(Y).  \label{add2}
\end{equation}%
If $H$ satisfies equation (\ref{add2}) for independent random variables $X$
and $Y$, then it is called \textit{additive}.

It was proved in \cite{Shannon1948} and \cite{khinchin1957} that a positive functional $H$ on $\mathcal{P}$ that
fulfills Axioms SK1-SK4 is necessarily of the form%
\begin{equation}
H(p_{1},...,p_{W})=-k\sum\limits_{i=1}^{W}p_{i}\log
p_{i}=:S_{BGS}(p_{1},...,p_{W})  \label{G_ent}
\end{equation}%
for every $W\geq 2$, where $k$ is a positive constant. For historical reasons, $S_{BGS}$ is usually called the \textit{Boltzmann-Gibbs-Shannon entropy}. In Physics, $k$ is the Boltzmann constant $k_{B}=1.3806504(24)\times10^{-23}$ J/K and $\log$ is the natural logarithm. In Information Theory, $k=1$ and $\log$ is the base 2 logarithm when dealing with digital communications. The particular case%
\begin{equation}
S_{BGS}(1/W,...,1/W)=k\log W,  \label{Boltzmann}
\end{equation}%
obtained for uniform distributions is sometimes referred to as the \textit{%
Boltzmann entropy}, although the expression (\ref{Boltzmann}) is actually due to Planck \cite{planck1900}. According to Axiom SK2, the Boltzmann entropy is the maximum of $S_{BGS}$.

The same conclusion about the uniqueness of $S_{BGS}$ can be derived using
other equivalent properties \cite{csiszar2008}. Since we are not interested in physical applications here, we set $k=1$ and generally refer to $S_{BGS}$ as Shannon's entropy.  

In Section \ref{sec1.2} we mentioned the Boltzmann distribution, not to be confused
with the Boltzmann entropy (\ref{Boltzmann}). It is defined as $p_{i}\propto \exp (-E_{i})$, $%
1\leq i\leq W$, where $E_{i}\geq 0$ is known as the energy function, so that 
$p_{i}>0$ for all $i$. In statistical mechanics, $E_{i}=\beta \varepsilon
_{i}$, where $\varepsilon _{i}$ is the energy of the $i$th state and $%
1/\beta =k_{B}T$, $T$ being the absolute temperature of the system; the
normalization factor $Z=\sum_{i}\exp (-\varepsilon _{i}/k_{B}T)$ is called
the partition function of the system. While the uniform distribution
maximizes the unsconstrained Shannon entropy, the Boltzmann distribution is
the maximizer of Shannon's entropy under the contraint of a prescribed
average energy, that is, $\sum_{i}p_{i}\varepsilon _{i}=\bar{\varepsilon}$ \cite{Cover2006,Amigo2022}.   

In 1961 R\'{e}nyi proposed a generalization of Shannon's entropy by using a
different, more general definition of expectation value \cite{renyi1961entropy,Amigo2018}: For any real $\alpha >0$%
, $\alpha \neq 1$, \textit{R\'{e}nyi entropy} $R_{\alpha }$ is defined as 
\begin{equation}
R_{\alpha }(p_{1},...,p_{W})=\frac{1}{1-\alpha }\log \left(
\sum\limits_{i=1}^{W}p_{i}^{\alpha }\right) .  \label{Renyi_ent}
\end{equation}%
So, R\'{e}nyi entropy is actually a family of entropies; in particular, $%
R_{1}:=\lim_{\alpha \rightarrow 1}R_{\alpha }=S_{BGS}$. Other limiting cases
are $R_{0}:=\lim_{\alpha \rightarrow 0}R_{\alpha }=\log W$, called \textit{%
Hartley} or \textit{max-entropy}, which coincides with the Boltzmann entropy (\ref{Boltzmann}) except for the value of the constant $k$, and $R_{\infty }:=\lim_{\alpha
\rightarrow \infty }R_{\alpha }=\min_{1\leq i\leq W}(-\ln p_{i})$, called
the \textit{min-entropy}. These names are due to the non-increasing monotonicity
of R\'{e}nyi's entropy with respect to the parameter: $R_{\alpha
}\geq R_{\beta }$ for $\alpha <\beta $.

It is easy to show that R\'{e}nyi's entropy satisfies Axioms SK1-SK3 but not
SK4. Instead of strong additivity, $R_{\alpha }$ satisfies \textit{additivity}: 
\begin{equation*}
R_{\alpha }(\mathrm{p}_{U}\times \mathrm{q}_{W})=R_{\alpha }(\mathrm{p}%
_{U})+R_{\alpha }(\mathrm{q}_{W}),
\end{equation*}%
see equation (\ref{add2}).

A final milestone in this short history of entropy is the introduction of 
\textit{non-additive entropies} by Havrda and Charv\'{a}t in Information
Theory \cite{HCentropy} and Tsallis in Statistical Mechanics \cite{Tsallis1988}, which are equivalent and
usually called the Tsallis entropy:%
\begin{equation}
T_{q}(p_{1},...,p_{W})=\frac{1}{1-q}\left(
\sum\limits_{i=1}^{W}p_{i}^{q}-1\right)  \label{Tsallis_entr}
\end{equation}%
for any real $q>0$, $q\neq 1$. Again, Tsallis entropy is a family of entropies that
satisfies Axioms SK1-SK3 but not SK4. Instead, $T_{q}$ is \textquotedblleft $%
q$-additive\textquotedblright , meaning that 
\begin{equation*}
T_{q}(\mathrm{p}_{U}\times \mathrm{q}_{W})=T_{q}(\mathrm{p}_{U})+T_{q}(%
\mathrm{q}_{W})+(1-q)T_{q}(\mathrm{p}_{U})T_{q}(\mathrm{q}_{W}).
\end{equation*}%
As R\'{e}nyi's entropy, the Tsallis entropy is a generalization of Shannon's
entropy in the sense that $T_{1}:=\lim_{q\rightarrow 1}T_{q}=S_{BGS}$.
Formally, $T_{q}$ can be obtained from $S_{BGS}$ by replacing the logarithm
in equation (\ref{G_ent}) by the \textquotedblleft $q$%
-logarithm\textquotedblright\ \cite{Amigo2018}.

The appearance of generalizations of the Shannon entropy prompted the weaker
concept of \textit{generalized entropy}: a positive functional on
probability distributions that satisfies Axioms SK1-SK3. Therefore, the BGS
entropy, together with the R\'{e}nyi and Tsallis entropies are examples of
generalized entropies. Shannon's uniqueness theorem can then be rephrased by
saying that, the only generalized entropy that is strongly additive is the
BGS entropy. Axioms SK1-SK3 are arguably the minimal requirements for a positive functional on probability mass distributions to be called an entropy. Most \textquotedblleft entropies\textquotedblright\ proposed since the formulation of Rényi and Tsallis' entropies are precisely generalized entropies in the axiomatic sense. 

To wrap up this short account of the classical and generalized entropies,
let us mention that Shannon's, R\'{e}nyi's and Tsallis'  entropies (and other entropies for that matter) have
counterparts for continuous-valued random variables and processes (i.e.,
defined on probability densities). These \textquotedblleft
differential\textquotedblright\ versions are formally obtained by replacing
probability mass functions by probability densities and summations by
integrations in equations (\ref{Shannon}), (\ref{Renyi_ent}) and (\ref%
{Tsallis_entr}), respectively. Although also useful in applications,
differential entropies may lack important properties of their discrete
counterparts. For example, \textit{differential} (Shannon's) \textit{entropy} lacks positivity \cite{Cover2006}.


\subsection{Methodology and organization of this review}
\label{sec1.3}

As said above, the primary objective of this work is to review the
applications of entropy in the fields of data analysis and machine
learning. In view of the many versions of entropy currently in use, we had to
make a selection of them based on their relevance in general, and the
interest of their applications in particular. Inevitably, this selection may have been influenced by the experience and research interests of the authors, so we apologize for any oversights of relevant applications and references. Apart from the group of classical entropies of Section \ref{secSKaxioms}, which we will call \textbf{G0} for reference purposes, the remaining entropies selected for this review can be classified into the following two groups. 

\begin{description}

\item[G1] Entropies based on (or inspired by) the entropies in Group G0, especially Shannon's entropy (\ref{Shannon}). This group comprises: \textit{Bubble entropy}, \textit{dispersion entropy} and \textit{fluctuation-based dispersion entropy}, \textit{energy entropy} and \textit{empirical mode decomposition energy entropy}, \textit{entanglement entropy}, \textit{Fourier entropy and fractional Fourier entropy}, \textit{graph entropy}, \textit{Kaniadakis entropy}, \textit{permutation entropy}, \textit{Rao's quadratic entropy}, \textit{spectral entropy}, \textit{tone entropy}, and \textit{wavelet entropy}.

\item[G2] Entropies based on information-theoretical concepts such as
the correlation integral, divergences, unconditional or conditional mutual information, or based on other entropies in this group. This group comprises: \textit{Approximate entropy}, \textit{cross entropy} and \textit{categorical cross entropy}, \textit{excess entropy}, \textit{fuzzy entropy}, \textit{intrinsic mode entropy}, \textit{kernel entropy}, \textit{rank-based entropy}, \textit{relative entropy} or \textit{Kullback-Leibler divergence}, \textit{sample entropy}, and \textit{transfer entropy}.

\end{description}

In Section \ref{secEntropyList}, each of the 33 selected entropies is assigned a subsection in alphabetical order. For brevity, mathematical definitions are reminded only when feasible in a short space; otherwise, we give a qualitative account and refer the reader to the original publications or standard bibliography for the formulas. The corresponding applications to data analysis and machine learning are explained with brief but sufficient descriptions and provided with specific references. Practical issues such as the choice of parameters or computational implementations are beyond the scope of this review.

Section \ref{sec3.0} contains general comments on the previous sections. Moreover, it presents the most common applications of Section \ref{secEntropyList} and a citation analysis of the references to identify the most influential papers to date in the field of entropy and its applications. Section \ref{sec4.0} begins with the concluding remarks of this review and ends with some recent developments and challenges. The aim of that outlook is to attract new researchers to the field. This review ends with a list of references that is long by usual standards, though very short compared to the vast and ever-growing literature on entropy and applications thereof.

There is a number of excellent general reviews on entropy, entropy-like quantities and their applications. Thus, Katok revisits in \cite{Katok2007} the fifty years 1958-2007 of entropy in the theory of dynamical systems (i.e., Kolmogorov-Sinai and topological entropies), which is recommended to the readers interested not only in the mathematical aspects of those entropies but also in their historical development. The reviews \cite{Amigo2022} and  \cite{Amigo2018} are mainly mathematical accounts of the classical and generalized entropies, along with some typical applications. The review “The entropy Universe” by M. Ribeiro et al. \cite{Ribeiro2021} is a fine blend of mathematical descriptions of several entropies and their applications in science and engineering. Unlike the aforementioned reviews, the subject of the present work are the applications of entropy to data analysis and machine learning. Hence, this review differs from \cite{Katok2007,Amigo2022,Amigo2018} in that the latter focus on the mathematical underpinnings of the considered entropies. And it also differs from \cite{Ribeiro2021} in that that review focuses on the definitions of the selected entropies rather than on the applications. Furthermore,  both the entropies and applications considered in \cite{Ribeiro2021} are more limited in number and scope than in ours; in particular, applications to machine learning are not contemplated therein.

\section{\large  Applications in Data Analysis and Machine Learning}
\label{secEntropyList}

In this section, the selected entropies are sorted alphabetically. To streamline the exposition, multiscale and weighted versions \cite{madalena2005} are included in the same section as the original entropy. The corresponding applications to data analysis and machine learning are tagged with some keywords in alphabetical order so that reverse search (i.e., searching for entropies for a given application) can also be easily performed.

\subsection{Approximate Entropy}
\label{sec2}

Approximate entropy was proposed by Pincus in 1991 \cite{pincusAppox1991} to analyze medical data. Loosely speaking, approximate entropy is a heuristic implementation of the correlation integral with time series data \cite{Amigo2022}. The approximate entropy depends on a parameter $r>0$, sometimes called tolerance, which is a cut-off that defines the concept of proximity between points via the Heaviside step function $\Theta (r-\left\Vert x_{i}-x_{j}\right\Vert )$ (where $\Theta(z)=1$ if $z\ge 0$, and 0 otherwise). It quantifies the change in the relative frequencies of length-$k$ time-delay vectors with increasing $k$. A modified version of approximate entropy was proposed in 2000 under the name sample entropy (Section \ref{sec33}). See \cite{DelgadoBonal2019} for a tutorial.

\begin{description}

\item[{Applications}]
\end{description}

\begin{itemize}

\item \textbf{Alzheimer's disease.} Approximate entropy has been used in the non-linear analysis of EEGs and MEGs from patients with Alzheimer's disease \cite{hornero2009}, \cite{morabito2012}.

\item \textbf{Anesthetic drug effects.} Another field of applications is the quantification of anesthetic drug effects on the brain activity as measured by EEGs, including comparative testing of different anesthetics \cite{liang2015}.

\item \textbf{Emotion recognition.} Along with other entropies, approximate entropy has been used for EEG-based human emotion recognition \cite{patel2021}. 

\item \textbf{Epileptic seizure detection.} Approximate entropy has also been used as biomarker in algorithms for epileptic EEG analysis, in particular, for epileptic seizure detection \cite{kannathal2005}, \cite{srinivasan2007}, \cite{jouny2012}.

\item \textbf{Physiological time series.} See \cite{richman2000} for an overview of applications of approximate entropy to the analysis of physiological time series. 

\item \textbf{Sleep research.} The applications of approximate entropy include sleep research, in particular, the separation of sleep stages based on EEG data \cite{acharya2005}. 

\end{itemize}


\subsection{Bubble Entropy}
\label{sec5}
   
Bubble entropy is a metric that evaluates changes in the order of data segments in time series when a new element is added. It was proposed by Manis et al. in 2017 \cite{Manis2017} as \textquotedblleft an entropy almost free of parameters", inspired by permutation entropy (Section \ref{sec30}) and rank-based entropy (Section \ref{sec32}). Bubble entropy relies on the Bubble Sort algorithm \cite{Manis2017}, which compares and swaps adjacent items until ordered; see Section 2.7 of \cite{Ribeiro2021} for an algorithmic description. 

\begin{description}
\item[{Applications}] 
\end{description}

\begin{itemize}

\item \textbf{Biomedical applications.} Due to its minimal dependency on parameters, bubble entropy is particularly useful in biomedical applications (e.g., analysis of heart rate variability) to distinguish healthy from pathological conditions \cite{Manis2017,Manis2021}.

\item \textbf{Fault bearing detection.} Bubble entropy is used to reinforce the accuracy of fault bearing diagnosis through the Gorilla Troops Optimization (GTO) algorithm for classification \cite{Gong2022}. A similar application can be found for the so-called Improved Hierarchical Refined Composite Multiscale Multichannel Bubble Entropy \cite{Gong2023}.

\item \textbf{Feature extraction.} Bubble entropy is compared with dispersion entropy (Section \ref{sec9}) in the extraction of single and double features in \cite{Jiang2023}.

\end{itemize}


\subsection{Categorical cross entropy}
\label{sec6}

According to \cite{Goodfellow2016,Bishop2006}, using cross entropy (Section \ref{sec7}) as an error (cost,
loss, objective,...) function speeds up the training of neural networks and
yields better results than the mean-squared error function for
classification problems. The cross entropy-based error function for
multi-class classification is called categorical cross entropy (CCE) \cite{Spindelbock2021}.
Specifically, if $t_{mk}$, $1\leq k\leq K$, are target binary variables for
input $m$, $1\leq m\leq M$, and $y_{mk}$ are the corresponding predicted
output variables, then the categorical cross entropy is given as
\begin{equation}
\textrm{CCE}=-\sum\limits_{m=1}^{M}\sum\limits_{k=1}^{K}t_{mk}\ln y_{mk}.  \label{CCE}
\end{equation}

For clarity, we consider CCE in this section and the conventional cross entropy in the next.

\begin{description}
\item[{Applications}] 
\end{description}

\begin{itemize}
  
\item \textbf{Deep Learning.} An improved form of CCE is used in deep neural networks to train them when dealing with noisy labels \cite{Li2021}.

\item \textbf{Multi-class classification.} CCE is used in \cite{Spindelbock2021} to train a convolutional neural network tailored to a multi-sensor, multi-channel time-series classification of cardiography signals. Likewise, CCE can be used for image classification. 

\item \textbf{Reinforcement learning.} CCE is used as an improvement of value function training (using classification instead of regression) mainly in games \cite{Farebrother2024}.
    
\item \textbf{Semi-supervised learning.} CCE is used in pseudo-labeling to optimize convolutional neural networks parameters \cite{Arazo2020}.

\end{itemize}


\subsection{Cross Entropy }
\label{sec7}

The cross entropy between the probability distributions $\mathrm{p}=(p_{1},...,p_{W})$ and $\mathrm{q}=(q_{1},...,q_{W})$ is defined as 
\begin{equation}
C(\mathrm{p,q})=-\sum_{i=1}^{W}p_{i}\log q_{i}.  \label{cross entropy}
\end{equation}%
It is related to the Shannon entropy $S(\mathrm{p})$ (equation \ref%
{Shannon}) and the Kullback-Leibler divergence 
$D\left(\mathrm{p}\right\Vert \mathrm{q})$ (Section \ref{sec28}, equation (\ref{KL Div})) through the equation.
\begin{equation}
C(\mathrm{p,q})=S(\mathrm{p})+D\left( \mathrm{p}\right\Vert 
\mathrm{q}).          \label{cross entropy2}
\end{equation}
Like $D\left( \mathrm{p}\right\Vert 
\mathrm{q})$, the cross entropy is used to quantify the difference between
two probability distributions $\mathrm{p}$ and $\mathrm{q}$. A typical example is determining whether an email is spam or not. In particular, when it comes to select parameters $\theta$ in a model that predicts a distribution $\mathrm{q}_{\theta}$ to fit an empirical distribution $\mathrm{p}$, one can minimize $C(\mathrm{p},\mathrm{q}_{\theta})$ instead of $D\left( \mathrm{p}\right\Vert 
\mathrm{q}_{\theta})$ because $S(\mathrm{p})$ does not depend on $\theta$.

\begin{description}
\item[{Applications}] 
\end{description}

\begin{itemize}

\item \textbf{Deep learning.} Cross entropy is a standard loss function for training deep neural networks, particularly those involving softmax activation functions. It is very useful for applications such as object detection, language translation, and sentiment analysis \cite{Mao2023}. In this regard, empirical evidence with limited and noisy data suggests that to measure the top-$\kappa$ error (a common measure of performance in machine learning performed with deep neural networks trained with the cross entropy loss), the loss function must be smooth, meaning that it should incorporate a smoothing parameter $\epsilon$ to handle small probability events \cite{Berrada2018}.

\item \textbf{Feature selection.} Cross entropy is used to select significant features of binary values from highly imbalanced large datasets via a framework called FMC Selector \cite{Wang2022}.
 
\item \textbf{Image analysis.} Wavelet analysis together with cross entropy are used in image segmentation, object recognition, texture analysis (e.g., fabric defect detection) and pattern classification \cite{Kim2007}.

\item \textbf{Learning-to-rank methods.} In \cite{Bruch2019} the author proposes a learning-to-rank loss function that is based on cross entropy. Learning-to-rank methods form a class of ranking algorithms that are widely applied in information retrieval.

\item \textbf{Multiclass classification.} Cross entropy is used to enhance the efficiency of solving support vector machines for multi-class classification problems \cite{Santosa2015}.

\item \textbf{Semi-supervised clustering.} Cross entropy is employed along with the information bottleneck method in semi-supervised clustering. It is robust to noisy labels and automatically determines the optimal number of clusters under mild conditions \cite{Smieja2017}.

\end{itemize}


\subsection{Differential entropy}
\label{sec8}
  
As said in Section \ref{secSKaxioms}, differential entropy is the continuous counterpart of Shannon entropy: If $X$ is a continuous random variable with density $\rho (x)$ and support
set $S$, then the differential entropy of $X$ is \cite{Cover2006}
\begin{equation}
h(X)=h(\rho )=-\int\nolimits_{S}\rho (x)\log \rho (x)dx. \label{diff entropy}
\end{equation}
Therefore, differential entropy is used in the analysis of continuous random variables and processes, e.g., analog signals.

\begin{description}
\item[{Applications}] 
\end{description}

\begin{itemize}

 \item \textbf{Anomaly detection.} Differential entropy can measure changes in the probability density function of an analog signal which reveals an anomaly in the source, whether it is a mechanical system or a patient \cite{Orchard2012}.

 \item \textbf{Emotion recognition.} Differential entropy has been used in \cite{patel2021} to extract features in EEG-based human emotion recognition.

 \item \textbf{Feature selection.} A feature selection algorithm based on differential entropy to evaluate feature subsets has been proposed in \cite{Bishop2006}. This algorithm effectively represents uncertainty in the boundary region of a fuzzy rough model and demonstrates improved performance in selecting optimal feature subsets, thereby enhancing classification accuracy; see \cite{qu2020} for an implementation.

 \item \textbf{Gaussianity of a population.} The differential entropy $h(X)$ reaches its maximun among all random variables with the same variance $\sigma$, $h_{\max }(X)=\log (2\pi e\sigma ^{2})/2$, when $X$ is normal \cite{Cover2006}. Therefore, a sample estimate of $h(X)$ that is significantly lower than that upper bound indicates that the underlying population is not Gaussian.

 \item \textbf{Generative models.} Variational Autoencoders and other generative models leverage differential entropy to model the latent space of continuous data distributions. These models can learn better representations of the input data, thus improving the performance \cite{e23070856}.

 \item \textbf{Mutual information.} Differential entropy is instrumental to compute the mutual information of continuous-valued random variables and processes, e.g., autoregressive processes. Seen in speech processing (linear prediction), seismic signal processing and biological signal processing \cite{e20100750}.

 \item \textbf{Probabilistic models.} Differential entropy is utilized in probabilistic models such as Gaussian Mixture Models to describe the uncertainty and distribution of continuous variables. This approach is applicable to image processing and network inference as well \cite{Robin2023}.

\end{itemize}


\subsection{Dispersion Entropy}
\label{sec9}

Dispersion entropy was introduced by Rostaghi and Azami in 2016 \cite{7434608} to solve the shortcoming of permutation entropy (Section \ref{sec30}) of only taking into account the ranking of the
data amplitudes but not their values. This explains why the definition of dispersion entropy resembles that of permutation entropy. 

Indeed, to compute the dispersion entropy of a
time series, the series is first transformed into a string of symbols called \textquotedblleft dispersion patterns\textquotedblright\ via a number of linear or nonlinear mapping techniques, then the probabilities of the different dispersion patterns are
estimated by their relative frequencies and, finally, the Shannon entropy of
the resulting probability distribution is calculated. The dispersion entropy has three parameters: (i) $c$, the number of letters, running from 1 to $c$, that build the dispersion patterns; $m$, the length of the data blocks (or \textquotedblleft embedding vectors\textquotedblright\ ) used to obtain dispersion patterns of the same length, so that there are $c^{m}$ possible dispersion patterns of length $m$; and $d$, the time delay. See \cite{e20030210} for details.

\begin{description}
\item[{Applications}] 
\end{description}

\begin{itemize}
  
\item \textbf{Feature extraction.} Multiscale fuzzy dispersion entropy is applied in fault diagnosis of rotating machinery to capture the dynamical variability of time series across various scales of complexity \cite{e25111494}.

\item \textbf{Image classification.} A multiscale version of dispersion entropy called MDispEn\textsubscript{2D} has been used with biomedical data to measure the impact of key parameters that may greatly influence the entropy values obtained in image classification \cite{e23101303}.
 
\item \textbf{Signal classification.} Other generalization of dispersion entropy, namely, fractional fuzzy dispersion entropy, has been proposed as a fuzzy membership function for signal classification tasks \cite{Hu2024}.

\item \textbf{Signal denoising.} Dispersion entropy is also used in signal denoising via (i) adaptive techniques and group-sparse total variation \cite{e23121567}, or (ii) empirical mode decomposition with adaptive noise \cite{electronics8060597}.

\item \textbf{Time series analysis.} Multiscale graph-based dispersion entropy is a generalization of dispersion entropy used to analyze multivariate time series data in graph and complex network frameworks, e.g., weather and two-phase flow data; it combines temporal dynamics with topological relationships \cite{arXiv:2405.00518}.

\end{itemize}


\subsection{Energy Entropy and Empirical Mode Decomposition Energy Entropy }
\label{sec11}
   
Energy entropy is the Boltzmann-Gibbs-Shannon entropy of a normalized distribution of energy levels or modes. So it is mainly used in data-driven analysis of physical systems, whether in natural sciences or technology.

In nonlinear time series analysis, one uses empirical mode decomposition (EMD) to decompose a time series into a set of intrinsic mode functions, each representing a simple oscillatory mode inherent to the data. This decomposition is adaptive and data-driven, making it suitable for analysing complex signals without requiring predefined basis functions \cite{sym10110623}. As its name suggests, EMD energy entropy combines energy entropy with EMD to provide information about the energy distribution across different intrinsic mode functions derived from a signal.

\begin{description}
\item[{Applications}] 
\end{description}

\begin{itemize}

\item \textbf{Chatter detection.} This application involves detecting vibrations and noise in machining operations that can indicate chattering. In this regard, energy entropy can detect chatter in robotic milling \cite{LIU2018169}.

\item \textbf{Fault prediction in vibration signal analysis.} EMD energy entropy has been employed to predict early fault of bearings in rotating machinery \cite{GAO2022110417,YU2006269}.

\item \textbf{Feature extraction.} Energy entropy is calculated via the empirical decomposition of the signal into intrinsic mode functions and serves as a feature for machine learning models used in chatter detection. The chatter feature extraction method draws on the largest energy entropy \cite{doi:10.1142/S0219519423400638}.
   
\item \textbf{Time-series forecasting.} EMD energy entropy was used in  \cite{zhu2021short} to predict short-term electricity consumption by taking into account the data variability, i.e., that power consumption data is non-stationary, nonlinear, and influenced by the season, holidays, and other factors. In \cite{gao2019analysis}, this entropy was the tool to distinguish two kinds of financial markets.

\end{itemize}


\subsection{Entanglement Entropy } 
\label{sec12}
   
Entanglement entropy originated in Quantum Mechanics as a measure of the degree of quantum entanglement between two subsystems of a quantum system. To be more precise, entanglement entropy is the von Neumann entropy (Section \ref{sec41}) of the reduced density matrix for any of the subsystems \cite{headrick2019lectures}; see \cite{rieger2024sample} for the estimation of entanglement entropy through supervised learning. In addition to its important role in quantum mechanics and quantum information theory, entanglement entropy is increasingly finding more applications in machine learning as well. This is why we have included entanglement entropy in this review.

\begin{description}
\item[{Applications}] 
\end{description}

\begin{itemize}
\item \textbf{Feature extraction.} In quantum machine learning, entanglement entropy is used for feature extraction by representing data in a form that highlights quantum correlations and thus, leveraging the quantum properties of the data \cite{liu2021entanglement}.

\item \textbf{Neural networks in quantum models.} Entanglement entropy is used in quantum models to quantify unknown entanglement by using neural networks to predict entanglement measures of unknown quantum states based on experimentally measurable data: moments or correlation data produced by local measurements \cite{lin2022quantifying}.

\end{itemize}

   
\subsection{Excess Entropy}
\label{sec13}

The \textit{mutual information} between the random variables $X$
and $Y$ measures the average reduction in uncertainly about one of
variables that results from learning the value of the other. It is defined as \cite{Cover2006}
\begin{equation}
I(X;Y)=I(Y;X)=S(X)+S(Y)-S(X,Y)\geq 0,  \label{mutual info}
\end{equation}%
where $S(...)$ is Shannon's entropy (\ref{Shannon}).

\textit{Excess entropy} (also called dual total correlation) is a non-negative
generalization of mutual information to more than two random variables
defined as \cite{ABDALLAH2012275}

\begin{equation}
D(X_{1},...,X_{N})=S(X_{1},...,X_{N})-\sum\limits_{i=1}^{N}S\left(
X_{i}\right\vert X_{1},...,X_{i-1},X_{i+1},...,X_{N})  \label{excess ent}
\end{equation}%
where $S\left( X_{i}\right\vert ...)$ is the Shannon conditional entropy of $X_{i}$ given the other variables. Of course, $D(X_{1},X_{2})=I(X_{1};X_{2})$. An alternative definition called \textit{total excess entropy} was introduced by Crutchfield and Packard in \cite{crutchfield1983}. 

\begin{description}
\item[{Applications}] 
\end{description}

\begin{itemize}

\item \textbf{Image segmentation.} Excess entropy is used to measure the structural information of a 2D or 3D image and then determine the optimal threshold in a segmentation algorithm proposed in \cite{bardera2009}. The working hypothesis of this thresholding-based segmentation algorithm is that the optimal threshold corresponds to the maximum excess entropy (i.e., to a segmentation with maximum structure).  

\item \textbf{Machine learning.} In \cite{nir2020machine}, the authors present a method called machine-learning iterative calculation of entropy, for calculating the entropy of physical systems by iteratively dividing the system into smaller subsystems and estimating the mutual information between each pair of halves.

\item \textbf{Neural estimation in adversarial generative models.} Mutual Information Neural Estimator is a scalable estimator used in high dimensional continuous data analysis that optimizes mutual information. The authors apply this estimator to Generative Adversarial Networks (GANs) \cite{belghazi2021minemutualinformationneural}.

\item \textbf{Time series classification.} In this application, total excess entropy is used for classifying stationary time series into long-term and short-term memory. A stationary sequence with finite block entropy is long-term memory if its excess entropy is infinite \cite{math11112448}.

\end{itemize}

  
\subsection{Fluctuation-based Dispersion Entropy }
\label{sec14}
   
Fluctuation-based dispersion entropy (FDispEn) was introduced by Azami and Escudero in 2018 \cite{e20030210} to account for the variability of time-series by incorporating information about fluctuations into the definition of dispersion entropy (Section \ref{sec6}).

Specifically, FDispEn considers the differences between adjacent symbols (belonging to the alphabet $\{1,2,...,c\}$) of the dispersion patterns, termed fluctuation-based dispersion patterns. Therefore, if the dispersion patterns are vectors of length $m$, the fluctuation-based ones have length $m-1$ and components that range from $-c+1$ to $c-1$. As a result, there are $(2c-1)^{m-1}$ potential fluctuation-based dispersion patterns. The number of possible patterns for the same parameters is the only difference between the algorithms for the conventional dispersion entropy and FDispEn.

\begin{description}
\item[{Applications}] 
\end{description}

\begin{itemize}

\item \textbf{Fault diagnosis.} The so-called refined composite moving average FDispEn is used in machinery fault diagnosis by analysing vibration signals \cite{Chen2023}. 

Refined composite multiscale FDispEn and supervised manifold mapping are used in fault diagnosis for feature extraction in planetary gearboxes \cite{machines11010047}.

Multivariate hierarchical multiscale FDispEn along with multi-cluster feature selection and Gray-Wolf Optimization-based Kernel Extreme Learning Machine helps diagnose faults in rotating machinery. It also captures the high-dimensional fault features hidden in multichannel vibration signals \cite{Zhou2021}.

\item \textbf{Feature extraction.} FDispEn gives rise to hierarchical refined multi-scale fluctuation-based dispersion entropy, used to extract underwater target features in marine environments and weak target echo signals, thereby improving the detection performance of active sonars \cite{Li20231}.

\item \textbf{Robustness in spectrum sensing.} Reference \cite{e23121611} proposes a machine learning implementation of spectrum sensing using an improved version of the FDispEnt as a feature vector. This improved version shows enhanced robustness to noise. 
     
\item \textbf{Signal classification.} FDispEn helps distinguish various physiological states of biomedical time series and it is commonly used in biomedicine. It is also used to estimate the dynamical variability of the fluctuations of signals applied to neurological diseases \cite{azami2019multiscalefluctuationbaseddispersionentropy}. Fluctuation-based reverse dispersion entropy is applied to signal classification combined with $k$-nearest neighbor \cite{Jiao2021}.

\item \textbf{Time series analysis.} FDispEn is used to quantify the uncertainty of time series to account for knowledge on parameters sensitivity and studying the effects of linear and nonlinear mapping on the defined entropy in \cite{e20030210}. FDispEn is defined as a measure for dealing with fluctuations in time series. Then, the performance is compared to complexity measures such as permutation entropy (Section \ref{sec30}), sample entropy (Section \ref{sec33}), and Lempel-Ziv complexity \cite{Cover2006,Amigo2004}.

\end{itemize}


\subsection{Fourier Entropy }
\label{sec15}
   
The Fourier entropy $h(f)$ of a Boolean function $f:\{-1,1\}^{n}\rightarrow \{-1,1\}$ is the Shannon entropy of its power spectrum $\{\hat{f}(S)^{2}:S\subset \lbrack n]\}$, where $S\subset \lbrack n]$
stands for the $2^{n}$ subsets of $\{1,2,...,n\}$ (including the empty set) and $\hat{f}$ is the Fourier transform of $f$. By Parseval's Theorem, $%
\sum_{S\subset \lbrack n]}\hat{f}(S)^{2}=1$, so the power spectrum of $f$ is a probability distribution. 

\begin{description}
\item[{Applications}] 
\end{description}

\begin{itemize}
  
\item \textbf{Decision trees.} The Fourier Entropy-Influence Conjecture, made by Friedgut and Kalai \cite{everymonotone}, says that the Fourier entropy of any Boolean function $f$ is upper bounded, up to a constant factor, by the total influence (or average sensitivity) of $f$. This conjecture, applied to decision trees, gives interesting results that boil down to $h(f)=O(\text{log}(L(f)))$, meaning that $\left\vert h(f)\right\vert <C\log (L(f))$, where $L(f)$ denotes the minimum number of leaves in a decision tree that computes \textit{f} and $C>0$ is independent of $f$ and $L(f)$ \cite{CHAKRABORTY201692}. Another similar application to decision trees can be found in \cite{10.1007/978-3-642-22006-7_28}.

\item \textbf{Learning theory.} The Fourier Entropy-Influence Conjecture is closely related to the problem of learning functions in the membership model. It is said that if a function has small Fourier entropy it means that its Fourier transform is concentrated on a few characters, i.e. the function can be approximated by a sparse polynomial, a class that is very important in the context of learning theory \cite{9317968}. Learning theory provides the mathematical foundation for understanding how algorithms learn from data, guiding the development of machine learning models.

\end{itemize}

     
\subsection{Fractional Fourier Entropy }
\label{sec16}

Fractional Fourier entropy is calculated in two steps: first, take the fractional Fourier transform of the data \cite{330368}; second, compute the Shannon entropy of the probability distribution defined by the normalized frequency spectrum.

\begin{description}
\item[{Applications}] 
\end{description}

\begin{itemize}

\item \textbf{Anomaly detection in remote sensing.} Fractional Fourier entropy is used in hyperspectral remote sensing to distinguish signals from background and noise \cite{8847346}, \cite{rs14030797}.

\item \textbf{Artificial intelligence.} Two-dimensional fractional Fourier entropy helps to diagnose COVID-19 by extracting features from chest CT images \cite{10.1145/3451357}.

\item \textbf{Biomedical image classification.} Fractional Fourier entropy has proven helpful in detecting pathological brain conditions. By using it as a new feature in Magnetic Resonance Imaging, the classification of images is improved in time and cost \cite{e17127877}.

\item \textbf{Deep learning.} Fractional Fourier entropy is used in the detection of gingivitis via feed-forward neural networks. It reduces the complexity of image extraction before classification and can obtain better image eigenvalues \cite{YAN202036}.

\item \textbf{Emotion recognition.} Fractional Fourier entropy, along with two binary support vector machines, helps improve the accuracy of emotion recognition from physiological signals in electrocardiogram and galvanic skin responses \cite{PANAHI2021102863}.

\item \textbf{Multilabel classification.} Fractional Fourier entropy has been used in a tea-category identification system, which can automatically determine tea category from images captured by a 3 charge-coupled device digital camera \cite{e18030077}.

\end{itemize}


\subsection{Fuzzy Entropy}
\label{sec17}

Fuzzy entropy quantifies "fuzziness" (uncertainty due to imprecision) in fuzzy set theory \cite{Zadeh1965} and time series analysis. There are different definitions of fuzzy entropy in the literature that draws on
the concept of fuzzy sets \cite{DeLuca1972,ISHIKAWA1979113}. In 2007, Chen et al. \cite{ChenYu2007} introduced a
concept that is akin to approximate and sample entropies (Sections \ref{sec2} and \ref{sec33}).
In this version, the Heaviside function $\Theta (r-\left\Vert
x_{i}-x_{j}\right\Vert )$ used in the latter entropies to define
neighborhoods is replaced by the exponential function $\exp (-\delta
_{ij}^{m}/r)$, where $r>0$ is the tolerance and $\delta _{ij}^{m}$
is a distance between the embedding vectors $(x_{i},...,x_{i+m-1})$ and $%
(x_{j},...,x_{+m-1})$ of length $m$. This distance includes also the
subtraction of the mean value of the vector components to minimize the
effects of non-stationarity.

In 2014, Zheng proposed multiscale fuzzy entropy \cite{ZHENG2014187}. Fuzzy entropy was introduced into fuzzy dynamical systems in \cite{e18010019}.

\begin{description}
\item[{Applications}] 
\end{description}

\begin{itemize}

\item \textbf{Clustering and time-series analysis.} Fuzzy entropy is used in problems of robustness against outliers in clustering techniques in \cite{Durso2023}.

\item \textbf{Data analysis.} Fuzzy entropy is proposed in \cite{e20060424} to assess the strength of fuzzy rules with respect to a dataset, based on the greatest energy and smallest entropy of a fuzzy relation.

\item \textbf{Fault detection.} Fuzzy entropy (along with dispersion entropy, Section \ref{sec9}) was the best performer in a comparative study of entropy-based methods for detecting motor faults \cite{MotorFault2023}. Multiscale fuzzy entropy is used to measure complexity in time series in rolling bearing fault diagnosis \cite{Jind2014MultiscaleFE}.

\item \textbf{Feature selection and mathematical modelling.} Fuzzy entropy is used in feature selection to evaluate the relevance and contribution of each feature in Picture Fuzzy Sets \cite{KUMAR2023100351}.

\item \textbf{Image classification.} Fuzzy entropy, in the form of multivariate multiscale fuzzy entropy, is proposed and tested in \cite{e24111577} for the study of texture in color images and their classification.

\item \textbf{Image segmentation.} Fuzzy entropy is the objective function of a colour image segmentation technique based on an improved cuckoo search algorithm \cite{Tan2021}.

\end{itemize}


\subsection{Graph Entropy }
\label{sec19}
   
Graph entropy was introduced by Korner in 1971 \cite{korner1971} to quantify the complexity or information content of a graph. It is usually defined as the Shannon entropy (although any other entropy would do) of a probability distribution over the graph's vertex set. In addition to applications in data analysis and machine learning, graph entropy is also applied in combinatorics; see \cite{simonyi1995} for a survey of graph entropy.

A particular case of graph entropy is the horizontal visibility (HV) graph entropy of a time series, which is the graph entropy of the so-called HV graph of the time series \cite{luque2010horizontalvisibilitygraphsexact}. In particular, this method is useful for distinguishing between different types of dynamical behaviours in nonlinear time series, such as chaotic versus regular dynamics \cite{LACASA201835}.

\begin{description}
\item[{Applications}] 
\end{description}

\begin{itemize}

\item \textbf{Dimension reduction and feature selection.} Graph entropy gives rise to the Conditional Graph Entropy that helps in the alternating minimization problem \cite{harangi2023conditionalgraphentropyalternating}.

\item \textbf{Graph structure.} Graph entropy is used to measure the information content of graphs, as well as to evaluate the complexity of the hierarchical structure of a graph \cite{wu2022structuralentropyguidedgraph}.

\item \textbf{Graph-based time series analysis.} Graph entropy can be used in time series analysis in conjunction with any method that transforms time series into graphs. An example is the HV graph entropy presented above; see \cite{juhnkekubitzke2021countinghorizontalvisibilitygraphs} and references therein.

\item \textbf{Node embedding dimension selection.} Graph entropy is applied in Graph Neural Networks through the Minimum Graph Entropy algorithm. It calculates the ideal node embedding dimension of any graph \cite{luo2021graphentropyguidednode}.

\item \textbf{Time series analysis.} HV graph entropy along with sample entropy has been used in \cite{Zhua2014a} to identify abnormalities in the EEGs of alcoholic subjects. HV transfer entropy was proposed in \cite{YU2017249} to estimate the direction of the information flow between pairs of coupled time series.

\end{itemize}


\subsection{Havrda–Charvát Entropy }
\label{sec21}
   
Havrda–Charvát (HC) entropy, also known as the Havrda–Charvát $\alpha $-entropy, was introduced by Havrda and Charvát \cite{HCentropy} in 1967 in Information Theory. It is formally identical to the Tsallis entropy (Section \ref{sec40}, introduced by Tsallis in 1988 in Statistical Mechanics \cite{Tsallis1988}. Similar to Rényi entropy (Section \ref{sec1}), HC entropy is a family of entropies parameterized by $\alpha>0$ and generalizes the Shannon entropy in the sense that the HC entropy coincides with the Shannon entropy in the limit $\alpha \rightarrow 1$. 

Although nowadays the most popular name for this entropy is Tsallis entropy, we present in this section applications published in articles that refer to this entropy as the Havrda–Charvát entropy.

\begin{description}
\item[{Applications}] 
\end{description}

\begin{itemize}

\item \textbf{Computer vision.} An HC entropy-based technique for group-wise registration of point sets with unknown correspondence is used in graphics, medical imaging and pattern recognition. By defining the HC entropy for cumulative distribution functions (CDFs), the corresponding CDF-HC divergence quantifies the dissimilarity between CDFs estimated from each point-set in the given population of point sets \cite{Chen2010}.

\item \textbf{Financial time series analysis.} Weighted HC entropy outperforms regular HC entropy when used as a complexity measure in financial time series. The weights turn out to be useful for showing amplitude differences between series with the same order mode (i.e. similarities in patterns or specific states) and robust to noise \cite{SHI2021125914}.

\item \textbf{Image segmentation and classification.} HC entropy is applied as loss function in image segmentation and classification tasks using convolutional neural networks in \cite{brochet2021deeplearningusinghavrdacharvat}.

\item \textbf{Loss functions in deep learning.} HC entropy can be used to design loss functions in deep learning models. These loss functions are particularly useful in scenarios with small datasets, common in medical applications \cite{e24040436}.

\end{itemize}

   
\subsection{Intrinsic Mode Entropy }
\label{sec23}

Intrinsic mode entropy (IME)  was introduced by Amoud et al. in 2007 \cite{4154718} to study time series in different scales, in the presence of dominant local trends and low-frequency components. It is obtained by computing the sample entropy (Section \ref{sec33}) of the cumulative sums of the intrinsic mode functions extracted by the empirical mode decomposition method (Section \ref{sec11}).

\begin{description}
\item[{Applications}] 
\end{description}

\begin{itemize}

\item \textbf{Language gesture recognition.} IME is used in \cite{4650350} to analyse data from a 3-dimensio\-nal accelerometer and a five-channel surface electromyogram of the user’s dominant forearm for automated recognition of Greek sign language gestures.

\item \textbf{Neural data analysis.} An IME version with improved discriminatory capacity in the analysis of neural data is proposed in \cite{Hu2011}.

\item \textbf{Time series analysis.} IME is used in nonlinear time series analysis to efficiently characterize the underlying dynamics \cite{4154718}. As any multiscale entropy, IME is particularly useful for the analysis of physiological time series \cite{Hu2017}. See also \cite{Amoud2009} for an application to the analysis of postural steadiness.

\end{itemize}

   
\subsection{Kaniadakis Entropy }
\label{sec25}
   
Kaniadakis entropy, also known as $\kappa$-entropy due to its dependence on a parameter $0<\kappa<1$, was introduced by the physicist G. Kaniadakis in 2002 \cite{PhysRevE.66.056125} to address the limitations of classical entropy in systems exhibiting relativistic effects. It is defined as
\begin{equation}
S_{\kappa}(p_{1},...,p_{W})=\sum_{i=1}^{W}\frac{p_{i}^{1-\kappa
}-p_{i}^{1+\kappa }}{2\kappa }.   \label{Kaniadakis}
\end{equation}%
Kaniadakis entropy is a relativistic generalization of the BGS entropy (\ref{G_ent}) in the sense that the latter is recovered in the $\kappa \to 0$ limit \cite{e26050406}. Like other physical entropies such as Tsallis' and von Neumann's, Kaniadakis entropy has also found interesting applications in applied mathematics.

\begin{description}
\item[{Applications}] 
\end{description}

\begin{itemize}

\item \textbf{Image segmentation.} Kaniadakis entropy is used in image thresholding to segment images with long-tailed distribution histograms; the parameter $\kappa$ is selected via a swarm optimization search algorithm \cite{lei2020adaptive}.

\item \textbf{Images threshold selection.} Kaniadakis entropy can be used to construct an objective function for image thresholding. By using the energy curve and the Black Widow optimization algorithm with Gaussian mutation, this approach can be performed on both grayscale and colour images of different modalities and dimensions \cite{10201718}.

\item \textbf{Seismic imaging.} Application of the Maximum Entropy Principle with $S_{\kappa}$ leads to the Kaniadakis distribution, a deformation of the Gaussian distribution that has application, e.g., in seismic imaging \cite{e25070990}.

\end{itemize}


\subsection{Kernel Entropy}
\label{sec26}

Kernel (or kernel-based) entropy is an evolution of the approximate entropy (Section \ref{sec2}) that consists of replacing the Heaviside step function in the definition of proximity by other functions (or \textquotedblleft kernels\textquotedblright ) to give more weight to the nearest neighbors. Thus, in the Gaussian kernel entropy, the Gaussian kernel 
\begin{equation}
\ker (i,j;r)=\exp \left( -\frac{\left\Vert x_{i}-x_{j}\right\Vert ^{2}}{%
10r^{2}}\right)   \label{Gaussian kernel}
\end{equation}
is used. Here $x_{i}$, $x_{j}$ are entries of a time series and $r$ is the parameter of the approximate entropy. Other popular kernels include the spherical, Laplacian and Cauchy functions \cite{Mekyska_2015}. 

Of course, the same refinement can be done in the computation of the sample entropy (Section \ref{sec33}), an improvement of the approximate entropy. To distinguish between the two resulting kernel entropies, one speaks of kernel-based approximate or sample entropy.

\begin{description}
\item[{Applications}] 
\end{description}

\begin{itemize}

\item \textbf{Complexity of time series.} The authors of \cite{1527935} present experimental evidence that Gaussian kernel entropy outperforms approximate entropy when it comes to analyze the complexity of time series.

\item \textbf{Fetal heart rate discrimination.} In \cite{ZaylaaProceed2016} the authors compare the performance of several kernel entropies on fetal heart rates discrimination, with the result that the circular and Cauchy kernels outperform other, more popular kernels, such as the Gaussian or the spherical ones.

\item \textbf{Pathological speech signal analysis.} Reference \cite{Mekyska_2015} is a study of several approaches in the field of pathological speech signal analysis. Among the new pathological voice measures, the authors include different kernel-based approximate and sample entropies.

\item \textbf{Speech signal classification in Parkinson’s disease.} Gaussian kernel entropy, along with other nonlinear features, is used in \cite{OrozcoProceed2013} in the task of automatic classification of speech signals from subjects with Parkinson's disease and a control set.

\end{itemize}


\subsection{Kolmogorov-Sinai Entropy }
\label{sec27}
   
As mentioned in Section \ref{secSKaxioms}, Kolmogorov-Sinai (KS) entropy is a classical entropy introduced by Kolmogorov in ergodic theory in 1958 \cite{kolmogorov1986}, and extended by Sinai to the theory of measure-preserving dynamical systems in 1959 \cite{sinai1959}. KS entropy is a fundamental invariant in the theory of metric dynamical systems \cite{walters2000}.

The calculation of the KS entropy of a dynamical system requires in general
three steps: (i) coarse-graining of the state space by a finite partition $\alpha $, (ii) computation of the Shannon entropy rate of the resulting symbolic dynamics $\mathbf{X}^{\alpha }$ (a finite-state, stationary random process with so many states as the cardinality of $\alpha $ \cite{Amigo2022}), called the Shannon entropy with respect to $\alpha $, $S(\mathbf{X}^{\alpha })$, and (iii) taking the supremum of $S(\mathbf{X}^{\alpha })$ over all finite partitions $\alpha $, the latter being equivalent to taking the limit of $S(\mathbf{X}^{\alpha })$ for ever finer partitions \cite{walters2000}.

Apart from approximating the limits in steps (ii) and (iii), other popular estimators of the KS entropy resort to Pesin's formula \cite{YaBPesin_1977}, which involves the strictly positive Lyapunov exponents of the system; see also \cite{SHIOZAWA2024129531} for high dimensional complex systems and \cite{Parlitz2016} for the estimation of the Lyapunov exponents.

\begin{description}
\item[{Applications}] 
\end{description}

\begin{itemize}

\item \textbf{Time series analysis.} The perhaps main practical application of the Kolmogorov-Sinai entropy is the analysis of nonlinear, real-valued time series, where it is used to characterize the underlying dynamical system, in particular, its chaotic behavior. Recent practical examples include short-term heart rate variability \cite{e22121396}, physical models of the vocal membranes \cite{SHIOZAWA2024129531}, autonomous driving \cite{app14062261}, and EEG-based human emotion recognition \cite{AFTANAS199713,patel2021}.

\end{itemize}


\subsection{Permutation Entropy}
\label{sec30}
    
The conventional (or Shannon) permutation entropy of a time series was introduced by Bandt and Pompe in 2002 \cite{Bandt2002}. It is the Shannon entropy (Section \ref{sec34}) of the probability distribution obtained from the ordinal patterns of length $L \ge 2$ in the time series, i.e., the rankings (or permutations) of the series entries in sliding windows of size $L$. Therefore, permutation entropy depends on the parameter $L$. Furthermore, permutation entropy is easy to program, relatively robust to noise, and can be computed practically in real time since knowledge of the data range is not needed \cite{e2013practical}. See, e.g., \cite{amigo2010permutation,e14081553,e2013permutationEditorial,e2023permutationEditorial} for references on theoretical and practical aspects of permutation entropy.

If, instead of the Shannon entropy of the ordinal patterns distribution, we use other entropy, e.g., Rényi entropy (Section \ref{sec1}) or Tsallis entropy (Section \ref{sec40}), then we obtain the corresponding "permutational version": permutation Rényi entropy, permutation Tsallis entropy, and more \cite{e17074627,ZUNINO20086057}. There are also "weighted versions" that take into account not only the rank order of the entries in a window but also their amplitudes; see, e.g., \cite{Stosic_2022}. In turn, "multiscale versions" (including multiscale permutation Rényi and Tsallis entropy) account for multiple time scales in time series by using different time delays \cite{Yin2018,li2019multiscale,AZAMI201628}.
  
\begin{description}
\item[{Applications}] 
\end{description}

\begin{itemize}

\item \textbf{Analysis and classification of EEGs.} One of the first applications of permutation entropy was the analysis of EEGs of subjects with epilepsy because normal and abnormal signals (during epileptic seizures) have different complexities \cite{doi101142S02181274030}. Furthermore, since permutation entropy can be computed in virtually real time, it has been used to predict seizures in epilepsy patients by tracking dynamical changes in EEGs \cite{e14081553}. Further examples can be found in the reviews \cite{Amigo2022}. Results can be improved using permutation Rényi and Tsallis entropy due to their additional, fine-tunable parameter \cite{e19030134,e17074627}.

\item \textbf{Analysis of unstructured data.} Nearest-neighbor permutation entropy is an innovative extension of permutation entropy tailored for unstructured data, irrespective of their spatial or temporal configuration and dimensionality, including, e.g., liquid crystal textures \cite{10106350209206}.

\item \textbf{Classification for obstructive sleep apnea.} A combination of permutation entropy-based indices and other entropic metrics was used in \cite{Graff2023} to distinguish subjets with obstructive sleep apnea from a control group. The data consisted of heart rate and beat-to-beat blood pressure recordings.

\item \textbf{Determinism detection.} Time series generated by one-dimensional maps have necessarily forbidden ordinal patterns of all sufficiently large lengths $L$ \cite{amigo2010permutation}. Theoretical results under some provisos and numerical results in other cases show that the same happens with higher dimensional maps \cite{amigo2008forbidden,e2013permutation}. Therefore, the scaling of permutation entropy with $L$ can distinguish noisy deterministic signals from random signals \cite{amigo2010permutation,CARPI20102020}.

\item \textbf{Emotion recognition.} Permutation entropy is used to help in tasks of feature extraction in EEGs \cite{patel2021}.

\item \textbf{Estimation of the Kolmogorov-Sinai (KS) entropy.} Given a piecewise monotone map $f$ of a one-dimensional interval, the permutation entropy rate converges to the KS entropy of $f$ (Section \ref{sec27}) when $L \rightarrow \infty$  \cite{entropyofintervalmaps}. Therefore, the permutation entropy calculated with a sufficiently large $L$ (and divided by $L-1$) is a good estimator of the KS entropy of one-dimensional dynamics. This result is remarkable because the computation of the KS entropy requires in general two infinite limits (see Section \ref{sec27}), while permutation entropy requires only one.   

\item \textbf{Nonlinear time series analysis.} Permutation entropy has been extensively used in the analysis of continuous-valued time series for its straightforward discretization of the data and ease of calculation. Numerous applications can be found, e.g., in \cite{e14081553,e2013permutationEditorial,e2023permutationEditorial} and the references therein.

\item \textbf{Speech signals analysis.} In their seminal paper \cite{Bandt2002}, Bandt and Pompe used precisely permutation entropy to analyze speech signals and showed that it is robust with respect to the window length, sampling frequency and observational noise.

\item \textbf{The causality-complexity plane.} Permutation entropy together with the so-called statistical complexity builds the causality-complexity plane, that has proven to be a powerful tool to discriminate and classify time series \cite{PhysRevLett99154102}. By using variants of the permutation entropy and the statistical complexity, the corresponding variants of the causality-complexity plane are obtained, possibly with enhanced discriminatory abilities for the data at hand \cite{e2023permutationEditorial}.
  
\item \textbf{Wind power prediction.} Permutation entropy has been used along with variational modal decomposition to predict wind power \cite{qu2023wind}.

\end{itemize}


\subsection{Rank-based entropy }
\label{sec32}
    
Rank-based entropy (RbE) was introduced by Citi et al. in 2014 \cite{CitiGuffantiMainardi2014} in the framework of multiscale entropy, where traditionally sample entropy is used. RbE measures the unpredictability of a time series quantifying
the \textquotedblleft amount of shuffling\textquotedblright\ that the ranks
of the mutual distances between pairs of $m$-long embedding vectors $%
(x_{i},x_{i+1},...,x_{i+m-1})$ and $(x_{j},x_{j+1},...,x_{j+m-1})$ undergo
when considering the next observation, i.e., the corresponding
$(m+1)$-long embedding vectors. See, e.g., Section 2.7 of \cite{Ribeiro2021} for an algorithmic description.

\begin{description}
\item[{Applications}] 
\end{description}

\begin{itemize}

\item \textbf{Anomaly detection.} RbE is applied in mixed data analysis to check the influence of categorical features, using Jaccard index for anomaly ranking and classification \cite{GarcheryGranitzer2018}.

\item \textbf{Feature selection.} RbE is used in the Entropy-and-Rank-based-Correlation framework to select features, e.g., in the detection of fruit diseases \cite{KhanAkramSharif2021}.

\item \textbf{Mutual information.} RbE is used to rank mutual information in decision trees for monotonic classification \cite{HuCheZhangEtAl2012}.

\item \textbf{Node importance.} RbE is employed in the analysis of graphs to rank nodes taking into account the local and global structure of the information \cite{LiuGao2023}.

\item \textbf{QSAR models.} RbE is employed in Quantitative Structure-Activity Relationship models (QSAR) to analyse their stability via “rank order entropy”, suggesting that certain models typically used should be discarded \cite{McLellanRyanBreneman2011}.

\item \textbf{Time series analysis.} RbE is used in terms of correlation entropy to test serial independence in \cite{DiksPanchenko2008}.
A multiscale version was used in \cite{CitiGuffantiMainardi2014} to study data of heart rate variability.

\item \textbf{Time series classification.} RbE helps classify order of earliness in time series to generate probability distributions in different stages \cite{SunLiSongHong2023}.

\end{itemize}
   

\subsection{Rao's Quadratic Entropy}
\label{sec31}

Rao's quadratic entropy (RQE, not to be confused with Rényi's collision entropy $R_{2}$, Section \ref{sec1}, sometimes called quadratic entropy, too) was proposed in 1982 \cite{rao1982} as a measure of diversity in biological populations. Given $W$ species, RQE is defined as
\begin{equation}
\mathrm{RQE}(p_{1},...,p_{W})=\sum\limits_{i,j=1}^{W}\delta _{i,j}p_{i}p_{j}
\label{Rao ent}
\end{equation}%
where $\delta _{i,j}$ is the difference between the $i$-th and the $j$-th
specie and $\{p_{1},...,p_{W}\}$ is the probability distribution of the $W$
species in the multinomial model.

\begin{description}
\item[{Applications}] 
\end{description}

\begin{itemize}

\item \textbf{Environmental monitoring.} RQE helps calculate the environmental heterogeneity index and assist prioritization schemes \cite{DoxaPrastacos2020}.

\item \textbf{Genetic diversity metrics.} RQE is used to measure diversity for a whole collection of alleles to accommodate different genetic distance coding schemes and computational tractability in case of large datasets \cite{10.1371/journal.pone.0185499}.

\item \textbf{Unsupervised classification in risk management.} RQE is used as a framework in the support vector data description algorithm for risk management, enhancing knowledge in terms of interpretation, optimization, among others \cite{RePEc:ris:crcrmw:2018_006}.

\end{itemize}
   

\subsection{Relative Entropy}
\label{sec28}

Relative entropy, also known as Kullback-Leibler (KL) divergence \cite{Cover2006}, is
an information-theoretical measure that quantifies the difference or
\textquotedblleft distance\textquotedblright\ between two probability
distributions. Specifically, if $\mathrm{p}=(p_{1},...,p_{W})$ and $\mathrm{q%
}=(q_{1},...,q_{W})$ are two probability distributions, then the relative
entropy or KL divergence from $\mathrm{p}$ to $\mathrm{q}$, $D\left( 
\mathrm{p}\right\Vert \mathrm{q})$, is defined as%
\begin{equation}
D\left( \mathrm{p}\right\Vert \mathrm{q})=\sum_{i=1}^{W}p_{i}\log \frac{p_{i}%
}{q_{i}}.  \label{KL Div}
\end{equation}%
It follows that $D\left( \mathrm{p}\right\Vert \mathrm{q}) \geq 0$ (Gibb's inequality) and $D\left( \mathrm{p}\right\Vert \mathrm{q}) = 0$ if and only if $\mathrm{p}=\mathrm{q}$. Note that $D\left( \mathrm{p}\right\Vert \mathrm{q})$ is not a distance in the strict sense because $D\left( 
\mathrm{p}\right\Vert \mathrm{q})\neq D\left( \mathrm{q}\right\Vert \mathrm{p})$ in general, although it can be easily symmetrized by taking any mean (arithmetic, geometric, harmonic,...) of  $D\left( \mathrm{p}\right\Vert \mathrm{q})$ and $D\left( \mathrm{q}\right\Vert \mathrm{p})$. Therefore, if $\mathrm{p}$ is a true probability distribution approximated with $\mathrm{q}$, then $D\left( \mathrm{p}\right\Vert \mathrm{q})$ is a measure of the approximation error. As another useful example, if $\mathrm{p}$ is a bivariate joint distribution and $\mathrm{q}$ is the product distribution of the two marginals, then $D\left( \mathrm{p}\right\Vert \mathrm{q})$ is the mutual information between the random variables defined by the marginal distributions, equation (\ref{mutual info}).

See \cite{csiszar2008} for a generalizacion of divergence (or relative entropy), where the $\log
p_{i}/q_{i}$ in equation (\ref{KL Div}) is replaced by $%
f(p_{i}/q_{i})$ where $f$ is a convex function on $(0,\infty )$ with $f(1)=0$. 

\begin{description}
\item[{Applications}] 
\end{description}

\begin{itemize}

\item \textbf{Anomaly detection in plane control.} KL divergence has been used for plane control in Software-Defined Networking as a method to detect Denial of Service attacks in \cite{9869333}.

\item \textbf{Bayesian networks.} The efficient computation of the KL divergence of two probability distributions, each one coming from a different Bayesian network (with possibly different structures), has been considered in \cite{e23091122}.

\item \textbf{Feature selection.} The authors of  \cite{10.5555/2503308.2188387} show that the KL divergence is useful in information-theoretic feature selection due to the fact that maximising conditional likelihood corresponds to minimising KL-divergence between the true and predicted class posterior probabilities. 

\item \textbf{Multiscale errors.} KL divergence is a useful metric in ML for multiscale errors \cite{Bishop2006}. A recent application to the study and analysis of the behavior of various nonpolar liquids via the Relative Resolution algorithm can be found in \cite{doi:10.1021/acs.jctc.3c01052}.

\item \textbf{Parameter minimization in ML.} Parameters that minimize the KL divergence minimize also the cross entropy and the negative log likelihood. So, the KL divergence is useful in optimization problems where the loss function is a cross-entropy \cite{draelos2019connections}.

\end{itemize}


\subsection{Rényi Entropy}
\label{sec1}

Rényi entropy $R_{\alpha}$, where $\alpha > 0$ and $\alpha \neq 1$, was introduced by Alfréd Rényi in 1961 \cite{renyi1961entropy}  as a generalization of Shannon entropy in the sense that $R_1$ is set equal to the latter by continuity; see Section \ref{secSKaxioms} and the review \cite{Amigo2018} for detail. The parameter $\alpha$ allows for different emphasis on the probabilities of events, making it a versatile measure in information theory and its applications. Thus, for $\alpha < 1$ the central part of the distribution is flattened, i.e., high probability events are suppressed, and low-probability events are enhanced. The opposite happens when $\alpha < 1$. As a function of the parameter $\alpha$, the Rényi entropy is non-increasing. Particular cases include the Hartley entropy or max-entropy $R_{0}=\lim_{\alpha \rightarrow 0}R_{\alpha }$, the collision or quadratic entropy $R_{2}$, and the min-entropy $R_{\infty }=\lim_{\alpha \rightarrow \infty }R_{\alpha}$. 

\begin{description}
\item[{Applications}] 
\end{description}

\begin{itemize}

\item \textbf{Anomaly Detection.} Lower values of $\alpha$ highlight rare events, making this measure useful for identifying anomalies \cite{DeLaPavaPanche2019,Rioul2023}. In particular, Rényi entropy is used in network intrusion detection for detecting botnet-like malware based on anomalous patterns \cite{Berezinski2015}.

\item \textbf{Automated identification of EEGs.} Average Renyi entropy, along with other entropic measures, have been used as inputs for SVM algorithms to classify focal or non-focal EEGs of subjects affected by partial epilepsy \cite{Sharma2015}.

\item \textbf{Clustering.} Rényi entropy can provide robust similarity measures that are less sensitive to outliers \cite{DeLaPavaPanche2019}.

\item \textbf{Extreme entropy machines.} Rényi's quadratic entropy $R_{2}$ is used in the construction of extreme entropy machines to improve classification problems \cite{czarnecki2015extremeentropymachinesrobust}.

\item \textbf{Feature selection and character recognition.} Adjustment of the parameter $\alpha$ can help to emphasize different parts of the underlying probability distribution and hence the selection of the most informative features. Rényi entropy is used for feature selection in \cite{DeLaPavaPanche2019,Sluga2017}. Max-entropy is used in \cite{9254346} for convolutional feature extraction and improvement of image perception.

\item \textbf{Gaussianity of linear random processes}. Differential and conditional Rényi entropy rates were used in \cite{lake2005} to develop a measure of the Gaussianity of a continue-valued linear random process and study the dynamics of the heart rate.

\item \textbf{Medical time series analysis.} Applications of the Rényi entropy to medical time series analysis reach from epilepsy detection in EEG (see, e.g., \cite{kannathal2005}) over artifact rejection in multichannel scalp EEG (see \cite{mammone2012} and references therein) to early diagnosis of Alzheimer's disease in MEG data (see, e.g., \cite{poza2008}.

\end{itemize}


\subsection{Sample Entropy }
\label{sec33}
   
Sample Entropy was introduced by Richman and Moorman in 2000 \cite{richman2000} as an improvement over approximate entropy (Section \ref{sec2}), namely, its calculation is easier and independent of the time series length. Sample entropy is the negative natural logarithm of the conditional probability that close sequences of $m$ points remain close when one more point is added, within a tolerance $r>0$. 

\begin{description}
\item[{Applications}] 
\end{description}

\begin{itemize}

\item \textbf{Automated identification.} Average sample entropy and other entropy measures are used as input for an SVM algorithm to classify focal and non-focal EEG signals of subjects with epilepsy \cite{Sharma2015}.

\item \textbf{Fault diagnosis.} Sample entropy has been used for multi-fault diagnosis in lithium batteries \cite{ShangEtAl2020}.

\item \textbf{Image classification.} Sample entropy, in the form of multivariate multiscale sample entropy, is used for classifying RGB colour images to compare textures, based on a threshold to measure similarity \cite{e24111577}.

\item \textbf{Image texture analysis.} Two-dimensional sample entropy has shown to be a useful texture feature quantifier for the analysis of biomedical images \cite{SilvaEtAl2016}.

\item \textbf{Mutual information.} Modified sample entropy has been used in skin blood flow signals to analyse mutual information and, hence, study the association of microvascular dysfunction in different age groups \cite{LiaoJan2016}.

\item \textbf{Neonatal heart rate variability.} Sample entropy was used in \cite{Lake2002} to analyze the neonatal heart rate variability. The authors also address practical issues such as the selection of optimal tolerance and embedding dimension, and the impact of missing data. They find that entropy falls before clinical signs of neonatal sepsis and that missing points are well tolerated. 

\item \textbf{Neurodegenerative disease classification.} Sample entropy is used to classify neurodegenerative diseases. Gait signals, support vector machines and nearest neighbours are employed to process the features extracted using sample entropy \cite{NguyenLiuLin2020}.

\item \textbf{Short signal analysis.} The "coefficient of sample entropy" (COSEn) was introduced in \cite{lake2011} for the analysis of short-length physiological time series.

\item \textbf{Time series analysis.} Sample entropy, often in the form of multiscale sample entropy, is a popular tool in time series analysis, in particular with biomedical data \cite{HumeauHeurtier2018}. For example, it is used for the fast diagnosis and monitoring of Parkinson's disease \cite{BelyaevEtAl2023} and human emotion recognition \cite{patel2021} using EEGs. A modified version of multiscale sample entropy has recently been used for diagnosing epilepsy \cite{LinLin2022}. See \cite{richman2000} for an overview of applications of sample entropy to the analysis of physiological time series.

\item \textbf{Weather forecasting.} Sample entropy is applied in weather forecasting by using transductive feature selection methods based on clustering-based sample entropy \cite{KarevanSuykens2018}.

\end{itemize}


\subsection{Shannon Entropy }
\label{sec34} 

Shannon entropy is the prototypical entropy for characterizing known or empirical probability distributions, predicting unknown or unobservable distributions via the maximum entropy principle (possibly under constraints) \cite{PhysRev.106.620}, and optimizing models by minimizing entropy-based cost functions (see e.g. Sections \ref{sec7} and \ref{sec28}). As explained in Section \ref{secSKaxioms}, it was introduced by Claude Shannon in his foundational 1948 paper "A Mathematical Theory of Communication" \cite{Shannon1948} as the cornerstone of digital and analog Information Theory. Indeed, Shannon entropy informs the core theorems of Information Theory \cite{Cover2006}. In the case of discrete probability distributions, the Shannon entropy measures the uncertainty about the outcome of a random variable with the given distribution or, alternatively, the expected information conveyed by that outcome, maximum uncertainty (or minimum information) being achieved by uniform distributions. It also quantifies the rate of information growth produced by a data source modeled as a stationary random process. In the case of continuous probability distributions, the Shannon entropy is called differential entropy and its applications to data analysis and machine learning were the subject of Section \ref{sec8}. 

In this section we only consider discrete probability distributions, i.e., finite state random variables and processes (possibly after a discretization or symbolization of the data). A typical example are real-valued time series, where the Shannon entropy is used to measure their complexity and, hence, distinguish between different dynamics. The Shannon entropy of certain probability mass distributions (e.g., ordinal patterns of a time series, power spectrum of a signal, eigenvalues of a matrix) may have particular names (permutation, spectral, von Neumann entropies); in this case, the applications of such entropies are presented in the corresponding sections. 

\begin{description}
\item[{Applications}] 
\end{description}

\begin{itemize}

\item \textbf{Accurate prediction.} Shannon entropy is employed in machine learning models to improve the accuracy of predictions of molecular properties in the screening and development of drug molecules and other functional materials \cite{GuhaVelegol2023}.

\item \textbf{Anomaly detection.} Shannon entropy is employed in sensors (Internet of Things) to identify anomalies using the CorrAUC algorithm \cite{DeMedeirosEtAl2023}.

\item \textbf{Artificial intelligence.} 
Shannon entropy contributes to the creation of the Kolmogorov Learning Cycle, which acts as a framework to optimize "Entropy Economy", helped by the intersection of Algorithmic Information Theory (AIT) and Machine Learning (ML). This framework enhances the performance of the Kolmogorov Structure Function, leading to the development of "Additive AI". By integrating principles from both AIT and ML, this approach aims to improve algorithmic efficiency and effectiveness, driving innovation in AI by balancing information theory with practical machine learning applications \cite{Evans2024}.

\item \textbf{Automated identification of EEG signals.} Average Shannon entropy and other entropy measures are used as inputs of an SVM algorithm to classify focal or non-focal EEG signals of subjects with epilepsy \cite{Sharma2015}.

\item \textbf{Classification and feature detection.} The restricted Boltzmann machine is a probabilistic graphic network \cite{HintonSejnowski1986,Goodfellow2016,Oh2020} that is widely used for classification and feature detection \cite{marullo2021}. The name is due to the fact that its connectivity is constrained \cite{Goodfellow2016} and the probability of finding the network in a certain configuration is given by the Boltzmann distribution (Section \ref{secSKaxioms}). If considered as cognitive processes (i.e., with efficient learning and information retrieval), restricted Boltzmann machines are equivalent to Hopfield networks \cite{Goodfellow2016}, meaning that each one can be mapped onto the other \cite{marullo2021,smart2021}. Both of them are among the most popular examples of neural networks.

\item \textbf{Classification in the small data regime}. eSPA+ is an evolution of the entropy-optimal scalable probabilistic approximation algorithm (eSPA) \cite{HorenkoeSPA} that has better stability and a lower iteration cost scaling than its precursor. According to \cite{HorenkoeSPAplus}, eSPA+ outperforms other popular approaches like support vector machines, random forest, gradient boosting machine and long short-term memory (LSTM) networks in classification problems in the small data regime, i.e., when handling data sets where the number of feature dimension is considerably higher than the number of observations.

\item \textbf{Fault bearing diagnosis.} Multi-scale stationary wavelet packet analysis and the Fourier amplitude spectrum are combined to obtain a new discriminative Shannon entropy feature that is called stationary wavelet packet Fourier entropy in \cite{e21060540}. Features extracted by this method are then used to diagnose bearing failure.

\item \textbf{Feature selection.} Shannon’s mutual information entropy is used to design an entire information-theoretic framework that improves the selection of features  in \cite{10.5555/2503308.2188387}. Shannon entropy is employed in biological science to improve classification of data in clustering of genes using microarray data \cite{SoltanianEtAl2019}.

\item \textbf{Hard clustering.} Shannon entropy is used as a criterion to measure the confidence in unsupervised clustering tasks \cite{HoayekRulliere2023}.

\textbf{Maximum entropy and feature extraction}. By sampling from the maximum entropy distribution over possible sequences in multilayer artificial neural networks, the authors of \cite{Finnegan2017} present a method for interpreting neural networks and extracting the features that the network has learned from the input data. The authors also apply their approach to biological sequence analysis.

\item \textbf{Maximum entropy and scalability in AI/ML models}. In \cite{MeMe2019} the authors propose a new Maximum Entropy Method (MEMe) which improves upon the scalability of existing machine learning algorithms by efficiently approximating computational bottlenecks using maximum entropy and fast moment estimation techniques.

\item \textbf{Mislabeled data in supervised classification.} Entropic outlier sparsification (EOS) is used in robust learning when data present anomalies and outliers \cite{HorenkoEOS2022}. EOS leverages the analytic solution of the (weighted) expected loss minimization problem subject to Shannon entropy regularization. EOS is tested with biomedical datasets in \cite{HorenkoEOS2022} with favorable results as compared to other methods. EOS can be also applied to feature selection and novelty detection problems.

\item \textbf{Natural language processing.} Shannon entropy quantifies the predictability (or redundancy) of a text. Therefore, it is instrumental in language modelling, text compression, information retrieval, among others \cite{Shannon1948,Cover2006}. For example, it is used in \cite{YANG20134523} for keyword extraction, i.e., to rank the relevance of words.

\item \textbf{Policy learning.} Shannon entropy acts as a regularization inside of an iterative policy optimization method for certain quadratic linear control scenarios \cite{Guo2023}.

\item \textbf{Regression learning}. The principle of entropy maximization is implemented in the sparse probabilistic approximation for regression task analysis (SPARTAn) algorithm, which is a computationally cheap and robust algorithm for regression learning \cite{HorenkoCheapEntropySparsified}. SPARTAn has been applied to predict the El Ni\~{n}o Southern Oscillation in \cite{HorenkoCheapEntropySparsified}, providing more predictive, sparse and physically explainable data descriptions than other alternative methods. 

\item \textbf{Signal analysis.} Shannon entropy is used as the cost functional of compression algorithms in sound and image processing \cite{CoifmanWickerhauser1992}.

\item \textbf{Statistical inference.} According to the Maximum Entropy Principle of Jaynes \cite{PhysRev.106.620}, "in making inferences on the basis of partial information we must use the probability distribution which has maximum entropy subject to whatever is known". This principle has been traditionally applied with the Shannon entropy and several moment constraints of a probability distribution to infer the actual distribution \cite{Cover2006,Amigo2022}. 

\end{itemize}


\subsection{Spectral Entropy}
\label{sec36} 

Spectral entropy, proposed by Kapur and Kesavan in 1992 \cite{kapur1992}, is an entropy based on the Shannon entropy. Here, the probability distribution is the (continuous or discrete) power spectrum in a representative frequency band, obtained from a signal or time series via the Fourier transform, and conveniently normalized. Hence, the spectral entropy characterizes a signal by the distribution of power among its frequency components.

\begin{description}
\item[{Applications}] 
\end{description}

\begin{itemize}
   
\item \textbf{Audio analysis.} Spectral entropy has been applied for robust audio content classification in noisy signals \cite{Wang2020}. Specifically, spectral entropy is used to segment input signals into noisy audio and noise. Also, spectral entropy (in the form of Multiband Spectral Entropy Signature) has been shown to outperform other approaches in the task of sound recognition \cite{ManzoMartinez2022}. 

\item \textbf{Damage event detection.} Spectral entropy detects damage in vibration recordings from a wind turbine gearbox \cite{Civera2022}.

\item \textbf{Data time compression.} Spectral entropy has been successfully applied to identify important segments in speech, enabling time-compression of speech for skimming \cite{Ajmal2007}.

\item\textbf{Deep learning synchronization.} Spectral entropy evaluates synchronization in neuronal networks, providing analysis of possibly noisy recordings collected with microelectrode arrays \cite{Kapucu2016}.

\item \textbf{Feature extraction.} Spectral entropy is used to extract features from EEG signals in \cite{patel2021} (emotion recognition) and \cite{Ra2021} (assessment of the depth of anaesthesia).

\item \textbf{Hyperspectral anomaly detection.} Hyperspectral Conditional Entropy enters into the Entropy Rate Superpixel Algorithm, which is used in hyper spectral-spatial data to recognize unusual patterns \cite{Liu2024}.
   
\item \textbf{Signal detection.} Spectral entropy has been used to detect cetacean vocalization in marine audio data \cite{Rademan2023}. The time frequency decomposition was done with the short time Fourier transform and the continuous wavelet transform.

\end{itemize}


\subsection{Tone Entropy }
\label{sec37}

Given a time series $(x_{i})_{1\leq i\leq N}$, its tone
entropy is the Shannon entropy (Section \ref{sec34}) of the probability distribution derived from
the percentage indices $\mathrm{PI}(i)=100(x_{i}-x_{i+1})/x_{i}$, $1\leq i\leq N-1$. Tone entropy was proposed by Oida et al. in 1997 \cite{oida1997} to study heart period fluctuations in electrocardiograms. Therefore, its applications are mainly in cardiology. 

\begin{description}
\item[{Applications}] 
\end{description}

\begin{itemize}

\item \textbf{Biomedical analysis.} Tone entropy has been employed to study the autonomic nervous system in age groups at high-risk of cardiovascular diseases \cite{Khandoker2019}. In \cite{Khandoker2015}, tone entropy was used to study the influence of gestational ages on the development of the foetal autonomic nervous system by analyzing the foetal heart rate variability. 

\item \textbf{Time series.} Tone entropy has been used in time series analysis to differentiate between physiologic and synthetic interbeat time series \cite{Karmakar2013}.

\end{itemize}


\subsection{Topological and Topology-based Entropies}
\label{sec38}
    
Topological entropy was introduced in \cite{Adler1965} to measure the complexity of continuous dynamics on topological spaces. On metric spaces, topological entropy measures the exponential growth rate of the number of distinguishable orbits with finite precision. See, e.g., \cite{Amigo2014,Amigo2015} for exact formulas and fast algorithms to compute the topological entropy of piecewise monotone maps and multimodal maps.

In time series analysis and digital communication technology, one is mostly interested in spaces with a finite number of states. Such time series can be the result symbolizing a continuous-valued time series. In this case, the states are usually called letters (or symbols), the state space is called alphabet and the blocks of letters are called words (which correspond to the "admissible" or "allowed" strings of letters). If $A(n)$ is the number of words of length $n$, then the topological entropy of the time series is defined as
\begin{equation}  
h_{top}=\lim_{n\rightarrow \infty }\log (A(n)/n),  \label{Top ent}
\end{equation}
where the base of the logarithm is usually 2 or e. 

Along with the above "conventional" topological entropies in dynamical systems and time series analysis, there are a number of adhoc entropies in time series analysis, sometimes also called topological entropies. This name is due to the fact that those entropies draw on topological properties extracted from the data, for example, via graphs or persistent homology. For clarity, here we refer to them as topology-based entropies. Examples include graph entropy and horizontal visibility graph entropy (Section \ref{sec19}). See \cite{Lum2013} for an account of topological methods in data analysis.

\begin{description}
\item[{Applications}] 
\end{description}

\begin{itemize}

\item \textbf{Cardiac dynamics classification.} Given a (finite) time series, the \textit{out-link entropy} is derived from the adjacency matrix of its ordinal network. This entropy has been used to classify cardiac dynamics in \cite{mccullough2017}.

\item \textbf{Convolutional neural networks.} The authors of \cite{Zhao2022} propose a method for quantitatively clarifying the status of single unit in convolutional neural networks using algebraic topological tools. Unit status is indicated via the calculation of a topology-based entropy, called \textit{feature entropy}.

\item \textbf{Damage detection in civil engineering.} Persistent entropy can also address the damage detection problem in civil engineering structures. In particular, to solve the supervised classification damage detection problem \cite{JimenezAlonso2019}.

\item \textbf{Detection of determinism in time series.} Permutation topological entropy (i.e., the topological entropy of the distribution of ordinal patterns of length $L\geq 2$ obtained from a time series) can be used to detect determinism in continuous-valued time series. Actually, it suffices to check the growth of ordinal patterns with increasing $L$'s, since this growth is exponential for deterministic signals ($h_{top}$ converges to a finite number) and factorial for random ones ($h_{top}$ diverges) \cite{amigo2010permutation,amigo2008forbidden}. 

\item \textbf{Financial time series analysis.} Topological entropy has been applied to horizontal visibility graphs for financial time series in \cite{Rong2018} to help quantifying
changes in complex stock market data.

\item \textbf{Similarity of piecewise linear functions.} Piecewise linear functions are a useful mathematical tool in different areas of applied mathematics, including signal processing and machine learning methods. In this regard, persistent entropy (a topological entropy based on persistent homology) can be used to measure their similarity \cite{Rucco2017}.

\end{itemize}


\subsection{Transfer Entropy}
\label{sec43}
    
A relevant question in time series analysis of coupled random or
deterministic processes is the causality relation, i.e., which process is
driving and which is responding. Transfer entropy, introduced by Schreiber
in 2000 \cite{PhysRevLett85461}, measures the information exchanged between two processes in both directions separately. It can be considered as an
information-theoretical (or nonlinear) implementation of Granger causality 
\cite{RePEc38}.

Given two stationary random processes $\mathbf{X}=(X_{t})_{t\ge 0}$
and $\mathbf{Y}=(Y_{t})_{t\ge 0}$, the transfer entropy from $%
\mathbf{Y}$ to $\mathbf{X}$, $T_{\mathbf{Y}\rightarrow \mathbf{X}}$, is the
reduction of uncertainty in future values of $\mathbf{X}$, given past values
of $\mathbf{X}$, due to the additional knowledge of past values of $\mathbf{Y%
}$. For simplicity, we consider here the simplest (lowest dimensional) case:%
\begin{equation}
T_{\mathbf{Y}\rightarrow \mathbf{X}}=S\left( X_{t+1}\right\vert
X_{t})-S\left( X_{t+1}\right\vert X_{t},Y_{t})  \label{transfer ent}
\end{equation}%
where $S\left( X_{t+1}\right\vert ...)$ is the conditional Shannon entropy
of the variable $X_{t+1}$ on the other variable(s) \cite{Cover2006}. If the process $\mathbf{Y%
}$ is not causal to $\mathbf{X}$ (i.e., they are independent), then $S\left(
X_{t+1}\right\vert X_{t},Y_{t})=S\left( X_{t+1}\right\vert X_{t})$ and $T_{%
\mathbf{Y}\rightarrow \mathbf{X}}=0$; otherwise, $S\left( X_{t+1}\right\vert
X_{t},Y_{t})<S\left( X_{t+1}\right\vert X_{t})$ and $T_{\mathbf{Y}%
\rightarrow \mathbf{X}}>0$. Observe that $T_{\mathbf{Y}\rightarrow \mathbf{X}%
}$ is not an entropy proper but a conditional mutual information, namely: $%
T_{\mathbf{Y}\rightarrow \mathbf{X}}=I(X_{t+1};Y_{t}\mid X_{t})$ \cite{Cover2006,Amigo2022}.

\begin{description}
\item[{Applications}] 
\end{description}

\begin{itemize}

\item \textbf{Accelerated training in Convolutional Neural Networks (CNN).} The authors of \cite{moldovan2021} propose a training mechanism for CNN architectures that integrates transfer entropy feedback connections. In this way, the training process is accelerated as fewer epochs are needed. Furthermore, it generates stability, hence, it can be considered a smoothing factor.

\item \textbf{Improving accuracy in Graph Convolutional Neural Networks (GCN).}  The accuracy of a GCN can be improved by using node relational characteristics (such as heterophily), degree information, and feature-based transfer entropy calculations. However, depending on the number of grah nodes, the computation of the transfer entropy can significantly increase the computational load \cite{moldovan2024transferentropygraphconvolutional}.

\item \textbf{Improving neural network performance.} A small, few-layer artificial neural network that employs feedback can reach top level performance on standard benchmark tasks, otherwise only obtained by large feed-forward structures. To show this, the authors of \cite{herzog2017} use feed-forward transfer entropy between neurons to structure feedback connectivity.

\item \textbf{Multivariate time series forecasting.} Transfer entropy is used to establish causal relationships in multivariate time series converted into graph neural networks, each node corresponding to a variable and edges representing the casual relationships between the variables. Such neural networks are then used for prediction \cite{9837007}.

\item \textbf{Time series analysis.} The main application of transfer entropy since its formulation has been the analysis of multivariate time series (whether biomedical, physical, economical, financial, ...) for revealing causal relationships via information directionality. See \cite{Amblard2013} and the references therein for the conceptual underpinnings and practical applications.

\end{itemize}


\subsection{Tsallis Entropy}
\label{sec40}
     
Tsallis entropy $T_{q}$, defined in Section \ref{secSKaxioms}, equation (\ref{Tsallis_entr}), was introduced by Tsallis in 1988 in Statistical Mechanics \cite{Tsallis1988}. The parameter $q$ takes the real values $q>0$, $q \ne 1$; $T_{q}$ converges to the Shannon entropy when $q\rightarrow 1$ \cite{Amigo2018}. Tsallis entropy is identical in form to the Havrda–Charvát entropy (Section \ref{sec21}).

Tsallis entropy is particularly useful for describing systems with non-extensive properties, such as long-range interactions, non-Markovian processes, and fractal structures. In machine learning, Tsallis entropy is used to improve algorithms in areas such as clustering, image segmentation, and anomaly detection by changing and fine tuning the parameter $q$. See \cite{Alomani2023} for the general properties of the Tsallis entropy.

\begin{description}
\item[{Applications}] 
\end{description}

\begin{itemize}

\item \textbf{Anomaly detection.} Tsallis entropy is used in network intrusion detection by detecting botnet-like malware based on anomalous patterns in the network \cite{Berezinski2015}.

\item \textbf{Clustering.} A Tsallis entropy based categorical data clustering algorithm is proposed in \cite{Sharma2019}. It is shown there, that when the attributes have a power law behavior the proposed algorithm outperforms existing Shannon entropy-based clustering algorithms.

\item \textbf{Feature selection.} Tsallis-entropy-based feature selection is used in  \cite{Wu2022} to identify significant features, which boosts the classification performance in machine learning. The authors propose an algorithm to optimize both the classifier (a Support Vector Machine) and Tsallis entropy parameters, so improving the classification accuracy.

\item \textbf{Image segmentation.} Tsallis Entropy is maximized to use it for segmenting images by maximizing the entropy within different regions of the image \cite{Naidu2017}.

\item \textbf{Pre-seismic signals.} Tsallis entropy has been used in \cite{Kalimari2008} to analyze pre-seismic electromagnetic signals.

\end{itemize}


\subsection{Von Neumann Entropy}
\label{sec41}

Von Neumann entropy \cite{MathematischeGrundlagen} is the equivalent of Shannon entropy in Quantum Statistical Mechanics and Quantum Information Theory. It is defined via the density matrix of an ensemble of quantum states, which (i) is Hermitian, (ii) has unit trace, and (iii) is positive semidefinite. Therefore, the eigenvalues of the density matrix build a probability distribution and, precisely, the von Neumann entropy of the ensemble is the Shannon entropy of the probability distribution defined by the eigenvalues of the corresponding density matrix. 

The same approach can be used with any matrix with the properties (i)-(iii), for example, the Pearson correlation matrix of a Markov chain (divided by the dimension of the matrix) or the Laplacian matrix of a graph. This explains the use of von Neumann entropy in classical data analysis as well.

\begin{description}
\item[{Applications}] 
\end{description}

\begin{itemize}

\item \textbf{Feature selection and dimensionality reduction.} Von Neumann entropy is employed in the case of kernelized relevance vector machines to asses dimensionality reduction for better model performances \cite{e25010154}.

\item \textbf{Graph-based learning.} In \cite{Hu2023} the authors propose a method to identify vital nodes in hypergraphs that draws on von Neumann entropy. More precisely, this method is based on the high-order line graph structure of hypergraphs and measures changes in network complexity using von Neumann entropy.

\item \textbf{Graph similarity and anomaly detection.} The von Neumann graph entropy (VNGE) is used to measure the information divergence and distance between graphs in a sequence. This is used for various learning tasks involving network-based data. The Fast Incremental von Neumann Graph Entropy algorithm reduces the computation time of the VNGE, making it feasible for real-time applications and large datasets \cite{Chen2018}.

\item \textbf{Network analysis.} Von Neumann entropy is used in \cite{YeEtAl2017} to build visualization histograms from the edges of networks and then component analysis is performed on a sample for different networks.

\item \textbf{Pattern recognition in neurological time series.} Von Neumann entropy has been used (together with other entropies) in \cite{huang2024entropy} for automated pattern recognition in neurological conditions, a crucial task in patient monitoring and medical diagnosis.

\end{itemize}


\subsection{Wavelet Entropy }
\label{sec42}
    
Wavelet entropy was introduced by Rosso et al. in 2001 \cite{Rosso2001}. It combines the wavelet transform with the concept of entropy to analyze the complexity and information content of multi-frequency signals. More precisely, the wavelet transform decomposes a signal into components at various scales, capturing both time and frequency information, while wavelet entropy quantifies the degree of disorder or unpredictability in those components, thus providing insights into the signal structure and complexity. Wavelet entropy is defined by the Shannon formula, but here the probability distribution corresponds to different resolution levels.

\begin{description}
\item[{Applications}] 
\end{description}

\begin{itemize}

\item \textbf{Emotion recognition.} Wavelet entropy can detect little variations in signals and it has been used in \cite{patel2021} to develop an automatic EEG classifier.

\item \textbf{Fault detection.} Wavelet entropy is applied to monitor the condition of machinery and detect faults by analysing vibration signals \cite{Hu20232}.

\item \textbf{Feature extraction.} Wavelet entropy is used to extract features from biomedical signals such as EEG and ECG to identify different physiological states or detect abnormalities \cite{Rosso2001}.

\end{itemize}


\section{\large Discussion}
\label{sec3.0}

In Section \ref{secIntro} of this review, we provided a brief historical account of
the concept of entropy. Due to its seemingly unconnected appearances in
Thermodynamics, Statistical Mechanics, Information Theory, Dynamical
Systems, etc., this was necessary to clarify its role in data analysis and
machine learning, the subject of the present review. To this end, we gave an
axiomatic characterization of entropy and generalized entropies, and
acknowledged the inclusion in our review of several entropy-based
probability functionals that also go by the name of entropy in the
literature and are useful for the mentioned applications. In the case of
univariate arguments, entropy quantifies the uncertainty, complexity, and
information content of the data. In the case of bivariate and multivariate
arguments, entropy quantifies similarity, distance, and information flow.

In Section \ref{secEntropyList} we collected a representative sample of 33 entropies (including
the classical entropies visited in Section \ref{secSKaxioms}) to show the versatility and
potential of the general concept of entropy in practical issues. Indeed,
applications such as biomedical signal analysis, fault diagnosis, feature
extraction, anomaly detection, optimization cost, and more highlight the
diversity of the applications of entropy in data analysis and machine
learning. 

Section \ref{secEntropyList} also showed that, more than 150 years after its formulation,
entropy and its applications remain the subject of intense research. In
fact, new concepts of entropy, evolutions and generalizations are constantly
being proposed in the literature to address new challenges. As a result, entropy is being applied to current topics of applied mathematics, in particular, data analysis and machine learning.

Next we turn our attention to a more quantitative description of the entropies used in this review, their applications, and the references cited. 

First of all, the Groups G0, G1 and G2 defined in Section \ref{sec1.3} (G0 comprising the classical entropies of Section \ref{secSKaxioms}), provide a coarse classification of the entropies that is sufficient for the following analysis of their relationship with the applications and the number of citations. 

Furthermore, Figure 1 (a WordCloud for ease of visualization) shows that data analysis is the application most commonly encountered in Section \ref{secEntropyList}, the reason being that this concept brings together all types of data analysis. The rest are applications in machine learning, where the most common ones are feature selection/extraction, series classification and anomaly detection.

\begin{figure}
\centering
\fbox{\includegraphics[width=10.5 cm]{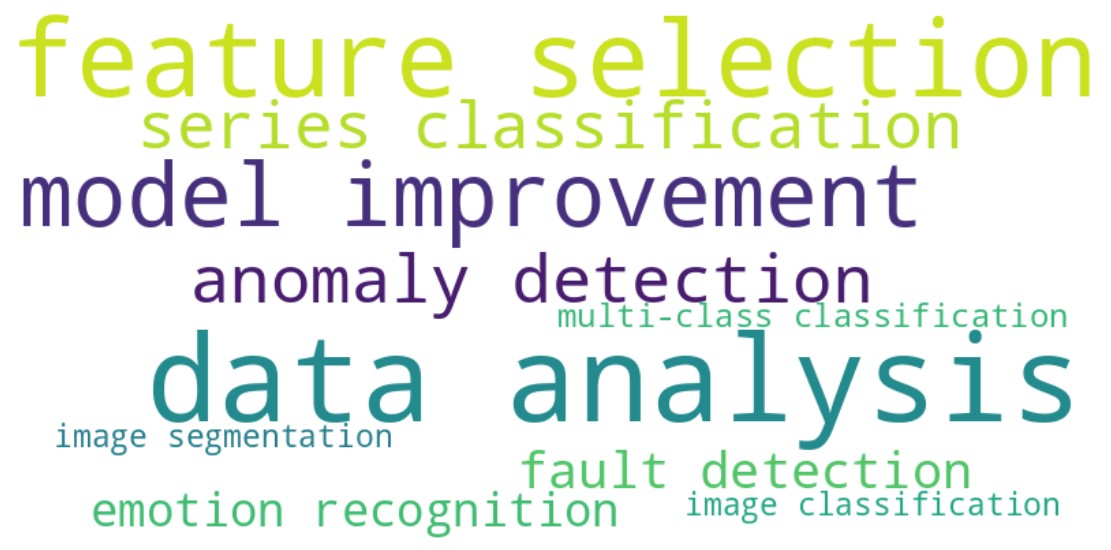}}
\caption{\centering Top 10 applications across entropies.}
\end{figure}

The relationship between applications and entropy groups is visualized in Figure 2 by a Sankey diagram, where the applications chosen (leftmost column) are the top 10 applications across entropies according to Figure 1. The diagram shows a satisfactory balance between the most popular applications of entropy and the three entropy groups, Group G0 ranking slightly higher than Groups G1 and G2.

\begin{figure}
\centering
\includegraphics[width=15.5cm]{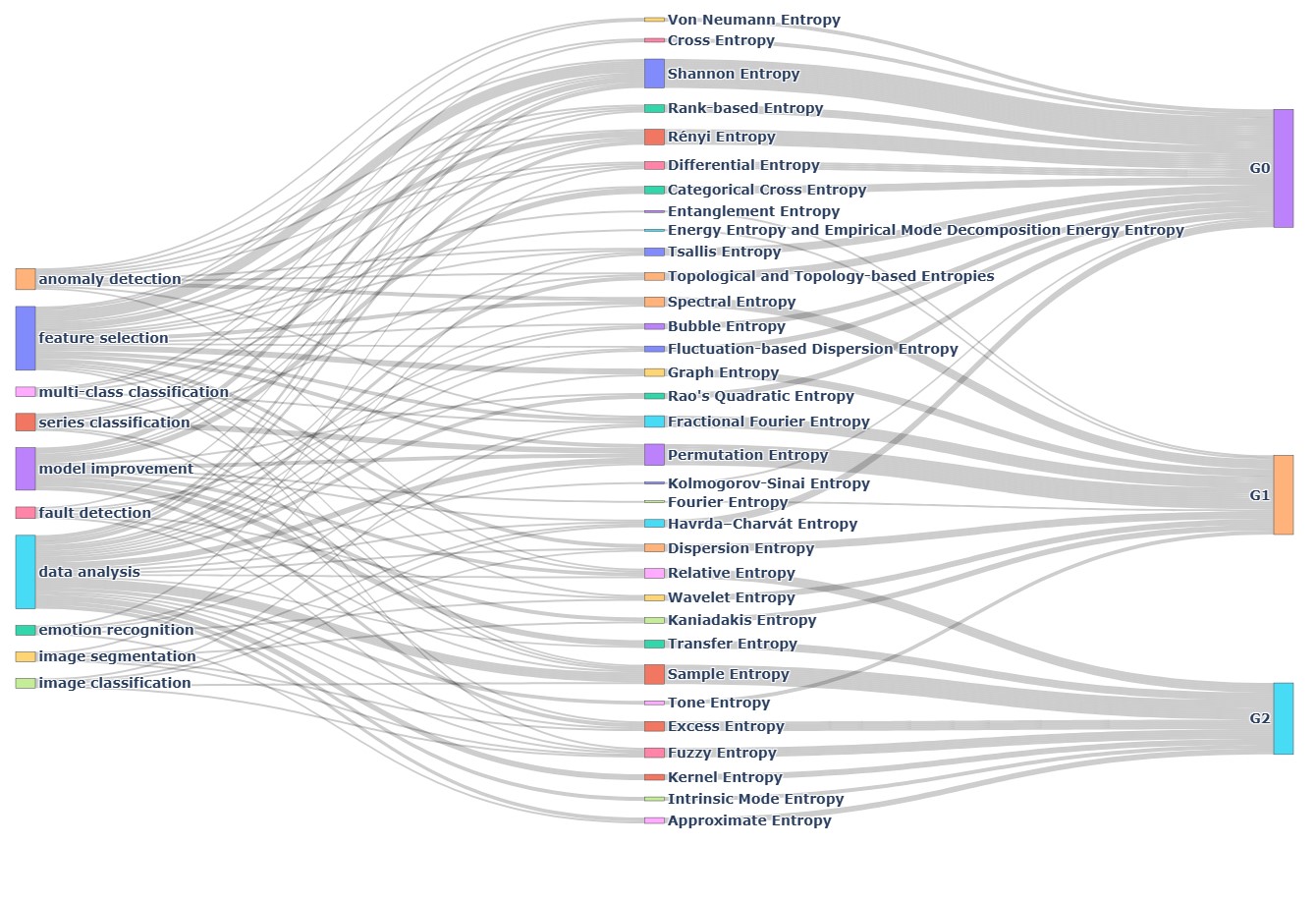}
\caption{\centering Relationship between applications and entropies (categorized as Groups G0/G1/G2).}
\end{figure}  

As for the number of citations and publications, Table 1 ranks the 10 most cited references, according to a DOIs citation analysis to date. Along with Shannon's and Tsallis' seminal papers (G0), there are 4 references on entropies from Group G1 and another 4 from Group G2, which shows again a tie between the two groups. Among these most influential papers, we see two G2 papers published in 2000 (authored by Richmann and Moorland, and Schreiber) and two G1 papers published shortly afterwards (by Bandt and Pompe, and Costa et al.), all of them related to signal analysis and, more specifically, to the analysis of biological signals in two of them.

\begin{table}
\centering
\scalebox{1.2}{\begin{tabular}{|c|c|c|c|c|}
\hline
\textbf{Reference} & \textbf{Year} & \textbf{Authors} & \textbf{Total
Citations} & \textbf{Taxonomy} \\ \hline\hline
1 & 1948 & Shannon & 43150 & G0 \\ \hline
31 & 2000 & Richman, Moorman & 16606 & G2 \\ \hline
18 & 1988 & Tsallis & 14855 & G0 \\ \hline
22 & 1991 & Pincus & 13763 & G2 \\ \hline
148 & 2002 & Bandt, Pompe & 10260 & G1 \\ \hline
248 & 2000 & Schreiber & 9743 & G2 \\ \hline
226 & 1992 & Coifman, Wickerhause & 8003 & G1 \\ \hline
21 & 2005 & Costa, Goldberger, Pen & 5624 & G1 \\ \hline
101 & 1972 & De Luca, Termini & 5217 & G2 \\ \hline
178 & 1982 & Rao & 4027 & G1 \\ \hline
\end{tabular}}
\vspace{6pt} 
\caption{Most referenced publications by DOI as of this writting (December 2024). The total citations are based on Semantic Scholar, CrossRef and OpenCitations.}
\end{table}

Finally, Figure 3 is a bar chart of the number of publications per year from 2000 on. In fact, it was around the year 2000 when the number of papers on entropy began to increase, after an initial period where that number was small and practically constant. This shows the growing interest that the concept of entropy and its applications has experienced in the last two decades or so among researchers in general and data analysts in particular. 

\begin{figure}
\begin{centering}
\includegraphics[width=12.5 cm]{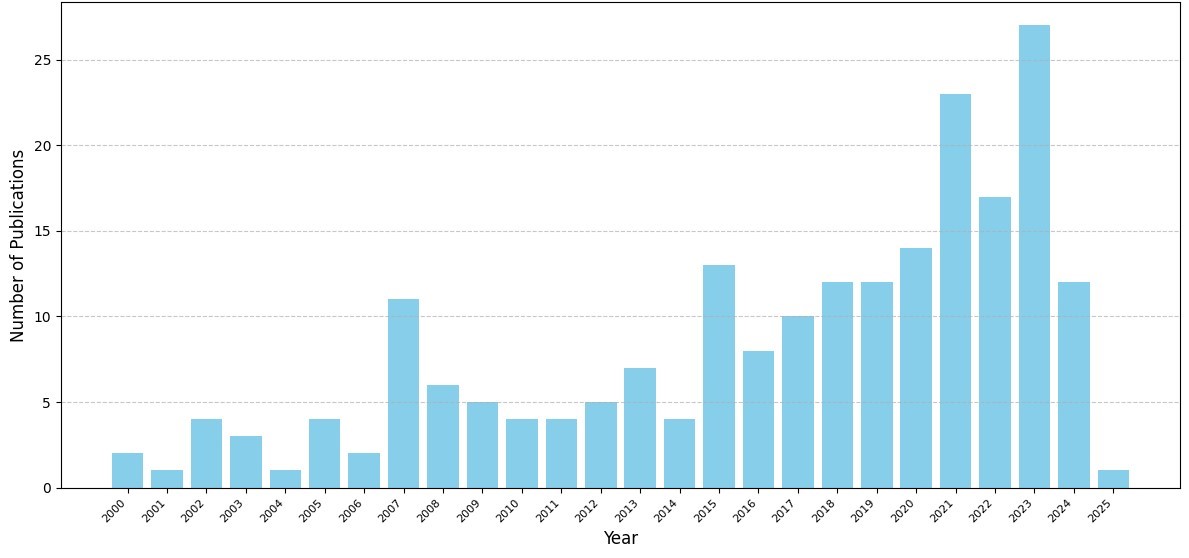}
\caption{\centering Number of publications per year on entropy and its applications (2000 onwards)}
\end{centering}
\end{figure}   
\unskip

\section{\large Conclusions and Outlook}
\label{sec4.0}

To conclude this review, we begin with a few remarks on the materials presented or omitted in the previous chapters.

\begin{itemize}

\item The choice of a particular entropy-based method depends in general on the application. Some methods may be more popular than others because they are designed for the purpose or dataset at hand, or simply because they have some computational advantage. In this regard, Section \ref{secEntropyList} presented possible candidates for different applications but no performance comparison between them was discussed. In fact, such a comparison would require a case-by-case test as, for example, in Reference \cite{MotorFault2023}, where the authors study motor fault detection with the approximate, dispersion, energy, fuzzy, permutation, sample and Shannon entropies (see Section \ref{sec17} for the best performer). Along with the selection of the \textquotedblleft right\textquotedblright\ entropy, a common concern among practitioners is the choice of parameters and hyperparameters. A combination of methods and parameter settings may be also a good approach in practice \cite{Graff2023} \cite{huang2024entropy}.

\item We have not included applications of entropy to cryptography in this
review because they belong to the general field of Data Science (through
data security) rather than to Data Analysis. Entropy (mainly Shannon's and R%
\'{e}nyi's entropies) have been applied to measure the \textquotedblleft
randomness\textquotedblright\ of encrypted messages in the so-called chaotic
cryptography, which applies ideas from Chaos Theory and chaos
synchronization to the masking of analog signals \cite{Cuomo1993}. To deal with digital signals, new tools such as discrete entropy \cite{AmigoDiscreteEnt2007} and discrete Lyapunov exponents \cite{AmigoDiscreteLE2007} have also been developed for application to chaotic cryptography, inspired by their conventional counterparts. For a general review of the applications of entropy and related information-theoretical concepts in cryptography, see the review \cite{EntropyInCrypto2022}.  

\item We have not delved into the numerical methods to compute entropies from the data.
Being functionals of probability distributions, most methods for computing entropy are based on the estimation of data probabilities. In the case of discrete-valued data, the probabilities are usually estimated via relative frequencies (the maximum likelihood estimator) and possibly extrapolation methods in case of undersampling \cite{amigo2010permutation}. In the case of continuous-valued data, the probability densities are usually estimated via kernel density estimation (also called Parzen-Rosenblatt windowing) \cite{Hastie2009}. Furthermore, there are particular methods that do not rely on probability estimation like, e.g., Lempel-Ziv complexity, which resorts to pattern matching and approximates the Shannon entropy rate of stationary binary sequences \cite{Amigo2004}. Also, in some particular cases, the entropy can be estimated via spectral information. For example, the quadratic R\'{e}nyi entropy $R_{2}$ can be estimated via spectral data transformations and kernel matrices \cite{Jenssen2010}. See \cite{Paninski2003} for a general review on the estimation of entropy.

\end{itemize}

So far, we have looked back and reviewed the applications of entropy in data analysis and machine learning to date. When we look forward, we see unsurprisingly many challenging topics waiting to be explored and further developed. To conclude this review, we comment on some promising, future research directions.

\begin{itemize}

\item \textbf{Algebraic Representations of Entropy}. The study and application of information-theoretical tools and, in particular, entropies, can benefit from an algebraic approach in different ways. A first example is permutation entropy (Section \ref{sec30}), which is the Shannon entropy of the symbolic representation of a continuous-valued stochastic process obtained using sliding windows and ordinal patterns. Since ordinal patterns can be interpreted as permutations, such representations can harness the algebraic structure of the symmetric group. This is what happens, e.g., with the concept of transcript; see \cite{AmigoMonetti2016} for definitions and applications. A second example is the correspondence between Shannon entropy (as well as other multivariate entropic quantities for that matter) and set-theoretic expressions \cite{Yeung1991}. This correspondence provides a handle to apply algebraic concepts and tools to information-theoretical quantities such as several multivariate generalizations of mutual information; see, e.g., \cite{Mediano2024,Baudot2015}. 

\item \textbf{Data Dimensional Reduction}. In general data analysis it is crucial to transform potentially high-dimensional data into a lower dimensional representation. Along with linear methods, such as principal component analysis (PCA) and metric multidimensional scaling, there are also nonlinear spectral methods such as kernel PCA \cite{Schoelkopf1998}. Interestingly, some of these methods involve entropies. Thus, in kernel entropy component analysis (ECA), the data transformation is related to the Rényi entropy of the input space data set \cite{Jenssen2010}. This and similar approaches could be generalized with the help of other entropies.

\item \textbf{Partial Information Decomposition} (PID) decomposes the mutual information $I(X_{1}$ $X_{2},...,X_{n};T)$ (equation (\ref{mutual info}) with a vector variable in the first argument) between a series of random source variables $X_{1},...,X_{n}$ and a target variable $T$ into the amount of influence that each $X_{i}$ has on $T$. In the simplest case of two sources $X_{1}$ and $X_{2}$, that influence splits into four components, although there is much controversy as to how these four components should be quantified \cite{James2017}. Other issues such as the applications of PID \cite{James2019}, and the generalization of PID to continuous-valued variables offer new topics worth researching; see, e.g., \cite{Ehrlich2024}. 

\item \textbf{Quantum generalized entropies.} In our list of entropies, we included two that arise from quantum mechanics, namely, von Neumann entropy (Section \ref{sec41}) and entanglement entropy (Section \ref{sec12}). With the advent of quantum information theory and quantum neural networks, the possible applications of conventional entropic quantities and, above all, quantum generalized entropies is a wide open field for further research. See \cite{Bosyk2016} for the definition, properties and applications of quantum generalized entropies.

\item Finally, let us point out that, many of the applications of entropy mentioned in Section 2 can be refined and diversified, for example, in the formulation and/or regularization of cost functions, decision trees, or generalization of concepts. Other interesting research field is the improvement of existing algorithms for computing entropies and entropic quantities (especially of multivariate quantities and time-continuous signals) and the introduction of new numerical techniques, possibly adapted to new progress in deep learning, reservoir computing and more.  

\end{itemize}

\vspace{6pt}


\section{Acknowledgments}
This work was financially supported by Generalitat Valenciana, Spain, grant PROMETEO/2021/063.


\bibliography{ref}

\begin{thebibliography}{100}
\expandafter\ifx\csname url\endcsname\relax
  \def\url#1{\texttt{#1}}\fi
\expandafter\ifx\csname urlprefix\endcsname\relax\def\urlprefix{URL }\fi
\expandafter\ifx\csname href\endcsname\relax
  \def\href#1#2{#2} \def\path#1{#1}\fi

\bibitem{Shannon1948}
C.~E. Shannon, A mathematical theory of communication, The Bell System
  Technical Journal 27~(3) (1948) 379--423.
\newblock \href {https://doi.org/10.1002/j.1538-7305.1948.tb01338.x}
  {\path{doi:10.1002/j.1538-7305.1948.tb01338.x}}.

\bibitem{Kantz1999}
H.~Kantz, T.~Schreiber, Nonlinear Time Series Analysis, Cambridge University
  Press, Cambridge UK, 1999.
\newblock \href {https://doi.org/10.1017/CBO9780511755798}
  {\path{doi:10.1017/CBO9780511755798}}.

\bibitem{Goodfellow2016}
I.~Goodfellow, Y.~Bengio, A.~Courville, Deep Learning, The MIT Press, Cambridge
  Massachusetts, 2016.
\newblock \href {https://doi.org/10.1007/s10710-017-9314-z}
  {\path{doi:10.1007/s10710-017-9314-z}}.

\bibitem{MathematischeGrundlagen}
J.~von Neumann, Mathematische Grundlagen der Quantenmechanik, Springer Verlag,
  Berlin, Heidelberg, New York, 1971.
\newblock \href {https://doi.org/10.1007/978-3-642-96048-2}
  {\path{doi:10.1007/978-3-642-96048-2}}.

\bibitem{gibbs1902}
J.~W. Gibbs, Elementary Principles in Statistical Mechanics: Developed with
  Especial Reference to the Rational Foundation of Thermodynamics, Charles
  Scribner's Sons, Farmington Hills, MI, USA, 1902.

\bibitem{kolmogorov1959}
A.~N. Kolmogorov, Entropy per unit time as a metric invariant of automorphisms,
  Doklady of the Russian Academy of Sciences 124 (1959) 754--755.

\bibitem{sinai1959}
Y.~G. Sinai, On the notion of entropy of a dynamical system, Doklady of the
  Russian Academy of Sciences 124 (1959) 768--771.

\bibitem{Adler1965}
R.~L. Adler, A.~G. Konheim, M.~H. McAndrew, Topological entropy, Transactions
  of the American Mathematical Society 114~(2) (1965) 309--319.
\newblock \href {https://doi.org/10.1090/S0002-9947-1965-0175106-9}
  {\path{doi:10.1090/S0002-9947-1965-0175106-9}}.

\bibitem{walters2000}
P.~Walters, An Introduction to Ergodic Theory, Springer Verlag, New York, 2000.

\bibitem{khinchin1957}
A.~I. Khinchin, Mathematical Foundations of Information Theory, Dover, New
  York, 1957.

\bibitem{Cover2006}
T.~M. Cover, J.~A. Thomas, Elements of Information Theory, 2nd Edition, John
  Wiley \& Sons, Inc., Hoboken, NJ, 2005.

\bibitem{planck1900}
M.~Planck, Zur theorie des gesetzes der energieverteilung im normalspectrum,
  Verhandlungen der Deutschen Physikalischen Gesellschaft 2 (1900) 237--245.

\bibitem{csiszar2008}
I.~Csiszár, Axiomatic characterization of information measures, Entropy 10
  (2008) 261--273.
\newblock \href {https://doi.org/10.3390/e10030261}
  {\path{doi:10.3390/e10030261}}.

\bibitem{Amigo2022}
J.~M. Amig\'{o}, K.~Keller, V.~Unakafova, On entropy, entropy-like quantities,
  and applications, in: W.~Freeden, M.~Nashed (Eds.), Frontiers in Entropy
  Across the Disciplines, World Scientific, 2022, Ch.~8.
\newblock \href {https://doi.org/10.1142/12920} {\path{doi:10.1142/12920}}.

\bibitem{renyi1961entropy}
A.~R{\'e}nyi, On measures of entropy and information, in: Proceedings of the
  4th Berkeley Symposium on Mathematical Statistics and Probability, Vol.~1,
  1961, pp. 547--561.

\bibitem{Amigo2018}
J.~M. Amigó, S.~G. Balogh, S.~Hernández, A brief review of generalized
  entropies, Entropy 20~(11) (2018) 813.
\newblock \href {https://doi.org/10.3390/e20110813}
  {\path{doi:10.3390/e20110813}}.

\bibitem{HCentropy}
J.~Havrda, F.~Charvát,
  \href{https://www.kybernetika.cz/content/1967/1/30/paper.pdf}{Quantification
  method of classification processes}, Kybernetika 3~(1) (1967) 30--35.
\newline\urlprefix\url{https://www.kybernetika.cz/content/1967/1/30/paper.pdf}

\bibitem{Tsallis1988}
C.~Tsallis, Possible generalization of boltzmann-gibbs statistics, Journal of
  Statistical Physics 52~(1) (1988) 479--487.
\newblock \href {https://doi.org/10.1007/BF01016429}
  {\path{doi:10.1007/BF01016429}}.

\bibitem{Katok2007}
A.~Katok, Fifty years of entropy in dynamics: 1958–2007, Journal of Modern
  Dynamics 1 (2007) 545--596.
\newblock \href {https://doi.org/10.3934/jmd.2007.1.545}
  {\path{doi:10.3934/jmd.2007.1.545}}.

\bibitem{Ribeiro2021}
M.~Ribeiro, T.~Henriques, L.~Castro, A.~Souto, L.~Antunes, C.~Costa-Santos,
  A.~Teixeira, The entropy universe, Entropy 23~(2) (2021) 222.
\newblock \href {https://doi.org/10.3390/e23020222}
  {\path{doi:10.3390/e23020222}}.

\bibitem{madalena2005}
M.~Costa, A.~L. Goldberger, C.~K. Peng, Multiscale entropy analysis of
  biological signals., Physical Review E 71 (2005) 021906.
\newblock \href {https://doi.org/10.1103/physreve.71.021906}
  {\path{doi:10.1103/physreve.71.021906}}.

\bibitem{pincusAppox1991}
J.~M. Pincus, Approximate entropy as a measure of system complexity.,
  Proceedings of the National Academy of Sciences 88~(6) (1991) 2297--2301.
\newblock \href {https://doi.org/10.1073/pnas.88.6.2297}
  {\path{doi:10.1073/pnas.88.6.2297}}.

\bibitem{DelgadoBonal2019}
A.~Delgado-Bonal, A.~Marshak, Approximate entropy and sample entropy: A
  comprehensive tutorial, Entropy 21~(6) (2019) 541.
\newblock \href {https://doi.org/10.3390/e21060541}
  {\path{doi:10.3390/e21060541}}.

\bibitem{hornero2009}
R.~Hornero, D.~Abásolo, J.~Escudero, C.~Gómez, Nonlinear analysis of
  electroencephalogram and magnetoencephalogram recordings in patients with
  alzheimer’s disease, Philosophical Transactions of the Royal Society of
  London, Series A 367~(1887) (2009) 317--336.
\newblock \href {https://doi.org/10.1098/rsta.2008.0197}
  {\path{doi:10.1098/rsta.2008.0197}}.

\bibitem{morabito2012}
F.~C. Morabito, D.~Labate, F.~La~Foresta, A.~Bramanti, G.~Morabito,
  I.~Palamara, Multivariate multi-scale permutation entropy for complexity
  analysis of alzheimer’s disease eeg, Entropy 14~(7) (2012) 1186--1202.
\newblock \href {https://doi.org/10.3390/e14071186}
  {\path{doi:10.3390/e14071186}}.

\bibitem{liang2015}
Z.~Liang, Y.~Wang, X.~Sun, D.~Li, L.~J. Voss, J.~W. Sleigh, S.~Hagihira, X.~Li,
  Eeg entropy measures in anesthesia, Frontiers in Computational Neuroscience 9
  (2015) 16.
\newblock \href {https://doi.org/10.3389/fncom.2015.00016}
  {\path{doi:10.3389/fncom.2015.00016}}.

\bibitem{patel2021}
P.~R. Patel, R.~N. Annavarapu, Eeg-based human emotion recognition using
  entropy as a feature extraction measure, Brain Informatics 8~(1) (2021) 20.
\newblock \href {https://doi.org/10.1186/s40708-021-00141-5}
  {\path{doi:10.1186/s40708-021-00141-5}}.

\bibitem{kannathal2005}
N.~Kannathal, M.~L. Choo, U.~R. Acharya, P.~K. Sadasivan, Entropies for
  detection of epilepsy in eeg, Computer Methods and Programs in Biomedicine
  80~(3) (2005) 187--194.
\newblock \href {https://doi.org/10.1016/j.cmpb.2005.06.012}
  {\path{doi:10.1016/j.cmpb.2005.06.012}}.

\bibitem{srinivasan2007}
V.~Srinivasan, C.~Eswaran, N.~Sriraam, Approximate entropy-based epileptic eeg
  detection using artificial neural networks, IEEE Transactions on Information
  Technology in Biomedicine 11 (2007) 288--295.
\newblock \href {https://doi.org/10.1109/TITB.2006.884369}
  {\path{doi:10.1109/TITB.2006.884369}}.

\bibitem{jouny2012}
C.~C. Jouny, G.~K. Bergey, Characterization of early partial seizure onset:
  Frequency, complexity and entropy, Clinical Neurophysiology 123~(4) (2012)
  658--669.
\newblock \href {https://doi.org/10.1016/j.clinph.2011.08.003}
  {\path{doi:10.1016/j.clinph.2011.08.003}}.

\bibitem{richman2000}
J.~Richman, J.~R. Moorman, Physiological time-series analysis using approximate
  entropy and sample entropy, American Journal of Physiology - Heart and
  Circulatory Physiology 278~(6) (2000) H2039--H2049.
\newblock \href {https://doi.org/10.1152/ajpheart.2000.278.6.H2039}
  {\path{doi:10.1152/ajpheart.2000.278.6.H2039}}.

\bibitem{acharya2005}
U.~R. Acharya, O.~Faust, N.~Kannathal, T.~Chua, S.~Laxminarayan, Non-linear
  analysis of eeg signals at various sleep stages, Computer Methods and
  Programs in Biomedicine 80~(1) (2005) 37--45.
\newblock \href {https://doi.org/10.1016/j.cmpb.2005.06.011}
  {\path{doi:10.1016/j.cmpb.2005.06.011}}.

\bibitem{Manis2017}
G.~Manis, M.~Aktaruzzaman, R.~Sassi, Bubble entropy: An entropy almost free of
  parameters, IEEE Transactions on Biomedical Engineering 64~(11) (2017)
  2711--2718.
\newblock \href {https://doi.org/10.1109/TBME.2017.2664105}
  {\path{doi:10.1109/TBME.2017.2664105}}.

\bibitem{Manis2021}
G.~Manis, M.~Bodini, M.~W. Rivolta, R.~Sassi, A two-steps-ahead estimator for
  bubble entropy, Entropy 23~(6) (2021) 761.
\newblock \href {https://doi.org/10.3390/e23060761}
  {\path{doi:10.3390/e23060761}}.

\bibitem{Gong2022}
J.~Gong, X.~Yang, H.~Wang, J.~Shen, W.~Liu, F.~Zhou, Coordinated method fusing
  improved bubble entropy and artificial gorilla troops optimizer optimized
  kelm for rolling bearing fault diagnosis, Applied Acoustics 195 (2022)
  108844.
\newblock \href {https://doi.org/10.1016/j.apacoust.2022.108844}
  {\path{doi:10.1016/j.apacoust.2022.108844}}.

\bibitem{Gong2023}
J.~Gong, X.~Yang, K.~Qian, Z.~Chen, T.~Han, Application of improved bubble
  entropy and machine learning in the adaptive diagnosis of rotating machinery
  faults, Alexandria Engineering Journal 80 (2023) 22--40.
\newblock \href {https://doi.org/10.1016/j.aej.2023.08.006}
  {\path{doi:10.1016/j.aej.2023.08.006}}.

\bibitem{Jiang2023}
X.~Jiang, Y.~Yi, J.~Wu, Analysis of the synergistic complementarity between
  bubble entropy and dispersion entropy in the application of feature
  extraction, Frontiers in Physics 11 (2023) 1163767.
\newblock \href {https://doi.org/10.3389/fphy.2023.1163767}
  {\path{doi:10.3389/fphy.2023.1163767}}.

\bibitem{Bishop2006}
C.~M. Bishop, \href{https://link.springer.com/book/9780387310732}{Pattern
  Recognition and Machine Learning}, Springer Science + Business Media, 2006.
\newline\urlprefix\url{https://link.springer.com/book/9780387310732}

\bibitem{Spindelbock2021}
T.~Spindelböck, S.~Ranftl, W.~von~der Linden, Cross-entropy learning for
  aortic pathology classification of artificial multi-sensor impedance
  cardiography signals, Entropy 23~(12) (2021) 1661.
\newblock \href {https://doi.org/10.3390/e23121661}
  {\path{doi:10.3390/e23121661}}.

\bibitem{Li2021}
P.~Li, X.~He, D.~Song, Z.~Ding, M.~Qiao, X.~Cheng, R.~Li, Improved categorical
  cross-entropy loss for training deep neural networks with noisy labels, in:
  H.~Ma, L.~Wang, C.~Zhang, F.~Wu, T.~Tan, Y.~Wang, J.~Lai, Y.~Zhao (Eds.),
  Pattern Recognition and Computer Vision, Springer International Publishing,
  2021, pp. 78--89.
\newblock \href {https://doi.org/10.1007/978-3-030-88013-2_7}
  {\path{doi:10.1007/978-3-030-88013-2_7}}.

\bibitem{Farebrother2024}
J.~Farebrother, J.~Orbay, Q.~Vuong, A.~A. Taïga, Y.~Chebotar, T.~Xiao,
  A.~Irpan, S.~Levine, P.~S. Castro, A.~Faust, A.~Kumar, R.~Agarwal, Stop
  regressing: Training value functions via classification for scalable deep rl,
  arXiv:2403.03950 (2024).
\newblock \href {https://doi.org/10.48550/arXiv.2403.03950}
  {\path{doi:10.48550/arXiv.2403.03950}}.

\bibitem{Arazo2020}
E.~Arazo, D.~Ortego, P.~Albert, N.~E. O'Connor, K.~McGuinness, Pseudo-labeling
  and confirmation bias in deep semi-supervised learning, arXiv:1908.02983
  (2020).
\newblock \href {https://doi.org/10.48550/arXiv.1908.02983}
  {\path{doi:10.48550/arXiv.1908.02983}}.

\bibitem{Mao2023}
A.~Mao, M.~Mohri, Y.~Zhong, Cross-entropy loss functions: theoretical analysis
  and applications, arXiv:2304.07288 (2023).
\newblock \href {https://doi.org/10.48550/arXiv.2304.07288}
  {\path{doi:10.48550/arXiv.2304.07288}}.

\bibitem{Berrada2018}
L.~Berrada, A.~Zisserman, M.~P. Kumar, Smooth loss functions for deep top-k
  classification, arXiv:1802.07595 (2018).
\newblock \href {https://doi.org/10.48550/arXiv.1802.07595}
  {\path{doi:10.48550/arXiv.1802.07595}}.

\bibitem{Wang2022}
Z.~Wang, Q.~Zhu, A cross-entropy based feature selection method for binary
  valued data classification, in: A.~Abraham, N.~Gandhi, T.~Hanne, T.~Hong,
  T.~Nogueira~Rios, W.~Ding (Eds.), Intelligent Systems Design and
  Applications, Springer International Publishing, 2022, pp. 1406--1416.
\newblock \href {https://doi.org/10.1007/978-3-030-96308-8_130}
  {\path{doi:10.1007/978-3-030-96308-8_130}}.

\bibitem{Kim2007}
S.~C. Kim, T.~J. Kang, Texture classification and segmentation using wavelet
  packet frame and gaussian mixture model, Pattern Recognition 40~(4) (2007)
  1207--1221.
\newblock \href {https://doi.org/10.1016/j.patcog.2006.09.012}
  {\path{doi:10.1016/j.patcog.2006.09.012}}.

\bibitem{Bruch2019}
S.~Bruch, {An Alternative Cross Entropy Loss for Learning-to-Rank},
  arXiv:1911.09798 (2021).
\newblock \href {https://doi.org/10.48550/arXiv.1911.09798}
  {\path{doi:10.48550/arXiv.1911.09798}}.

\bibitem{Santosa2015}
B.~Santosa, Multiclass classification with cross entropy-support vector
  machines, Procedia Computer Science 72 (2015) 345--352.
\newblock \href {https://doi.org/10.1016/j.procs.2015.12.149}
  {\path{doi:10.1016/j.procs.2015.12.149}}.

\bibitem{Smieja2017}
M.~Śmieja, B.~C. Geiger, Semi-supervised cross-entropy clustering with
  information bottleneck constraint, Information Sciences 421 (2017) 254–271.
\newblock \href {https://doi.org/10.48550/arXiv.1705.01601}
  {\path{doi:10.48550/arXiv.1705.01601}}.

\bibitem{Orchard2012}
M.~E. Orchard, B.~Olivares, M.~Cerda, J.~F. Silva, Anomaly detection based on
  information-theoretic measures and particle filtering algorithms, Annual
  Conference of the Prognostics and Health Management (PHM) Society 4(1)
  (2012).
\newblock \href {https://doi.org/10.36001/phmconf.2012.v4i1.2113}
  {\path{doi:10.36001/phmconf.2012.v4i1.2113}}.

\bibitem{qu2020}
Y.~Qu, R.~Li, A.~Deng, C.~Shang, Q.~Shen, Non-unique decision differential
  entropy-based feature selection, Neurocomputing 393 (2020) 187--193.
\newblock \href {https://doi.org/10.1016/j.neucom.2018.10.112}
  {\path{doi:10.1016/j.neucom.2018.10.112}}.

\bibitem{e23070856}
E.~Grassucci, D.~Comminiello, A.~Uncini, An information-theoretic perspective
  on proper quaternion variational autoencoders, Entropy 23~(7) (2021) 856.
\newblock \href {https://doi.org/10.3390/e23070856}
  {\path{doi:10.3390/e23070856}}.

\bibitem{e20100750}
J.~Gibson, Entropy power, autoregressive models, and mutual information,
  Entropy 20~(10) (2018) 750.
\newblock \href {https://doi.org/10.3390/e20100750}
  {\path{doi:10.3390/e20100750}}.

\bibitem{Robin2023}
S.~Robin, L.~Scrucca, Mixture-based estimation of entropy, Computational
  Statistics \& Data Analysis 177 (2023) 107582.
\newblock \href {https://doi.org/10.1016/j.csda.2022.107582}
  {\path{doi:10.1016/j.csda.2022.107582}}.

\bibitem{7434608}
M.~Rostaghi, H.~Azami, Dispersion entropy: A measure for time-series analysis,
  IEEE Signal Processing Letters 23~(5) (2016) 610--614.
\newblock \href {https://doi.org/10.1109/LSP.2016.2542881}
  {\path{doi:10.1109/LSP.2016.2542881}}.

\bibitem{e20030210}
H.~Azami, J.~Escudero, Amplitude- and fluctuation-based dispersion entropy,
  Entropy 20~(3) (2018) 210.
\newblock \href {https://doi.org/10.3390/e20030210}
  {\path{doi:10.3390/e20030210}}.

\bibitem{e25111494}
M.~Rostaghi, M.~M. Khatibi, M.~R. Ashory, H.~Azami, Refined composite
  multiscale fuzzy dispersion entropy and its applications to bearing fault
  diagnosis, Entropy 25~(11) (2023) 1494.
\newblock \href {https://doi.org/10.3390/e25111494}
  {\path{doi:10.3390/e25111494}}.

\bibitem{e23101303}
R.~Furlong, M.~Hilal, V.~O’Brien, A.~Humeau-Heurtier, Parameter analysis of
  multiscale two-dimensional fuzzy and dispersion entropy measures using
  machine learning classification, Entropy 23~(10) (2021) 1303.
\newblock \href {https://doi.org/10.3390/e23101303}
  {\path{doi:10.3390/e23101303}}.

\bibitem{Hu2024}
B.~Hu, Y.~Wang, J.~Mu, A new fractional fuzzy dispersion entropy and its
  application in muscle fatigue detection, Mathematical Biosciences and
  Engineering 21~(1) (2024) 144--169.
\newblock \href {https://doi.org/10.3934/mbe.2024007}
  {\path{doi:10.3934/mbe.2024007}}.

\bibitem{e23121567}
R.~Dhandapani, I.~Mitiche, S.~McMeekin, V.~S. Mallela, G.~Morison, Enhanced
  partial discharge signal denoising using dispersion entropy optimized
  variational mode decomposition, Entropy 23~(12) (2021) 1567.
\newblock \href {https://doi.org/10.3390/e23121567}
  {\path{doi:10.3390/e23121567}}.

\bibitem{electronics8060597}
G.~Li, Z.~Yang, H.~Yang, A denoising method of ship radiated noise signal based
  on modified ceemdan, dispersion entropy, and interval thresholding,
  Electronics 8~(6) (2019) 597.
\newblock \href {https://doi.org/10.3390/electronics8060597}
  {\path{doi:10.3390/electronics8060597}}.

\bibitem{arXiv:2405.00518}
J.~S. Fabila-Carrasco, C.~Tan, J.~Escudero, Graph-based multivariate multiscale
  dispersion entropy: Efficient implementation and applications to real-world
  network data, arXiv:2405.00518 (2024).
\newblock \href {https://doi.org/10.48550/arXiv.2405.00518}
  {\path{doi:10.48550/arXiv.2405.00518}}.

\bibitem{sym10110623}
H.~Ge, G.~Chen, H.~Yu, H.~Chen, F.~An, Theoretical analysis of empirical mode
  decomposition, Symmetry 10~(11) (2018) 623.
\newblock \href {https://doi.org/10.3390/sym10110623}
  {\path{doi:10.3390/sym10110623}}.

\bibitem{LIU2018169}
C.~Liu, L.~Zhu, C.~Ni, Chatter detection in milling process based on vmd and
  energy entropy, Mechanical Systems and Signal Processing 105 (2018) 169--182.
\newblock \href {https://doi.org/10.1016/j.ymssp.2017.11.046}
  {\path{doi:10.1016/j.ymssp.2017.11.046}}.

\bibitem{GAO2022110417}
Z.~Gao, Y.~Liu, Q.~Wang, J.~Wang, Y.~Luo, Ensemble empirical mode decomposition
  energy moment entropy and enhanced long short-term memory for early fault
  prediction of bearing, Measurement 188 (2022) 110417.
\newblock \href {https://doi.org/10.1016/j.measurement.2021.110417}
  {\path{doi:10.1016/j.measurement.2021.110417}}.

\bibitem{YU2006269}
Y.~Yu, Y.~Dejie, C.~Junsheng, A roller bearing fault diagnosis method based on
  emd energy entropy and ann, Journal of Sound and Vibration 294~(1) (2006)
  269--277.
\newblock \href {https://doi.org/10.1016/j.jsv.2005.11.002}
  {\path{doi:10.1016/j.jsv.2005.11.002}}.

\bibitem{doi:10.1142/S0219519423400638}
Z.~Yang, S.~Luo, P.~Zhong, R.~Chen, C.~Pan, K.~Li, An {EMD} and {IMF} {Energy}
  entropy-based optimized feature extraction and classification scheme for
  single trial eeg signal, Journal of Mechanics in Medicine and Biology 23~(08)
  (2023) 2340063.
\newblock \href {https://doi.org/10.1142/S0219519423400638}
  {\path{doi:10.1142/S0219519423400638}}.

\bibitem{zhu2021short}
G.~Zhu, S.~Peng, Y.~Lao, Q.~Su, Q.~Sun, Short-term electricity consumption
  forecasting based on the emd-fbprophet-lstm method, Mathematical Problems in
  Engineering 2021~(1) (2021) 6613604.
\newblock \href {https://doi.org/10.1155/2021/6613604}
  {\path{doi:10.1155/2021/6613604}}.

\bibitem{gao2019analysis}
J.~Gao, P.~Shang, Analysis of complex time series based on emd energy entropy
  plane, Nonlinear Dynamics 96~(1) (2019) 465--482.
\newblock \href {https://doi.org/10.1007/s11071-019-04800-5}
  {\path{doi:10.1007/s11071-019-04800-5}}.

\bibitem{headrick2019lectures}
M.~Headrick, Lectures on entanglement entropy in field theory and holography,
  arXiv:1907.08126 (2019).
\newblock \href {https://doi.org/10.48550/arXiv.1907.08126}
  {\path{doi:10.48550/arXiv.1907.08126}}.

\bibitem{rieger2024sample}
M.~Rieger, M.~Reh, M.~Gärtner, Sample-efficient estimation of entanglement
  entropy through supervised learning, Physical Review A 109~(1) (2024) 012403.
\newblock \href {https://doi.org/10.1103/PhysRevA.109.012403}
  {\path{doi:10.1103/PhysRevA.109.012403}}.

\bibitem{liu2021entanglement}
Y.~Liu, W.~J. Li, X.~Zhang, M.~Lewenstein, G.~Su, S.~J. Ran, Entanglement-based
  feature extraction by tensor network machine learning, Frontiers in Applied
  Mathematics and Statistics 7 (2021) 716044.
\newblock \href {https://doi.org/10.3389/fams.2021.716044}
  {\path{doi:10.3389/fams.2021.716044}}.

\bibitem{lin2022quantifying}
X.~Lin, Z.~Chen, Z.~Wei, Quantifying unknown quantum entanglement via a hybrid
  quantum-classical machine learning framework, Physical Review A 107 (2023)
  062409.
\newblock \href {https://doi.org/10.1103/PhysRevA.107.062409}
  {\path{doi:10.1103/PhysRevA.107.062409}}.

\bibitem{ABDALLAH2012275}
S.~A. Abdallah, M.~D. Plumbley, A measure of statistical complexity based on
  predictive information with application to finite spin systems, Physics
  Letters A 376~(4) (2012) 275--281.
\newblock \href {https://doi.org/10.1016/j.physleta.2011.10.066}
  {\path{doi:10.1016/j.physleta.2011.10.066}}.

\bibitem{crutchfield1983}
J.~P. Crutchfield, N.~H. Packard, Symbolic dynamics of noisy chaos., Physica D
  7 (1983) 201–223.
\newblock \href {https://doi.org/10.1016/0167-2789(83)90127-6}
  {\path{doi:10.1016/0167-2789(83)90127-6}}.

\bibitem{bardera2009}
A.~Bardera, I.~Boada, M.~Feixas, M.~Sbert, Image segmentation using excess
  entropy, Journal of Signal Processing Systems 54~(1-3) (2009) 205--214.
\newblock \href {https://doi.org/10.1007/s11265-008-0194-6}
  {\path{doi:10.1007/s11265-008-0194-6}}.

\bibitem{nir2020machine}
A.~Nir, E.~Sela, R.~Beck, Y.~Bar-Sinai, Machine-learning iterative calculation
  of entropy for {Physical} systems, Proceedings of the National Academy of
  Sciences 117~(48) (2020) 30234--30240.
\newblock \href {https://doi.org/10.1073/pnas.2017042117}
  {\path{doi:10.1073/pnas.2017042117}}.

\bibitem{belghazi2021minemutualinformationneural}
M.~I. Belghazi, A.~Baratin, S.~Rajeswar, S.~Ozair, Y.~Bengio, A.~Courville,
  R.~D. Hjelm, {MINE: Mutual Information Neural Estimation}, arXiv:1801.04062
  (2021).
\newblock \href {https://doi.org/10.48550/arXiv.1801.04062}
  {\path{doi:10.48550/arXiv.1801.04062}}.

\bibitem{math11112448}
X.~Xiang, J.~Zhou, An excess entropy approach to classify long-term and
  short-term memory stationary time series, Mathematics 11~(11) (2023) 2448.
\newblock \href {https://doi.org/10.3390/math11112448}
  {\path{doi:10.3390/math11112448}}.

\bibitem{Chen2023}
Y.~Chen, J.~Chen, Y.~Qiang, Z.~Yuan, J.~Yang, Refined composite moving average
  fluctuation dispersion entropy and its application on rolling bearing fault
  diagnosis, Review of Scientific Instruments 94~(10) (2023) 105110.
\newblock \href {https://doi.org/10.1063/5.0165430}
  {\path{doi:10.1063/5.0165430}}.

\bibitem{machines11010047}
H.~Su, Z.~Wang, Y.~Cai, J.~Ding, X.~Wang, L.~Yao, Refined composite multiscale
  fluctuation dispersion entropy and supervised manifold mapping for planetary
  gearbox fault diagnosis, Machines 11~(1) (2023) 47.
\newblock \href {https://doi.org/10.3390/machines11010047}
  {\path{doi:10.3390/machines11010047}}.

\bibitem{Zhou2021}
F.~Zhou, J.~Han, X.~Yang, Multivariate hierarchical multiscale fluctuation
  dispersion entropy: Applications to fault diagnosis of rotating machinery,
  Applied Acoustics 182 (2021) 108271.
\newblock \href {https://doi.org/10.1016/j.apacoust.2021.108271}
  {\path{doi:10.1016/j.apacoust.2021.108271}}.

\bibitem{Li20231}
Z.~Li, T.~Lan, Z.~Li, P.~Gao, Exploring relationships between boltzmann entropy
  of images and building classification accuracy in land cover mapping, Entropy
  25~(8) (2023) 1182.
\newblock \href {https://doi.org/10.3390/e25081182}
  {\path{doi:10.3390/e25081182}}.

\bibitem{e23121611}
G.~Baldini, J.~M. Chareau, F.~Bonavitacola, Spectrum sensing implemented with
  improved fluctuation-based dispersion entropy and machine learning, Entropy
  23~(12) (2021) 1611.
\newblock \href {https://doi.org/10.3390/e23121611}
  {\path{doi:10.3390/e23121611}}.

\bibitem{azami2019multiscalefluctuationbaseddispersionentropy}
H.~Azami, S.~E. Arnold, S.~Sanei, Z.~Chang, G.~Sapiro, J.~Escudero, A.~S.
  Gupta, Multiscale fluctuation-based dispersion entropy and its applications
  to neurological diseases, ArXiv:1902.10825 (2019).
\newblock \href {https://doi.org/10.48550/arXiv.1902.10825}
  {\path{doi:10.48550/arXiv.1902.10825}}.

\bibitem{Jiao2021}
S.~Jiao, B.~Geng, Y.~Li, Q.~Zhang, Q.~Wang, Fluctuation-based reverse
  dispersion entropy and its applications to signal classification, Applied
  Acoustics 175 (2021) 107857.
\newblock \href {https://doi.org/10.1016/j.apacoust.2020.107857}
  {\path{doi:10.1016/j.apacoust.2020.107857}}.

\bibitem{Amigo2004}
J.~M. Amigó, J.~Szczepanski, E.~Wajnryb, M.~V. Sanchez-Vives, Estimating the
  entropy of spike trains via lempel-ziv complexity, Neural Computation 16
  (2004) 717--736.
\newblock \href {https://doi.org/10.1162/089976604322860677}
  {\path{doi:10.1162/089976604322860677}}.

\bibitem{everymonotone}
E.~Friedgut, G.~Kalai, Every monotone graph property has a sharp threshold,
  Proceedings of the American Mathematical Society 124~(10) (1999) 2993--3002.
\newblock \href {https://doi.org/10.1090/S0002-9939-96-03732-X}
  {\path{doi:10.1090/S0002-9939-96-03732-X}}.

\bibitem{CHAKRABORTY201692}
S.~Chakraborty, R.~Kulkarni, S.~Lokam, N.~Saurabh, Upper bounds on {Fourier}
  entropy, Theoretical Computer Science 654 (2016) 92--112.
\newblock \href {https://doi.org/10.1016/j.tcs.2016.05.006}
  {\path{doi:10.1016/j.tcs.2016.05.006}}.

\bibitem{10.1007/978-3-642-22006-7_28}
R.~O'Donnell, J.~Wright, Y.~Zhou, The fourier entropy--influence conjecture for
  certain classes of boolean functions, in: L.~Aceto, M.~Henzinger, J.~Sgall
  (Eds.), Automata, Languages and Programming, Lecture Notes in Computer
  Science, vol 6755, Springer Berlin Heidelberg, 2011.
\newblock \href {https://doi.org/10.1007/978-3-642-22006-7_28}
  {\path{doi:10.1007/978-3-642-22006-7_28}}.

\bibitem{9317968}
E.~Kelman, G.~Kindler, N.~Lifshitz, D.~Minzer, M.~Safra, Towards a proof of the
  {Fourier}–entropy conjecture?, in: 2020 IEEE 61st Annual Symposium on
  Foundations of Computer Science (FOCS), 2020, pp. 247--258.
\newblock \href {https://doi.org/10.1109/FOCS46700.2020.00032}
  {\path{doi:10.1109/FOCS46700.2020.00032}}.

\bibitem{330368}
L.~B. Almeida, {The fractional Fourier transform and time-frequency
  representations}, IEEE Transactions on Signal Processing 42~(11) (1994)
  3084--3091.
\newblock \href {https://doi.org/10.1109/78.330368}
  {\path{doi:10.1109/78.330368}}.

\bibitem{8847346}
R.~Tao, X.~Zhao, W.~Li, H.~C. Li, Q.~Du, Hyperspectral anomaly detection by
  {Fractional} {Fourier} entropy, IEEE Journal of Selected Topics in Applied
  Earth Observations and Remote Sensing 12~(12) (2019) 4920--4929.
\newblock \href {https://doi.org/10.1109/JSTARS.2019.2940278}
  {\path{doi:10.1109/JSTARS.2019.2940278}}.

\bibitem{rs14030797}
L.~Zhang, J.~Ma, B.~Cheng, F.~Lin, Fractional fourier transform-based tensor rx
  for hyperspectral anomaly detection, Remote Sensing 14~(3) (2022) 797.
\newblock \href {https://doi.org/10.3390/rs14030797}
  {\path{doi:10.3390/rs14030797}}.

\bibitem{10.1145/3451357}
S.~H. Wang, X.~Zhang, Y.~D. Zhang, {DSSAE}: Deep stacked sparse autoencoder
  analytical model for {COVID-19} diagnosis by {Fractional Fourier} entropy,
  ACM Transactions on Management Information System (TMIS) 13~(1) (2021).
\newblock \href {https://doi.org/10.1145/3451357} {\path{doi:10.1145/3451357}}.

\bibitem{e17127877}
S.~Wang, Y.~Zhang, X.~Yang, P.~Sun, Z.~Dong, A.~Liu, T.~F. Yuan, Pathological
  brain detection by a novel image feature—fractional fourier entropy,
  Entropy 17~(12) (2015) 8278--8296.
\newblock \href {https://doi.org/10.3390/e17127877}
  {\path{doi:10.3390/e17127877}}.

\bibitem{YAN202036}
Y.~Yan, Gingivitis detection by fractional {Fourier} entropy with optimization
  of hidden neurons, International Journal of Cognitive Computing in
  Engineering 1 (2020) 36--44.
\newblock \href {https://doi.org/10.1016/j.ijcce.2020.09.003}
  {\path{doi:10.1016/j.ijcce.2020.09.003}}.

\bibitem{PANAHI2021102863}
F.~Panahi, S.~Rashidi, A.~Sheikhani, Application of fractional {Fourier}
  transform in feature extraction from {Electrocardiogram and Galvanic Skin}
  response for emotion recognition, Biomedical Signal Processing and Control 69
  (2021) 102863.
\newblock \href {https://doi.org/10.1016/j.bspc.2021.102863}
  {\path{doi:10.1016/j.bspc.2021.102863}}.

\bibitem{e18030077}
Y.~Zhang, X.~Yang, C.~Cattani, R.~V. Rao, S.~Wang, P.~Phillips, Tea category
  identification using a novel {Fractional} {Fourier} entropy and jaya
  algorithm, Entropy 18~(3) (2016) 77.
\newblock \href {https://doi.org/10.3390/e18030077}
  {\path{doi:10.3390/e18030077}}.

\bibitem{Zadeh1965}
L.~A. Zadeh, Fuzzy sets, Information and Control 8 (1965) 338--353.
\newblock \href {https://doi.org/10.1016/S0019-9958(65)90241-X}
  {\path{doi:10.1016/S0019-9958(65)90241-X}}.

\bibitem{DeLuca1972}
A.~De~Luca, S.~Termini, A definition of a nonprobabilistic entropy in the
  setting of fuzzy sets theory, Information and Control 20 (1972) 301--312.
\newblock \href {https://doi.org/10.1016/S0019-9958(72)90199-4}
  {\path{doi:10.1016/S0019-9958(72)90199-4}}.

\bibitem{ISHIKAWA1979113}
A.~Ishikawa, H.~Mieno, The fuzzy entropy concept and its application, Fuzzy
  Sets and Systems 2~(2) (1979) 113--123.
\newblock \href {https://doi.org/10.1016/0165-0114(79)90020-4}
  {\path{doi:10.1016/0165-0114(79)90020-4}}.

\bibitem{ChenYu2007}
W.~Chen, Z.~Wang, H.~Xie, W.~Yu, Characterization of surface emg signal based
  on fuzzy entropy, IEEE Transactions on Neural Systems and Rehabilitation
  Engineering 15 (2007) 266--272.
\newblock \href {https://doi.org/10.1109/TNSRE.2007.897025}
  {\path{doi:10.1109/TNSRE.2007.897025}}.

\bibitem{ZHENG2014187}
J.~Zheng, J.~Cheng, Y.~Yang, S.~Luo, A rolling bearing fault diagnosis method
  based on multi-scale fuzzy entropy and variable predictive model-based class
  discrimination, Mechanism and Machine Theory 78 (2014) 187--200.
\newblock \href {https://doi.org/10.1016/j.mechmachtheory.2014.03.014}
  {\path{doi:10.1016/j.mechmachtheory.2014.03.014}}.

\bibitem{e18010019}
D.~Markechová, B.~Riečan, Entropy of fuzzy partitions and entropy of fuzzy
  dynamical systems, Entropy 18~(1) (2016) 19.
\newblock \href {https://doi.org/10.3390/e18010019}
  {\path{doi:10.3390/e18010019}}.

\bibitem{Durso2023}
P.~D’Urso, L.~De~Giovanni, V.~Vitale, Robust {DTW-based} entropy fuzzy
  clustering of time series, Annals of Operations Research (2023).
\newblock \href {https://doi.org/10.1007/s10479-023-05720-9}
  {\path{doi:10.1007/s10479-023-05720-9}}.

\bibitem{e20060424}
F.~Di~Martino, S.~Sessa, Energy and entropy measures of fuzzy relations for
  data analysis, Entropy 20~(6) (2018) 424.
\newblock \href {https://doi.org/10.3390/e20060424}
  {\path{doi:10.3390/e20060424}}.

\bibitem{MotorFault2023}
S.~Aguayo-Tapia, G.~Avalos-Almazan, J.~J. Rangel-Magdaleno, Entropy-based
  methods for motor fault detection: A review, Entropy 26~(4) (2024) 299.
\newblock \href {https://doi.org/10.3390/e26040299}
  {\path{doi:10.3390/e26040299}}.

\bibitem{Jind2014MultiscaleFE}
Z.~Jinde, M.~J. Chen, C.~Junsheng, Y.~Yang,
  \href{https://www.researchgate.net/publication/287480256_Multiscale_fuzzy_entropy_and_its_application_in_rolling_bearing_fault_diagnosis}{Multiscale
  fuzzy entropy and its application in rolling bearing fault diagnosis},
  Zhendong Gongcheng Xuebao/Journal of Vibration Engineering 27 (2014)
  145--151.
\newline\urlprefix\url{https://www.researchgate.net/publication/287480256_Multiscale_fuzzy_entropy_and_its_application_in_rolling_bearing_fault_diagnosis}

\bibitem{KUMAR2023100351}
R.~Kumar, D.~C.~S. Bisht, Picture fuzzy entropy: A novel measure for managing
  uncertainty in multi-criteria decision-making, Decision Analytics Journal 9
  (2023) 100351.
\newblock \href {https://doi.org/10.1016/j.dajour.2023.100351}
  {\path{doi:10.1016/j.dajour.2023.100351}}.

\bibitem{e24111577}
E.~Lhermitte, M.~Hilal, R.~Furlong, V.~O’Brien, A.~Humeau-Heurtier, Deep
  learning and entropy-based texture features for color image classification,
  Entropy 24~(11) (2022) 1577.
\newblock \href {https://doi.org/10.3390/e24111577}
  {\path{doi:10.3390/e24111577}}.

\bibitem{Tan2021}
Z.~Tan, K.~Li, Y.~Wang, An improved cuckoo search algorithm for multilevel
  color image thresholding based on modified fuzzy entropy, Journal of Ambient
  Intelligence and Humanized Computing (2021).
\newblock \href {https://doi.org/10.1007/s12652-021-03001-6}
  {\path{doi:10.1007/s12652-021-03001-6}}.

\bibitem{korner1971}
J.~Korner,
  \href{https://scholar.google.com/citations?view_op=view_citation&hl=en&user=bQ8m_m4AAAAJ&citation_for_view=bQ8m_m4AAAAJ:dhFuZR0502QC}{Coding
  of an information source having ambiguous alphabet and the entropy of
  graphs}, Transactions of the 6th Prague conference on Information Theory
  (1971) 411--425.
\newline\urlprefix\url{https://scholar.google.com/citations?view_op=view_citation&hl=en&user=bQ8m_m4AAAAJ&citation_for_view=bQ8m_m4AAAAJ:dhFuZR0502QC}

\bibitem{simonyi1995}
G.~Simonyi,
  \href{http://scholar.google.com/scholar.bib?q=info:vh6sRffaxYEJ:scholar.google.com/&output=citation&hl=en&as_sdt=0,5&as_vis=1&ct=citation&cd=0}{Graph
  entropy: A survey}, Combinatorial Optimization 20 (1995) 399--441.
\newblock \href {https://doi.org/10.1090/dimacs/020/08}
  {\path{doi:10.1090/dimacs/020/08}}.
\newline\urlprefix\url{http://scholar.google.com/scholar.bib?q=info:vh6sRffaxYEJ:scholar.google.com/&output=citation&hl=en&as_sdt=0,5&as_vis=1&ct=citation&cd=0}

\bibitem{luque2010horizontalvisibilitygraphsexact}
B.~Luque, L.~Lacasa, F.~Ballesteros, J.~Luque, Horizontal visibility graphs:
  exact results for random time series, Physical Review E 80 (2009) 046103.
\newblock \href {https://doi.org/10.1103/PhysRevE.80.046103}
  {\path{doi:10.1103/PhysRevE.80.046103}}.

\bibitem{LACASA201835}
L.~Lacasa, W.~Just, Visibility graphs and symbolic dynamics, Physica D:
  Nonlinear Phenomena 374-375 (2018) 35--44.
\newblock \href {https://doi.org/10.1016/j.physd.2018.04.001}
  {\path{doi:10.1016/j.physd.2018.04.001}}.

\bibitem{harangi2023conditionalgraphentropyalternating}
V.~Harangi, X.~Niu, B.~Bai, Conditional graph entropy as an alternating
  minimization problem, arXiv:2209.00283 (2023).
\newblock \href {https://doi.org/10.48550/arXiv.2209.00283}
  {\path{doi:10.48550/arXiv.2209.00283}}.

\bibitem{wu2022structuralentropyguidedgraph}
J.~Wu, X.~Chen, K.~Xu, S.~Li, Structural entropy guided graph hierarchical
  pooling, arXiv:2206.13510 (2022).
\newblock \href {https://doi.org/10.48550/arXiv.2206.13510}
  {\path{doi:10.48550/arXiv.2206.13510}}.

\bibitem{juhnkekubitzke2021countinghorizontalvisibilitygraphs}
M.~Juhnke-Kubitzke, D.~Köhne, J.~Schmidt, Counting horizontal visibility
  graphs, arXiv:2111.02723 (2021).
\newblock \href {https://doi.org/10.48550/arXiv.2111.02723}
  {\path{doi:10.48550/arXiv.2111.02723}}.

\bibitem{luo2021graphentropyguidednode}
G.~Luo, J.~Li, J.~Su, H.~Peng, C.~Yang, L.~Sun, P.~S. Yu, L.~He, Graph entropy
  guided node embedding dimension selection for graph neural networks,
  arXiv:2105.03178 (2021).
\newblock \href {https://doi.org/10.48550/arXiv.2105.03178}
  {\path{doi:10.48550/arXiv.2105.03178}}.

\bibitem{Zhua2014a}
G.~Zhu, Y.~Li, P.~Wen, Analysis of alcoholic {EEG} signals based on horizontal
  visibility graph entropy, Brain Informatics 1~(1) (2014) 19--25.
\newblock \href {https://doi.org/10.1007/s40708-014-0003-x}
  {\path{doi:10.1007/s40708-014-0003-x}}.

\bibitem{YU2017249}
M.~Yu, A.~Hillebrand, A.~Gouw, C.~J. Stam, {Horizontal visibility graph
  transfer entropy (HVG-TE): A novel metric to characterize directed
  connectivity in large-scale brain networks}, NeuroImage 156 (2017) 249--264.
\newblock \href {https://doi.org/10.1016/j.neuroimage.2017.05.047}
  {\path{doi:10.1016/j.neuroimage.2017.05.047}}.

\bibitem{Chen2010}
T.~Chen, B.~Vemuri, A.~Rangarajan, S.~J. Eisenschenk, {Group-Wise Point-Set
  Registration Using a Novel CDF-Based Havrda-Charvát Divergence},
  International Journal of Computer Vision 86~(1) (2010) 111--124.
\newblock \href {https://doi.org/10.1007/s11263-009-0261-x}
  {\path{doi:10.1007/s11263-009-0261-x}}.

\bibitem{SHI2021125914}
Y.~Shi, Y.~Wu, P.~Shang, {Research on weighted Havrda–Charvat’s entropy in
  financial time series}, Physica A 572 (2021) 125914.
\newblock \href {https://doi.org/10.1016/j.physa.2021.125914}
  {\path{doi:10.1016/j.physa.2021.125914}}.

\bibitem{brochet2021deeplearningusinghavrdacharvat}
T.~Brochet, J.~Lapuyade-Lahorgue, S.~Bougleux, M.~Salaun, S.~Ruan, {Deep
  learning using Havrda-Charvat entropy for classification of pulmonary
  endomicroscopy}, arXiv:2104.05450 (2021).
\newblock \href {https://doi.org/10.48550/arXiv.2104.05450}
  {\path{doi:10.48550/arXiv.2104.05450}}.

\bibitem{e24040436}
T.~Brochet, J.~Lapuyade-Lahorgue, A.~Huat, S.~Thureau, D.~Pasquier, I.~Gardin,
  R.~Modzelewski, D.~Gibon, J.~Thariat, V.~Grégoire, P.~Vera, S.~Ruan, {A
  Quantitative Comparison between Shannon and Tsallis–Havrda–Charvat
  Entropies Applied to Cancer Outcome Prediction}, Entropy 24~(4) (2022) 436.
\newblock \href {https://doi.org/10.3390/e24040436}
  {\path{doi:10.3390/e24040436}}.

\bibitem{4154718}
H.~Amoud, H.~Snoussi, D.~Hewson, M.~Doussot, J.~Duchene, {Intrinsic Mode
  Entropy for Nonlinear Discriminant Analysis}, IEEE Signal Processing Letters
  14~(5) (2007) 297--300.
\newblock \href {https://doi.org/10.1109/LSP.2006.888089}
  {\path{doi:10.1109/LSP.2006.888089}}.

\bibitem{4650350}
K.~V. E., L.~J. Hadjileontiadis, {Intrinsic mode entropy: An enhanced
  classification means for automated Greek Sign Language gesture recognition},
  in: 2008 30th Annual International Conference of the IEEE Engineering in
  Medicine and Biology Society, 2008, pp. 5057--5060.
\newblock \href {https://doi.org/10.1109/IEMBS.2008.4650350}
  {\path{doi:10.1109/IEMBS.2008.4650350}}.

\bibitem{Hu2011}
M.~Hu, H.~Liang, Intrinsic mode entropy based on multivariate empirical mode
  decomposition and its application to neural data analysis, Cognitive
  Neurodynamics 5~(3) (2011) 277--284.
\newblock \href {https://doi.org/10.1007/s11571-011-9159-8}
  {\path{doi:10.1007/s11571-011-9159-8}}.

\bibitem{Hu2017}
M.~Hu, H.~Liang, {Multiscale Entropy: Recent Advances}, Springer International
  Publishing, 2017, pp. 115--138.
\newblock \href {https://doi.org/10.1007/978-3-319-58709-7_4}
  {\path{doi:10.1007/978-3-319-58709-7_4}}.

\bibitem{Amoud2009}
H.~Amoud, H.~Snoussi, D.~Hewson, J.~Duchêne, Intrinsic mode entropy for
  postural steadiness analysis, in: 4th European Conference of the
  International Federation for Medical and Biological Engineering, Vol.~22 of
  IFMBE Proceedings, Springer, Berlin, Heidelberg, 2009, pp. 212--215.
\newblock \href {https://doi.org/10.1007/978-3-540-89208-3_53}
  {\path{doi:10.1007/978-3-540-89208-3_53}}.

\bibitem{PhysRevE.66.056125}
G.~Kaniadakis, Statistical mechanics in the context of special relativity,
  Physical Review E 66~(5) (2002) 056125.
\newblock \href {https://doi.org/10.1103/PhysRevE.66.056125}
  {\path{doi:10.1103/PhysRevE.66.056125}}.

\bibitem{e26050406}
G.~Kaniadakis, {Relativistic Roots of $\kappa$-Entropy}, Entropy 26~(5) (2024)
  406.
\newblock \href {https://doi.org/10.3390/e26050406}
  {\path{doi:10.3390/e26050406}}.

\bibitem{lei2020adaptive}
B.~Lei, J.~I. Fan, {Adaptive Kaniadakis entropy thresholding segmentation
  algorithm based on particle swarm optimization}, Soft Computing 24~(10)
  (2020) 7305--7318.
\newblock \href {https://doi.org/10.1007/s00500-019-04351-2}
  {\path{doi:10.1007/s00500-019-04351-2}}.

\bibitem{10201718}
B.~Jena, M.~K. Naik, R.~Panda, {A novel Kaniadakis entropy-based multilevel
  thresholding using energy curve and Black Widow optimization algorithm with
  Gaussian mutation}, in: 2023 International Conference in Advances in Power,
  Signal, and Information Technology (APSIT), 2023, pp. 86--91.
\newblock \href {https://doi.org/10.1109/APSIT58554.2023.10201718}
  {\path{doi:10.1109/APSIT58554.2023.10201718}}.

\bibitem{e25070990}
S.~L. E.~F. da~Silva, J.~M. de~Araújo, E.~de~la Barra, G.~Corso, {A
  Graph-Space Optimal Transport Approach Based on Kaniadakis $\kappa$-Gaussian
  Distribution for Inverse Problems Related to Wave Propagation}, Entropy
  25~(7) (2023) 990.
\newblock \href {https://doi.org/10.3390/e25070990}
  {\path{doi:10.3390/e25070990}}.

\bibitem{Mekyska_2015}
J.~Mekyska, E.~Janousova, P.~Gomez-Vilda, Z.~Smekal, I.~Rektorova, I.~Eliasova,
  M.~Kostalova, M.~Mrackova, J.~B. Alonso-Hernandez, M.~Faundez-Zanuy,
  K.~López-de Ipiña, Robust and complex approach of pathological speech
  signal analysis, Neurocomputing 167 (2015) 94–111.
\newblock \href {https://doi.org/10.1016/j.neucom.2015.02.085}
  {\path{doi:10.1016/j.neucom.2015.02.085}}.

\bibitem{1527935}
L.~S. Xu, K.~Q. Wang, L.~Wang, Gaussian kernel approximate entropy algorithm
  for analyzing irregularity of time-series, in: 2005 International Conference
  on Machine Learning and Cybernetics, Vol.~9, 2005, pp. 5605--5608.
\newblock \href {https://doi.org/10.1109/ICMLC.2005.1527935}
  {\path{doi:10.1109/ICMLC.2005.1527935}}.

\bibitem{ZaylaaProceed2016}
A.~Zaylaa, S.~Saleh, F.~Karameh, A.~Nahas, Z.~andBouakaz, Cascade of nonlinear
  entropy and statistics to discriminate fetal heart rates, in: Proceedings of
  the 2016 3rd International Conference on Advances in Computational Tools for
  Engineering Applications (ACTEA), IEEE Xplore, 2016, pp. 152--157.
\newblock \href {https://doi.org/10.1109/ACTEA.2016.7560130}
  {\path{doi:10.1109/ACTEA.2016.7560130}}.

\bibitem{OrozcoProceed2013}
J.~R. Orozco-Arroyave, J.~D. Arias-Londono, J.~F. Vargas-Bonilla, E.~Nöth,
  Analysis of speech from people with parkinson's disease through nonlinear
  dynamics, in: T.~Drugman, T.~Dutoit (Eds.), Advances in Nonlinear Speech
  Processing. NOLISP 2013, Lecture Notes in Computer Science, vol 7911,
  Springer Berlin Heidelberg, 2013.
\newblock \href {https://doi.org/10.1007/978-3-642-38847-7_15}
  {\path{doi:10.1007/978-3-642-38847-7_15}}.

\bibitem{kolmogorov1986}
A.~N. Kolmogorov,
  \href{https://web.archive.org/web/20240601000000*/https://www.mathnet.ru/php/archive.phtml?wshow=paper&jrnid=dan&paperid=22922&option_lang=eng}{A
  new metric invariant of transitive dynamical systems and automorphisms of
  lebesgue spaces}, Proceedings of the Steklov Institute of Mathematics 169
  (1986) 97--102.
\newline\urlprefix\url{https://web.archive.org/web/20240601000000*/https://www.mathnet.ru/php/archive.phtml?wshow=paper&jrnid=dan&paperid=22922&option_lang=eng}

\bibitem{YaBPesin_1977}
Y.~B. Pesin, Characteristic lyapunov exponents and smooth ergodic theory,
  Russian Mathematical Surveys 32~(4) (1977) 55.
\newblock \href {https://doi.org/10.1070/rm1977v032n04abeh001639}
  {\path{doi:10.1070/rm1977v032n04abeh001639}}.

\bibitem{SHIOZAWA2024129531}
K.~Shiozawa, I.~Tokuda, {Estimating Kolmogorov-Sinai entropy from time series
  of high-dimensional complex systems}, Physics Letters A 510 (2024) 129531.
\newblock \href {https://doi.org/10.1016/j.physleta.2024.129531}
  {\path{doi:10.1016/j.physleta.2024.129531}}.

\bibitem{Parlitz2016}
U.~Parlitz, Estimating lyapunov exponents from time series, in: C.~Skokos,
  G.~Gottwald, J.~Laskar (Eds.), Chaos Detection and Predictability, Springer,
  2016, Ch.~1.
\newblock \href {https://doi.org/10.1007/978-3-662-48410-4_1}
  {\path{doi:10.1007/978-3-662-48410-4_1}}.

\bibitem{e22121396}
C.~Karmakar, R.~Udhayakumar, M.~Palaniswami, {Entropy Profiling: A
  Reduced-Parametric Measure of Kolmogorov-Sinai Entropy from Short-Term HRV
  Signal}, Entropy 22~(12) (2020) 1396.
\newblock \href {https://doi.org/10.3390/e22121396}
  {\path{doi:10.3390/e22121396}}.

\bibitem{app14062261}
G.~Kiss, P.~Bakucz, {Using Kolmogorov Entropy to Verify the Description
  Completeness of Traffic Dynamics of Highly Autonomous Driving}, Applied
  Sciences 14~(6) (2024) 2261.
\newblock \href {https://doi.org/10.3390/app14062261}
  {\path{doi:10.3390/app14062261}}.

\bibitem{AFTANAS199713}
L.~I. Aftanas, N.~V. Lotova, V.~I. Koshkarov, V.~L. Pokrovskaja, S.~A. Popov,
  V.~P. Makhnev, {Non-linear analysis of emotion EEG: calculation of Kolmogorov
  entropy and the principal Lyapunov exponent}, Neuroscience Letters 226~(1)
  (1997) 13--16.
\newblock \href {https://doi.org/10.1016/S0304-3940(97)00232-2}
  {\path{doi:10.1016/S0304-3940(97)00232-2}}.

\bibitem{Bandt2002}
C.~Bandt, B.~Pompe, Permutation entropy: A natural complexity measure for time
  series, Physical Review Letters 88~(17) (2002) 174102.
\newblock \href {https://doi.org/10.1103/PhysRevLett.88.174102}
  {\path{doi:10.1103/PhysRevLett.88.174102}}.

\bibitem{e2013practical}
M.~Riedl, A.~Müller, N.~Wessel, Practical considerations of permutation
  entropy, The European Physical Journal Special Topics 222~(2) (2013)
  249--262.
\newblock \href {https://doi.org/10.1140/epjst/e2013-01862-7}
  {\path{doi:10.1140/epjst/e2013-01862-7}}.

\bibitem{amigo2010permutation}
J.~M. Amigó, Permutation Complexity in Dynamical Systems, Springer Verlag,
  2010.
\newblock \href {https://doi.org/10.1007/978-3-642-04084-9}
  {\path{doi:10.1007/978-3-642-04084-9}}.

\bibitem{e14081553}
M.~Zanin, L.~Zunino, O.~A. Rosso, D.~Papo, {Permutation Entropy and Its Main
  Biomedical and Econophysics Applications: A Review}, Entropy 14~(8) (2012)
  1553--1577.
\newblock \href {https://doi.org/10.3390/e14081553}
  {\path{doi:10.3390/e14081553}}.

\bibitem{e2013permutationEditorial}
J.~M. Amigó, K.~Keller, J.~Kurths, Recent progress in symbolic dynamics and
  permutation complexity ---ten years of permutation entropy, The European
  Physical Journal Special Topics 222 (2013) 241--247.
\newblock \href {https://doi.org/10.1140/epjst/e2013-01839-6}
  {\path{doi:10.1140/epjst/e2013-01839-6}}.

\bibitem{e2023permutationEditorial}
J.~M. Amigó, O.~A. Rosso, Ordinal methods: Concepts, applications, new
  developments, and challenges---in memory of karsten keller (1961--2022),
  Chaos 33~(8) (2023) 080401.
\newblock \href {https://doi.org/10.1063/5.0167263}
  {\path{doi:10.1063/5.0167263}}.

\bibitem{e17074627}
N.~Mammone, J.~Duun-Henriksen, T.~W. Kjaer, F.~C. Morabito, {Differentiating
  Interictal and Ictal States in Childhood Absence Epilepsy through Permutation
  Rényi Entropy}, Entropy 17~(7) (2015) 4627--4643.
\newblock \href {https://doi.org/10.3390/e17074627}
  {\path{doi:10.3390/e17074627}}.

\bibitem{ZUNINO20086057}
L.~Zunino, D.~G. Pérez, A.~Kowalski, M.~T. Martin, M.~Garavaglia, A.~Plastino,
  O.~A. Rosso, {Fractional Brownian motion, fractional Gaussian noise, and
  Tsallis permutation entropy}, Physica A 387~(24) (2008) 6057--6068.
\newblock \href {https://doi.org/10.1016/j.physa.2008.07.004}
  {\path{doi:10.1016/j.physa.2008.07.004}}.

\bibitem{Stosic_2022}
D.~Stosic, D.~Stosic, T.~Stosic, B.~Stosic, Generalized weighted permutation
  entropy, Chaos 32 (2022) 103105.
\newblock \href {https://doi.org/10.1063/5.0107427}
  {\path{doi:10.1063/5.0107427}}.

\bibitem{Yin2018}
Y.~Yin, K.~Sun, S.~He, {Multiscale permutation Rényi entropy and its
  application for EEG signals}, PLOS ONE 13~(9) (2018) 1--15.
\newblock \href {https://doi.org/10.1371/journal.pone.0202558}
  {\path{doi:10.1371/journal.pone.0202558}}.

\bibitem{li2019multiscale}
C.~Li, P.~Shang, Multiscale tsallis permutation entropy analysis for complex
  physiological time series, Physica A 523 (2019) 10--20.
\newblock \href {https://doi.org/10.1016/j.physa.2019.01.031}
  {\path{doi:10.1016/j.physa.2019.01.031}}.

\bibitem{AZAMI201628}
H.~Azami, J.~Escudero, Improved multiscale permutation entropy for biomedical
  signal analysis: Interpretation and application to electroencephalogram
  recordings, Biomedical Signal Processing and Control 23 (2016) 28--41.
\newblock \href {https://doi.org/10.1016/j.bspc.2015.08.004}
  {\path{doi:10.1016/j.bspc.2015.08.004}}.

\bibitem{doi101142S02181274030}
K.~Keller, H.~Lauffer, Symbolic analysis of high-dimensional time series,
  International Journal of Bifurcation and Chaos 13~(09) (2003) 2657--2668.
\newblock \href {https://doi.org/10.1142/S0218127403008168}
  {\path{doi:10.1142/S0218127403008168}}.

\bibitem{e19030134}
K.~Keller, T.~Mangold, L.~Stolz, J.~Werner, Permutation entropy: New ideas and
  challenges, Entropy 19~(3) (2017) 134.
\newblock \href {https://doi.org/10.3390/e19030134}
  {\path{doi:10.3390/e19030134}}.

\bibitem{10106350209206}
L.~G. J.~M. Voltarelli, A.~A.~B. Pessa, L.~Zunino, R.~S. Zola, E.~K. Lenzi,
  M.~Perc, H.~V. Ribeiro, Characterizing unstructured data with the nearest
  neighbor permutation entropy, Chaos 34~(5) (2024) 053130.
\newblock \href {https://doi.org/10.1063/5.0209206}
  {\path{doi:10.1063/5.0209206}}.

\bibitem{Graff2023}
P.~Pilarczyk, G.~Graff, J.~M. Amigó, K.~Tessmer, K.~Narkiewicz, B.~Graff,
  Differentiating patients with obstructive sleep apnea from healthy controls
  based on heart rate-blood pressure coupling quantified by entropy-based
  indices, Chaos 33 (2023) 103140.
\newblock \href {https://doi.org/10.1063/5.0158923}
  {\path{doi:10.1063/5.0158923}}.

\bibitem{amigo2008forbidden}
J.~M. Amigó, M.~B. Kennel, Forbidden ordinal patterns in higher dimensional
  dynamics, Physica D 237~(22) (2008) 2893--2899.
\newblock \href {https://doi.org/10.1016/j.physd.2008.05.003}
  {\path{doi:10.1016/j.physd.2008.05.003}}.

\bibitem{e2013permutation}
J.~M. Amigó, K.~Keller, Permutation entropy: One concept, two approaches, The
  European Physical Journal Special Topics 222~(2) (2013) 263--273.
\newblock \href {https://doi.org/10.1140/epjst/e2013-01840-1}
  {\path{doi:10.1140/epjst/e2013-01840-1}}.

\bibitem{CARPI20102020}
L.~C. Carpi, P.~M. Saco, O.~A. Rosso, Missing ordinal patterns in correlated
  noises, Physica A 389~(10) (2010) 2020--2029.
\newblock \href {https://doi.org/10.1016/j.physa.2010.01.030}
  {\path{doi:10.1016/j.physa.2010.01.030}}.

\bibitem{entropyofintervalmaps}
C.~Bandt, G.~Keller, B.~Pompe, Entropy of interval maps via permutations,
  Nonlinearity 15 (2002) 1595.
\newblock \href {https://doi.org/10.1088/0951-7715/15/5/312}
  {\path{doi:10.1088/0951-7715/15/5/312}}.

\bibitem{PhysRevLett99154102}
O.~A. Rosso, H.~A. Larrondo, M.~T. Martin, A.~Plastino, M.~A. Fuentes,
  Distinguishing noise from chaos, Physical Review Letters 99 (2007) 154102.
\newblock \href {https://doi.org/10.1103/PhysRevLett.99.154102}
  {\path{doi:10.1103/PhysRevLett.99.154102}}.

\bibitem{qu2023wind}
Z.~Qu, X.~Hou, W.~Hu, R.~Yang, C.~Ju, Wind power forecasting based on improved
  variational mode decomposition and permutation entropy, Clean Energy 7~(5)
  (2023) 1032--1045.
\newblock \href {https://doi.org/10.1093/ce/zkad043}
  {\path{doi:10.1093/ce/zkad043}}.

\bibitem{CitiGuffantiMainardi2014}
L.~Citi, G.~G., L.~Mainardi,
  \href{https://ieeexplore.ieee.org/document/7043113/keywords#keywords}{{Rank-based
  Multi-Scale Entropy analysis of heart rate variability}}, in: Proceedings of
  the Computing in Cardiology, 2014, pp. 597--600.
\newline\urlprefix\url{https://ieeexplore.ieee.org/document/7043113/keywords#keywords}

\bibitem{GarcheryGranitzer2018}
M.~Garchery, M.~Granitzer, On the influence of categorical features in ranking
  anomalies using mixed data, Procedia Computer Science 126 (2018) 77--86.
\newblock \href {https://doi.org/10.1016/j.procs.2018.07.211}
  {\path{doi:10.1016/j.procs.2018.07.211}}.

\bibitem{KhanAkramSharif2021}
M.~A. Khan, T.~Akram, M.~Sharif, M.~Alhaisoni, T.~Saba, N.~Nawaz, A
  probabilistic segmentation and entropy-rank correlation-based feature
  selection approach for the recognition of fruit diseases, EURASIP Journal on
  Image and Video Processing 2021 (2021) 14.
\newblock \href {https://doi.org/10.1186/s13640-021-00558-2}
  {\path{doi:10.1186/s13640-021-00558-2}}.

\bibitem{HuCheZhangEtAl2012}
Q.~Hu, X.~Che, L.~Zhang, D.~Zhang, M.~Guo, D.~Yu, {Rank Entropy-Based Decision
  Trees for Monotonic Classification}, IEEE Transactions on Knowledge and Data
  Engineering 24~(11) (2012) 2052--2064.
\newblock \href {https://doi.org/10.1109/TKDE.2011.149}
  {\path{doi:10.1109/TKDE.2011.149}}.

\bibitem{LiuGao2023}
S.~Liu, H.~Gao, {The Structure Entropy-Based Node Importance Ranking Method for
  Graph Data}, Entropy 25~(6) (2023) 941.
\newblock \href {https://doi.org/10.3390/e25060941}
  {\path{doi:10.3390/e25060941}}.

\bibitem{McLellanRyanBreneman2011}
M.~R. McLellan, M.~D. Ryan, C.~M. Breneman, Rank order entropy: why one metric
  is not enough, Journal of Chemical Information and Modeling 51~(9) (2011)
  2302--2319.
\newblock \href {https://doi.org/10.1021/ci200170k}
  {\path{doi:10.1021/ci200170k}}.

\bibitem{DiksPanchenko2008}
C.~Diks, V.~Panchenko, {Rank-based Entropy Tests for Serial Independence},
  Studies in Nonlinear Dynamics \& Econometrics 12~(1) (2008).
\newblock \href {https://doi.org/10.2202/1558-3708.1476}
  {\path{doi:10.2202/1558-3708.1476}}.

\bibitem{SunLiSongHong2023}
S.~C., H.~Li, S.~M., H.~S., {A Ranking-Based Cross-Entropy Loss for Early
  Classification of Time Series}, IEEE Transactions on Neural Networks and
  Learning Systems 35~(8) (2024) 11194--11203.
\newblock \href {https://doi.org/10.1109/tnnls.2023.3250203}
  {\path{doi:10.1109/tnnls.2023.3250203}}.

\bibitem{rao1982}
C.~R. Rao, Diversity and dissimilarity coefficients: A unified approach,
  Theoretical Population Biology 21 (1982) 24--43.
\newblock \href {https://doi.org/10.1016/0040-5809(82)90004-1}
  {\path{doi:10.1016/0040-5809(82)90004-1}}.

\bibitem{DoxaPrastacos2020}
A.~Doxa, P.~Prastacos, {Using Rao's quadratic entropy to define environmental
  heterogeneity priority areas in the European Mediterranean biome}, Biological
  Conservation 241 (2020) 108366.
\newblock \href {https://doi.org/10.1016/j.biocon.2019.108366}
  {\path{doi:10.1016/j.biocon.2019.108366}}.

\bibitem{10.1371/journal.pone.0185499}
P.~E. Smouse, S.~C. Banks, R.~Peakall, Converting quadratic entropy to
  diversity: Both animals and alleles are diverse, but some are more diverse
  than others, PLOS ONE 12~(10) (2017) 1--19.
\newblock \href {https://doi.org/10.1371/journal.pone.0185499}
  {\path{doi:10.1371/journal.pone.0185499}}.

\bibitem{RePEc:ris:crcrmw:2018_006}
G.~Dionne, G.~Koumou,
  \href{https://ideas.repec.org/p/ris/crcrmw/2018_006.html}{{Machine Learning
  and Risk Management: SVDD Meets RQE}}, Tech. Rep. 18-6, HEC Montreal, Canada
  Research Chair in Risk Management (2018).
\newline\urlprefix\url{https://ideas.repec.org/p/ris/crcrmw/2018_006.html}

\bibitem{9869333}
N.~Niknami, J.~Wu, {Entropy-KL-ML: Enhancing the Entropy-KL-Based Anomaly
  Detection on Software-Defined Networks}, IEEE Transactions on Network Science
  and Engineering 9~(6) (2022) 4458--4467.
\newblock \href {https://doi.org/10.1109/TNSE.2022.3202147}
  {\path{doi:10.1109/TNSE.2022.3202147}}.

\bibitem{e23091122}
S.~Moral, A.~Cano, M.~Gómez-Olmedo, {Computation of Kullback--Leibler
  Divergence in Bayesian Networks}, Entropy 23~(9) (2021) 1122.
\newblock \href {https://doi.org/10.3390/e23091122}
  {\path{doi:10.3390/e23091122}}.

\bibitem{10.5555/2503308.2188387}
G.~Brown, A.~Pocock, M.~J. Zhao, M.~Luján,
  \href{https://dl.acm.org/doi/10.5555/2503308.2188387}{Conditional likelihood
  maximisation: a unifying framework for information theoretic feature
  selection}, The Journal of Machine Learning Research 13~(1) (2012) 27--66.
  https://dl.acm.org/doi/10.5555/2503308.2188387.
\newline\urlprefix\url{https://dl.acm.org/doi/10.5555/2503308.2188387}

\bibitem{doi:10.1021/acs.jctc.3c01052}
M.~Chaimovich, A.~Chaimovich, {Relative Resolution: An Analysis with the
  Kullback--Leibler Entropy}, Journal of Chemical Theory and Computation 20~(5)
  (2024) 2074--2087.
\newblock \href {https://doi.org/10.1021/acs.jctc.3c01052}
  {\path{doi:10.1021/acs.jctc.3c01052}}.

\bibitem{draelos2019connections}
R.~Draelos, {Connections: Log-Likelihood, Cross-Entropy, KL-Divergence,
  Logistic Regression, and Neural Networks}, GlassBox MedicineAvailable online:
  https://glassboxmedicine.com/2019/12/07/connections-log-likelihood-cross-entropy-kl-divergence-logistic-regression-and-neural-networks/(Accessed
  30 September 2024) (2019).

\bibitem{DeLaPavaPanche2019}
I.~De~La Pava~Panche, A.~M. Alvarez-Meza, A.~Orozco-Gutierrez, A data-driven
  measure of effective connectivity based on renyi's $\alpha$-entropy,
  Frontiers in Neuroscience 13 (2019) 1277.
\newblock \href {https://doi.org/10.3389/fnins.2019.01277}
  {\path{doi:10.3389/fnins.2019.01277}}.

\bibitem{Rioul2023}
O.~Rioul, The interplay between error, total variation, alpha-entropy and
  guessing: Fano and pinsker direct and reverse inequalities, Entropy 25~(7)
  (2023) 978.
\newblock \href {https://doi.org/10.3390/e25070978}
  {\path{doi:10.3390/e25070978}}.

\bibitem{Berezinski2015}
P.~Berezinski, B.~Jasiul, M.~Szpyrka, An entropy-based network anomaly
  detection method, Entropy 17~(4) (2015) 2367--2408.
\newblock \href {https://doi.org/10.3390/e17042367}
  {\path{doi:10.3390/e17042367}}.

\bibitem{Sharma2015}
R.~Sharma, R.~B. Pachori, U.~R. Acharya, Application of entropy measures on
  intrinsic mode functions for the automated identification of focal
  electroencephalogram signals, Entropy 17~(2) (2015) 669--691.
\newblock \href {https://doi.org/10.3390/e17020669}
  {\path{doi:10.3390/e17020669}}.

\bibitem{czarnecki2015extremeentropymachinesrobust}
W.~M. Czarnecki, J.~Tabor, {Extreme entropy machines: Robust information
  theoretic classification}, Pattern Analysis and Applications 20~(2) (2017)
  383--400.
\newblock \href {https://doi.org/10.1007/s10044-015-0497-8}
  {\path{doi:10.1007/s10044-015-0497-8}}.

\bibitem{Sluga2017}
D.~Sluga, U.~Lotrič, Quadratic mutual information feature selection, Entropy
  19~(4) (2017) 157.
\newblock \href {https://doi.org/10.3390/e19040157}
  {\path{doi:10.3390/e19040157}}.

\bibitem{9254346}
N.~Gowdra, R.~Sinha, S.~MacDonell, Examining convolutional feature extraction
  using maximum entropy (me) and signal-to-noise ratio (snr) for image
  classification, in: IECON 2020 The 46th Annual Conference of the IEEE
  Industrial Electronics Society, 2020, pp. 471--476.
\newblock \href {https://doi.org/10.1109/IECON43393.2020.9254346}
  {\path{doi:10.1109/IECON43393.2020.9254346}}.

\bibitem{lake2005}
D.~E. Lake, Renyi entropy measures of heart rate gaussianity, IEEE Transactions
  on Biomedical Engineering 53 (2005) 21--27.
\newblock \href {https://doi.org/10.1109/TBME.2005.859782}
  {\path{doi:10.1109/TBME.2005.859782}}.

\bibitem{mammone2012}
N.~Mammone, F.~La~Foresta, F.~C. Morabito, Automatic artifact rejection from
  multichannel scalp eeg by wavelet ica, IEEE Sensors Journal 12 (2012)
  533--542.
\newblock \href {https://doi.org/10.1109/JSEN.2011.2115236}
  {\path{doi:10.1109/JSEN.2011.2115236}}.

\bibitem{poza2008}
J.~Poza, J.~Escudero, R.~Hornero, A.~Fernandez, C.~I. Sanchez, Regional
  analysis of spontaneous meg rhythms in patients with alzheimer's disease
  using spectral entropies, Annals of Biomedical Engineering 36~(1) (2008)
  141--152.
\newblock \href {https://doi.org/10.1007/s10439-007-9402-y}
  {\path{doi:10.1007/s10439-007-9402-y}}.

\bibitem{ShangEtAl2020}
Y.~Shang, G.~Lu, Y.~Kang, Z.~Zhou, B.~Duan, C.~Zhang, A multi-fault diagnosis
  method based on modified sample entropy for lithium-ion battery strings,
  Journal of Power Sources 446 (2020) 227275.
\newblock \href {https://doi.org/10.1016/j.jpowsour.2019.227275}
  {\path{doi:10.1016/j.jpowsour.2019.227275}}.

\bibitem{SilvaEtAl2016}
L.~E.~V. Silva, A.~C.~S. Senra~Filho, V.~P.~S. Fazan, J.~C. Felipe, L.~O.
  Murta~Junior, Two-dimensional sample entropy: assessing image texture through
  irregularity, Biomedical Physics \& Engineering Express 2~(4) (2016) 045002.
\newblock \href {https://doi.org/10.1088/2057-1976/2/4/045002}
  {\path{doi:10.1088/2057-1976/2/4/045002}}.

\bibitem{LiaoJan2016}
F.~Liao, Y.~K. Jan, {Using Modified Sample Entropy to Characterize
  Aging-Associated Microvascular Dysfunction}, Frontiers in Physiology 7
  (2016).
\newblock \href {https://doi.org/10.3389/fphys.2016.00126}
  {\path{doi:10.3389/fphys.2016.00126}}.

\bibitem{Lake2002}
D.~E. Lake, J.~S. Richman, M.~P. Griffin, J.~R. Moorman, Sample entropy
  analysis of neonatal heart rate variability, American Journal of Physiology -
  Regulatory, Integrative and Comparative Physiology 283 (2002) 301--312.
\newblock \href {https://doi.org/10.1152/ajpregu.00069.2002}
  {\path{doi:10.1152/ajpregu.00069.2002}}.

\bibitem{NguyenLiuLin2020}
Q.~D. Nam~N., A.~B. Liu, C.~W. Lin, {Development of a Neurodegenerative Disease
  Gait Classification Algorithm Using Multiscale Sample Entropy and Machine
  Learning Classifiers}, Entropy 22~(12) (2020) 1340.
\newblock \href {https://doi.org/10.3390/e22121340}
  {\path{doi:10.3390/e22121340}}.

\bibitem{lake2011}
D.~E. Lake, J.~R. Moorman, Accurate estimation of entropy in very short
  physiological time series: the problem of atrial fibrillation detection in
  implanted ventricular devices., American Journal of Physiology - Heart and
  Circulatory Physiology 300~(1) (2011) H319–H325.
\newblock \href {https://doi.org/10.1152/ajpheart.00561.2010}
  {\path{doi:10.1152/ajpheart.00561.2010}}.

\bibitem{HumeauHeurtier2018}
A.~Humeau-Heurtier, {Evaluation of Systems’ Irregularity and Complexity:
  Sample Entropy, Its Derivatives, and Their Applications across Scales and
  Disciplines}, Entropy 20~(10) (2018) 794.
\newblock \href {https://doi.org/10.3390/e20100794}
  {\path{doi:10.3390/e20100794}}.

\bibitem{BelyaevEtAl2023}
M.~Belyaev, M.~Murugappan, A.~Velichko, D.~Korzun, {Entropy-Based Machine
  Learning Model for Fast Diagnosis and Monitoring of Parkinson’s Disease},
  Sensors 23~(20) (2023) 8609.
\newblock \href {https://doi.org/10.3390/s23208609}
  {\path{doi:10.3390/s23208609}}.

\bibitem{LinLin2022}
G.~Lin, A.~Lin, Modified multiscale sample entropy and cross-sample entropy
  based on horizontal visibility graph, Chaos, Solitons \& Fractals 165 (2022)
  112802.
\newblock \href {https://doi.org/10.1016/j.chaos.2022.112802}
  {\path{doi:10.1016/j.chaos.2022.112802}}.

\bibitem{KarevanSuykens2018}
Z.~Karevan, J.~A.~K. Suykens, {Transductive Feature Selection Using
  Clustering-Based Sample Entropy for Temperature Prediction in Weather
  Forecasting}, Entropy 20~(4) (2018) 264.
\newblock \href {https://doi.org/10.3390/e20040264}
  {\path{doi:10.3390/e20040264}}.

\bibitem{PhysRev.106.620}
E.~T. Jaynes, {Information Theory and Statistical Mechanics}, Physical Review
  106~(4) (1957) 620--630.
\newblock \href {https://doi.org/10.1103/PhysRev.106.620}
  {\path{doi:10.1103/PhysRev.106.620}}.

\bibitem{GuhaVelegol2023}
R.~Guha, D.~Velegol, Harnessing shannon entropy-based descriptors in machine
  learning models to enhance the prediction accuracy of molecular properties,
  Journal of Cheminformatics 15 (2023) 54.
\newblock \href {https://doi.org/10.1186/s13321-023-00712-0}
  {\path{doi:10.1186/s13321-023-00712-0}}.

\bibitem{DeMedeirosEtAl2023}
K.~DeMedeiros, A.~Hendawi, M.~Alvarez, {A Survey of AI-Based Anomaly Detection
  in IoT and Sensor Networks}, Sensors 23~(3) (2023) 1352.
\newblock \href {https://doi.org/10.3390/s23031352}
  {\path{doi:10.3390/s23031352}}.

\bibitem{Evans2024}
S.~C. Evans, T.~Shah, H.~Huang, S.~P. Ekanayake, The entropy economy and the
  kolmogorov learning cycle: Leveraging the intersection of machine learning
  and algorithmic information theory to jointly optimize energy and learning,
  Physica D 461 (2024) 134051.
\newblock \href {https://doi.org/10.1016/j.physd.2024.134051}
  {\path{doi:10.1016/j.physd.2024.134051}}.

\bibitem{HintonSejnowski1986}
G.~E. Hinton, T.~J. Sejnowski, Learning and relearning in boltzmann machines,
  in: D.~E. Rumelhart, J.~L. McClelland (Eds.), Parallel Distributed
  Processing: Explorations in the Microstructure of Cognition, Volume 1:
  Foundations, Springer, 2016, Ch. The MIT Press, p. 282–317.
\newblock \href {https://doi.org/10.1007/978-3-662-48410-4_1}
  {\path{doi:10.1007/978-3-662-48410-4_1}}.

\bibitem{Oh2020}
S.~Oh, A.~Baggag, H.~Nha, Entropy, free energy, and work of restricted
  boltzmann machines, Entropy 22 (2020) 538.
\newblock \href {https://doi.org/10.3390/e22050538}
  {\path{doi:10.3390/e22050538}}.

\bibitem{marullo2021}
C.~Marullo, E.~Agliari, Boltzmann machines as generalized hopfield networks: A
  review of recent results and outlooks, Entropy 23 (2021) 34.
\newblock \href {https://doi.org/10.3390/e23010034}
  {\path{doi:10.3390/e23010034}}.

\bibitem{smart2021}
M.~Smart, A.~Zilman, {On the mapping between Hopfield networks and Restricted
  Boltzmann Machines}, arXiv: 2101.11744 (2021).
\newblock \href {https://doi.org/10.48550/arXiv.2101.11744}
  {\path{doi:10.48550/arXiv.2101.11744}}.

\bibitem{HorenkoeSPA}
I.~Horenko, On a scalable entropic breaching of the overfitting barrier for
  small data problems in machine learning, Neural Computation 32 (2020)
  1563--1579.
\newblock \href {https://doi.org/10.1162/neco_a_01296}
  {\path{doi:10.1162/neco_a_01296}}.

\bibitem{HorenkoeSPAplus}
E.~Vecchi, L.~Posp\'{\i}\v{s}il, S.~Albrecht, T.~J. O'Kane, I.~Horenko, espa+:
  Scalable entropy-optimal machine learning classification for small data
  problems, Neural Computation 34 (2022) 1220--1255.
\newblock \href {https://doi.org/10.1162/neco_a_01490}
  {\path{doi:10.1162/neco_a_01490}}.

\bibitem{e21060540}
N.~Rodriguez, L.~Barba, P.~Alvarez, G.~Cabrera-Guerrero, Stationary
  wavelet-fourier entropy and kernel extreme learning for bearing multi-fault
  diagnosis, Entropy 21~(6) (2019) 540.
\newblock \href {https://doi.org/10.3390/e21060540}
  {\path{doi:10.3390/e21060540}}.

\bibitem{SoltanianEtAl2019}
A.~R. Soltanian, N.~Rabiei, F.~Bahreini, Feature selection in microarray data
  using entropy information, in: H.~Husi (Ed.), Computational Biology, Exon
  Publications, 2019, Ch.~10.
\newblock \href {https://doi.org/10.15586/computationalbiology.2019.ch10}
  {\path{doi:10.15586/computationalbiology.2019.ch10}}.

\bibitem{HoayekRulliere2023}
A.~Hoayek, D.~Rullière, \href{https://hal.science/hal-03812055v1}{Assessing
  clustering methods using shannon's entropy}, Information Sciences 689 (2025)
  121510.
\newblock \href {https://doi.org/https://doi.org/10.1016/j.ins.2024.121510}
  {\path{doi:https://doi.org/10.1016/j.ins.2024.121510}}.
\newline\urlprefix\url{https://hal.science/hal-03812055v1}

\bibitem{Finnegan2017}
A.~Finnegan, J.~S. Song, Maximum entropy methods for extracting the learned
  features of deep neural networks, PLOS Computational Biology 13 (2017)
  e1005836.
\newblock \href {https://doi.org/10.1371/journal.pcbi.1005836}
  {\path{doi:10.1371/journal.pcbi.1005836}}.

\bibitem{MeMe2019}
D.~Granziol, B.~Ru, S.~Zohren, X.~Dong, M.~Osborne, S.~Roberts, Meme: An
  accurate maximum entropy method for efficient approximations in large-scale
  machine learning, Entropy 21 (2019) 551.
\newblock \href {https://doi.org/10.3390/e21060551}
  {\path{doi:10.3390/e21060551}}.

\bibitem{HorenkoEOS2022}
I.~Horenko, Cheap robust learning of data anomalies with analytically solvable
  entropic outlier sparsification, Proceedings of the National Academy of
  Sciences 119 (2022) e2119659119.
\newblock \href {https://doi.org/10.1073/pnas.2119659119}
  {\path{doi:10.1073/pnas.2119659119}}.

\bibitem{YANG20134523}
Z.~Yang, J.~Lei, K.~Fan, Y.~Lai, Keyword extraction by entropy difference
  between the intrinsic and extrinsic mode, Physica A 392~(19) (2013)
  4523--4531.
\newblock \href {https://doi.org/10.1016/j.physa.2013.05.052}
  {\path{doi:10.1016/j.physa.2013.05.052}}.

\bibitem{Guo2023}
X.~Guo, X.~Li, R.~Xu, Fast policy learning for linear quadratic control with
  entropy regularization, arXiv:2311.14168 (2023).
\newblock \href {https://doi.org/10.48550/arXiv.2311.14168}
  {\path{doi:10.48550/arXiv.2311.14168}}.

\bibitem{HorenkoCheapEntropySparsified}
I.~Horenko, E.~Vecchi, J.~Kardo\v{s}, A.~Wächter, O.~Schenk, T.~O'Kane,
  P.~Gagliardine, S.~Gerber, On cheap entropy-sparsified regression learning,
  Proceedings of the National Academy of Science 120 (2023) e2214972120.
\newblock \href {https://doi.org/10.1073/pnas.2214972120}
  {\path{doi:10.1073/pnas.2214972120}}.

\bibitem{CoifmanWickerhauser1992}
R.~Coifman, M.~Wickerhauser, Entropy-based algorithms for best basis selection,
  IEEE Transactions on Information Theory 38~(2) (1992) 713--718.
\newblock \href {https://doi.org/10.1109/18.119732}
  {\path{doi:10.1109/18.119732}}.

\bibitem{kapur1992}
J.~N. Kapur, H.~K. Kesavan, Entropy optimization principles and their
  applications, in: Entropy and Energy Dissipation in Water Resources,
  Springer, Berlin/Heidelberg, Germany, 1992, pp. 3--20.
\newblock \href {https://doi.org/10.1007/978-94-011-2430-0_1}
  {\path{doi:10.1007/978-94-011-2430-0_1}}.

\bibitem{Wang2020}
K.~C. Wang, {Robust Audio Content Classification Using Hybrid-Based SMD and
  Entropy-Based VAD}, Entropy 22~(2) (2020) 183.
\newblock \href {https://doi.org/10.3390/e22020183}
  {\path{doi:10.3390/e22020183}}.

\bibitem{ManzoMartinez2022}
A.~Manzo-Martínez, F.~Gaxiola, G.~Ramírez-Alonso, F.~Martínez-Reyes, {A
  Comparative Study in Machine Learning and Audio Features for Kitchen Sounds
  Recognition}, Computación y Sistemas 26~(2) (2022) 4244.
\newblock \href {https://doi.org/10.13053/cys-26-2-4244}
  {\path{doi:10.13053/cys-26-2-4244}}.

\bibitem{Civera2022}
M.~Civera, C.~Surace, {An Application of Instantaneous Spectral Entropy for the
  Condition Monitoring of Wind Turbines}, Applied Sciences 12 (2022) 1059.
\newblock \href {https://doi.org/10.3390/app12031059}
  {\path{doi:10.3390/app12031059}}.

\bibitem{Ajmal2007}
M.~Ajmal, A.~Kushki, K.~N. Plataniotis, {Time-Compression of Speech in
  Information Talks Using Spectral Entropy}, in: {Eighth International Workshop
  on Image Analysis for Multimedia Interactive Services (WIAMIS '07)}, 2007,
  pp. 80--80.
\newblock \href {https://doi.org/10.1109/WIAMIS.2007.80}
  {\path{doi:10.1109/WIAMIS.2007.80}}.

\bibitem{Kapucu2016}
F.~E. Kapucu, I.~Välkki, J.~E. Mikkonen, C.~Leone, K.~Lenk, J.~M.~A.
  Tanskanen, J.~A.~K. Hyttinen, {Spectral Entropy Based Neuronal Network
  Synchronization Analysis Based on Microelectrode Array Measurements},
  Frontiers in Computational Neuroscience 10 (2016) 112.
\newblock \href {https://doi.org/10.3389/fncom.2016.00112}
  {\path{doi:10.3389/fncom.2016.00112}}.

\bibitem{Ra2021}
J.~S. Ra, T.~Li, Y.~Li, A novel spectral entropy-based index for assessing the
  depth of anaesthesia, Brain Informatics 8~(10) (2021) 1--12.
\newblock \href {https://doi.org/10.1186/s40708-021-00130-8}
  {\path{doi:10.1186/s40708-021-00130-8}}.

\bibitem{Liu2024}
S.~Liu, Z.~Li, G.~Wang, X.~Qiu, T.~Liu, J.~Cao, D.~Zhang, {Spectral–Spatial
  Feature Fusion for Hyperspectral Anomaly Detection}, Sensors 24~(5) (2024)
  1652.
\newblock \href {https://doi.org/10.3390/s24051652}
  {\path{doi:10.3390/s24051652}}.

\bibitem{Rademan2023}
M.~W. Rademan, D.~J.~J. Versfeld, J.~A. du~Preez, {Soft-Output Signal Detection
  for Cetacean Vocalizations Using Spectral Entropy, K-Means Clustering and the
  Continuous Wavelet Transform}, Ecological Informatics 74 (2023) 101990.
\newblock \href {https://doi.org/10.1016/j.ecoinf.2023.101990}
  {\path{doi:10.1016/j.ecoinf.2023.101990}}.

\bibitem{oida1997}
E.~Oida, T.~Moritani, Y.~Yamori, Tone-entropy analysis on cardiac recovery
  after dynamic exercise, Journal of Applied Physiology 82 (1997) 1794--1801.
\newblock \href {https://doi.org/10.1152/jappl.1997.82.6.1794#}
  {\path{doi:10.1152/jappl.1997.82.6.1794#}}.

\bibitem{Khandoker2019}
A.~H. Khandoker, Y.~Al~Zaabi, H.~F. Jelinek, What can tone and entropy tell us
  about risk of cardiovascular diseases?, in: 2019 Computing in Cardiology
  (CinC), 2019, pp. 1--4.
\newblock \href {https://doi.org/10.22489/CinC.2019.120}
  {\path{doi:10.22489/CinC.2019.120}}.

\bibitem{Khandoker2015}
A.~Khandoker, C.~Karmakar, Y.~Kimura, M.~Endo, S.~Oshio, M.~Palaniswami, {Tone
  Entropy Analysis of Foetal Heart Rate Variability}, Entropy 17~(3) (2015)
  1042--1053.
\newblock \href {https://doi.org/10.3390/e17031042}
  {\path{doi:10.3390/e17031042}}.

\bibitem{Karmakar2013}
C.~K. Karmakar, A.~H. Khandoker, M.~Palaniswami, {Multi-scale Tone Entropy in
  differentiating physiologic and synthetic RR time series}, in: 2013 35th
  Annual International Conference of the IEEE Engineering in Medicine and
  Biology Society (EMBC), 2013, pp. 6135--6138.
\newblock \href {https://doi.org/10.1109/EMBC.2013.6610953}
  {\path{doi:10.1109/EMBC.2013.6610953}}.

\bibitem{Amigo2014}
J.~M. Amig\'{o}, A.~Gim\'{e}nez, A simplified algorithm for the topological
  entropy of multimodal maps, Entropy 16~(2) (2014) 627--644.
\newblock \href {https://doi.org/10.3390/e16020627}
  {\path{doi:10.3390/e16020627}}.

\bibitem{Amigo2015}
J.~M. Amig\'{o}, A.~Gim\'{e}nez, Formulas for the topological entropy of
  multimodal maps based on min-max symbols, Discrete and Continuous Dynamical
  Systems B 20~(10) (2015) 3415--3434.
\newblock \href {https://doi.org/10.3934/dcdsb.2015.20.3415}
  {\path{doi:10.3934/dcdsb.2015.20.3415}}.

\bibitem{Lum2013}
P.~Lum, G.~Singh, A.~Lehman, T.~Ishkanov, M.~Vejdemo-Johansson, M.~Alagappan,
  J.~Carlsson, G.~Carlsson, Extracting insights from the shape of complex data
  using topology, Scientific Reports 3~(1) (2013) 1236.
\newblock \href {https://doi.org/10.1038/srep01236}
  {\path{doi:10.1038/srep01236}}.

\bibitem{mccullough2017}
M.~McCullough, M.~Small, H.~H.~C. Iu, T.~Stemler, Multiscale ordinal network
  analysis of human cardiac dynamics, Philosophical Transactions of the Royal
  Society A 375 (2017) 20160292.
\newblock \href {https://doi.org/10.1098/rsta.2016.0292}
  {\path{doi:10.1098/rsta.2016.0292}}.

\bibitem{Zhao2022}
Y.~Zhao, H.~Zhang, Quantitative performance assessment of cnn units via
  topological entropy calculation, arXiv:2103.09716 (2022).
\newblock \href {http://arxiv.org/abs/2103.09716} {\path{arXiv:2103.09716}},
  \href {https://doi.org/10.48550/arXiv.2103.09716}
  {\path{doi:10.48550/arXiv.2103.09716}}.

\bibitem{JimenezAlonso2019}
J.~F. Jiménez-Alonso, J.~López-Martínez, J.~L. Blanco-Claraco,
  R.~González-Díaz, A.~Sáez, A topological entropy-based approach for damage
  detection of civil engineering structures, in: 5th International Conference
  on Mechanical Models in Structural Engineering (CMMoST 2019), 2019, pp.
  55--62.

\bibitem{Rong2018}
L.~Rong, P.~Shang, Topological entropy and geometric entropy and their
  application to the horizontal visibility graph for financial time series,
  Nonlinear Dynamics 92~(1) (2018) 41--58.
\newblock \href {https://doi.org/10.1007/s11071-018-4120-6}
  {\path{doi:10.1007/s11071-018-4120-6}}.

\bibitem{Rucco2017}
M.~Rucco, R.~Gonzalez-Diaz, M.~Jimenez, N.~Atienza, C.~Cristalli,
  E.~Concettoni, A.~Ferrante, E.~Merelli, \href{10.1016/j.sigpro.2016.12.006}{A
  new topological entropy-based approach for measuring similarities among
  piecewise linear functions}, Signal Processing 134 (2017) 130--138.
\newblock \href {https://doi.org/https://doi.org/10.1016/j.sigpro.2016.12.006}
  {\path{doi:https://doi.org/10.1016/j.sigpro.2016.12.006}}.
\newline\urlprefix\url{10.1016/j.sigpro.2016.12.006}

\bibitem{PhysRevLett85461}
T.~Schreiber, Measuring information transfer, Physical Review Letters 85~(2)
  (2000) 461--464.
\newblock \href {https://doi.org/10.1103/PhysRevLett.85.461}
  {\path{doi:10.1103/PhysRevLett.85.461}}.

\bibitem{RePEc38}
C.~W.~J. Granger,
  \href{https://ideas.repec.org/a/ecm/emetrp/v37y1969i3p424-38.html}{{Investigating
  Causal Relations by Econometric Models and Cross-Spectral Methods}},
  Econometrica 37~(3) (1969) 424--438.
\newblock \href {https://doi.org/https://doi.org/10.2307/1912791}
  {\path{doi:https://doi.org/10.2307/1912791}}.
\newline\urlprefix\url{https://ideas.repec.org/a/ecm/emetrp/v37y1969i3p424-38.html}

\bibitem{moldovan2021}
A.~Moldovan, A.~Caţaron, A.~Răzvan, Learning in convolutional neural networks
  accelerated by transfer entropy, Entropy 23~(9) (2021) 1281.
\newblock \href {https://doi.org/10.3390/e23091218}
  {\path{doi:10.3390/e23091218}}.

\bibitem{moldovan2024transferentropygraphconvolutional}
A.~Moldovan, A.~Caţaron, R.~Andonie, Transfer entropy in graph convolutional
  neural networks, arXiv:2406.06632 (2024).
\newblock \href {https://doi.org/10.48550/arXiv.2406.06632}
  {\path{doi:10.48550/arXiv.2406.06632}}.

\bibitem{herzog2017}
S.~Herzog, C.~Tetzlaff, F.~Wörgötter, Transfer entropy-based feedback
  improves performance in artificial neural networks, arXiv:1706.04265 (2017).
\newblock \href {https://doi.org/10.48550/arXiv.1706.04265}
  {\path{doi:10.48550/arXiv.1706.04265}}.

\bibitem{9837007}
Z.~Duan, H.~Xu, Y.~Huang, J.~Feng, Y.~Wang, Multivariate time series
  forecasting with transfer entropy graph, Tsinghua Science and Technology
  28~(1) (2023) 141--149.
\newblock \href {https://doi.org/10.26599/TST.2021.9010081}
  {\path{doi:10.26599/TST.2021.9010081}}.

\bibitem{Amblard2013}
P.~O. Amblard, O.~J.~J. Michel, The relation between granger causality and
  directed information theory: A review, Entropy 15~(1) (2013) 113--143.
\newblock \href {https://doi.org/10.3390/e15010113}
  {\path{doi:10.3390/e15010113}}.

\bibitem{Alomani2023}
G.~Alomani, M.~Kayid, Further properties of tsallis entropy and its
  application, Entropy 25~(2) (2023) 199.
\newblock \href {https://doi.org/10.3390/e25020199}
  {\path{doi:10.3390/e25020199}}.

\bibitem{Sharma2019}
S.~Sharma, I.~Bassi, Efficacy of tsallis entropy in clustering categorical
  data, in: 2019 IEEE Bombay Section Signature Conference (IBSSC), 2019, pp.
  1--5.
\newblock \href {https://doi.org/10.1109/IBSSC47189.2019.8973057}
  {\path{doi:10.1109/IBSSC47189.2019.8973057}}.

\bibitem{Wu2022}
D.~Wu, H.~Jia, L.~Abualigah, Z.~Xing, R.~Zheng, H.~Wang, M.~Altalhi, {Enhance
  Teaching-Learning-Based Optimization for Tsallis-Entropy-Based Feature
  Selection Classification Approach}, Processes 10~(2) (2022) 360.
\newblock \href {https://doi.org/10.3390/pr10020360}
  {\path{doi:10.3390/pr10020360}}.

\bibitem{Naidu2017}
M.~S.~R. Naidu, P.~R. Kumar, Tsallis entropy based image thresholding for image
  segmentation, in: Computational Intelligence in Data Mining, Vol. 556,
  Springer, Singapore, 2017, pp. 371--379.
\newblock \href {https://doi.org/10.1007/978-981-10-3874-7_34}
  {\path{doi:10.1007/978-981-10-3874-7_34}}.

\bibitem{Kalimari2008}
M.~Kalimeri, C.~Papadimitriou, G.~Balasis, K.~Eftaxias, Dynamical complexity
  detection in pre-seismic emissions using nonadditive tsallis entropy, Physica
  A 387~(5) (2008) 1161--1172.
\newblock \href {https://doi.org/10.1016/j.physa.2007.10.053}
  {\path{doi:10.1016/j.physa.2007.10.053}}.

\bibitem{e25010154}
L.~A. Belanche-Muñoz, M.~Wiejacha, {Analysis of Kernel Matrices via the von
  Neumann Entropy and Its Relation to RVM Performances}, Entropy 25~(1) (2023)
  154.
\newblock \href {https://doi.org/10.3390/e25010154}
  {\path{doi:10.3390/e25010154}}.

\bibitem{Hu2023}
F.~Hu, K.~Tian, Z.~K. Zhang, Identifying vital nodes in hypergraphs based on
  von neumann entropy, Entropy 25~(9) (2023) 1263.
\newblock \href {https://doi.org/10.3390/e25091263}
  {\path{doi:10.3390/e25091263}}.

\bibitem{Chen2018}
P.~Y. Chen, L.~Wu, S.~Liu, I.~Rajapakse, Fast incremental von neumann graph
  entropy computation: Theory, algorithm, and applications, arXiv:1805.11769
  (2019).
\newblock \href {https://doi.org/10.48550/arXiv.1805.11769}
  {\path{doi:10.48550/arXiv.1805.11769}}.

\bibitem{YeEtAl2017}
C.~Ye, R.~C. Wilson, E.~R. Hancock, Network analysis using entropy component
  analysis, Journal of Complex Networks 6~(5) (2017) 831.
\newblock \href {https://doi.org/https://doi.org/10.1093/comnet/cnx045}
  {\path{doi:https://doi.org/10.1093/comnet/cnx045}}.

\bibitem{huang2024entropy}
Y.~Huang, Y.~Zhao, A.~Capstick, F.~Palermo, H.~Haddadi, P.~Barnaghi, Analyzing
  entropy features in time-series data for pattern recognition in neurological
  conditions, Artificial Intelligence in Medicine 150 (2024) 102821.
\newblock \href {https://doi.org/10.1016/j.artmed.2024.102821}
  {\path{doi:10.1016/j.artmed.2024.102821}}.

\bibitem{Rosso2001}
O.~A. Rosso, S.~Blanco, J.~Yordanova, V.~Kolev, A.~Figliola, M.~Schürmann,
  E.~Başar, {Wavelet entropy: a new tool for analysis of short duration brain
  electrical signals}, Journal of Neuroscience Methods 105~(1) (2001) 65--75.
\newblock \href {https://doi.org/10.1016/S0165-0270(00)00356-3}
  {\path{doi:10.1016/S0165-0270(00)00356-3}}.

\bibitem{Hu20232}
P.~Hu, C.~Zhao, J.~Huang, T.~Song, Intelligent and small samples gear fault
  detection based on wavelet analysis and improved cnn, Processes 11 (2023)
  2969.
\newblock \href {https://doi.org/10.3390/pr11102969}
  {\path{doi:10.3390/pr11102969}}.

\bibitem{Cuomo1993}
K.~M. Cuomo, A.~V. Oppenheim, S.~H. Strogatz, Synchronization of lorenz-based
  chaotic circuits with applications to communications, IEEE Transactions on
  Circuits and Systems II 40~(10) (1993) 626--633.
\newblock \href {https://doi.org/10.1109/82.246163}
  {\path{doi:10.1109/82.246163}}.

\bibitem{AmigoDiscreteEnt2007}
J.~M. Amig\'{o}, L.~Kocarev, I.~Tomovski, Discrete entropy, Physica D 228~(1)
  (2007) 77--85.
\newblock \href {https://doi.org/10.1016/j.physd.2007.03.001}
  {\path{doi:10.1016/j.physd.2007.03.001}}.

\bibitem{AmigoDiscreteLE2007}
J.~M. Amig\'{o}, L.~Kocarev, J.~Szczepanski, Discrete lyapunov exponent and
  resistance to differential cryptanalysis, IEEE Transactions on Circuits and
  Systems II 54~(10) (2007) 882--886.
\newblock \href {https://doi.org/10.1109/TCSII.2007.901576}
  {\path{doi:10.1109/TCSII.2007.901576}}.

\bibitem{EntropyInCrypto2022}
B.~Zolfaghari, K.~Bibak, T.~Koshiba, The odyssey of entropy: Cryptography,
  Entropy 24 (2022) 266.
\newblock \href {https://doi.org/10.3390/e24020266}
  {\path{doi:10.3390/e24020266}}.

\bibitem{Hastie2009}
T.~Hastie, R.~Tibshirani, J.~Friedman, The Elements of Statistical Learning,
  second edition Edition, Springer, New York, 2009.
\newblock \href {https://doi.org/10.1007/b94608} {\path{doi:10.1007/b94608}}.

\bibitem{Jenssen2010}
R.~Jenssen, Kernel entropy component analysis, IEEE Transactions on Pattern
  Analysis and Machine Intelligence 32~(5) (2010) 847--860.
\newblock \href {https://doi.org/10.1109/TPAMI.2009.100}
  {\path{doi:10.1109/TPAMI.2009.100}}.

\bibitem{Paninski2003}
L.~Paninski, Estimation of entropy and mutual information, Neural Computation
  15 (2003) 1191--1253.
\newblock \href {https://doi.org/10.1162/089976603321780272}
  {\path{doi:10.1162/089976603321780272}}.

\bibitem{AmigoMonetti2016}
J.~M. Amigó, R.~Monetti, B.~Graff, G.~Graff, Computing algebraic transfer
  entropy and coupling directions via transcripts, Chaos 26 (2019) 113115.
\newblock \href {https://doi.org/10.1063/1.4967803}
  {\path{doi:10.1063/1.4967803}}.

\bibitem{Yeung1991}
R.~W. Yeung, A new outlook on shannon's information measures, IEEE Transactions
  on Information Theory 37 (1991) 466--474.
\newblock \href {https://doi.org/10.1109/18.79902}
  {\path{doi:10.1109/18.79902}}.

\bibitem{Mediano2024}
K.~J.~A. Down, P.~A.~M. Mediano, Algebraic representations of entropy and
  fixed-parity information quantities, arXiv: 2409.04845 (2024).
\newblock \href {https://doi.org/10.48550/arXiv.2409.04845}
  {\path{doi:10.48550/arXiv.2409.04845}}.

\bibitem{Baudot2015}
P.~Baudot, D.~Bennequin, The homological nature of entropy, Entropy 17 (2015)
  3253--3318.
\newblock \href {https://doi.org/10.3390/e17053253}
  {\path{doi:10.3390/e17053253}}.

\bibitem{Schoelkopf1998}
B.~Schölkopf, A.~Smola, K.~R. Müller, Nonlinear component analysis as a
  kernel eigenvalue problem, Neural Computation 10 (1998) 1299–1319.
\newblock \href {https://doi.org/10.1162/089976698300017467}
  {\path{doi:10.1162/089976698300017467}}.

\bibitem{James2017}
R.~G. James, P.~Crutchfield, Multivariate dependence beyond shannon
  information, Entropy 19 (2017) 531.
\newblock \href {https://doi.org/10.3390/e19100531}
  {\path{doi:10.3390/e19100531}}.

\bibitem{James2019}
R.~G. James, J.~Emenheiser, P.~Crutchfield, Unique information and secret key
  agreement, Entropy 21 (2019) 12.
\newblock \href {https://doi.org/10.3390/e21010012}
  {\path{doi:10.3390/e21010012}}.

\bibitem{Ehrlich2024}
D.~A. Ehrlich, K.~Schick-Poland, A.~Makkeh, F.~Lanfermann, P.~Wollstadt,
  M.~Wibral, Partial information decomposition for continuous variables based
  on shared exclusions: Analytical formulation and estimation, Physical Review
  E 110 (2024) 014115.
\newblock \href {https://doi.org/10.1103/PhysRevE.110.014115}
  {\path{doi:10.1103/PhysRevE.110.014115}}.

\bibitem{Bosyk2016}
G.~M. Bosyk, S.~Zozor, F.~Holik, M.~Portesi, P.~W. Lamberti, A family of
  generalized quantum entropies: definition and properties, Quantum Information
  Processing 15 (2016) 3393--3420.
\newblock \href {https://doi.org/10.1007/s11128-016-1329-5}
  {\path{doi:10.1007/s11128-016-1329-5}}.

\end{thebibliography}
\bibliographystyle{elsarticle-num}

\end{document}